\documentclass[10pt]{article}

\usepackage{iclr2027_conference,times}
\iclrfinalcopy

\usepackage{amsmath,amssymb,mathtools,bm}
\usepackage{amsthm}
\usepackage{graphicx}
\graphicspath{{select/}}
\usepackage{booktabs}
\usepackage{array}
\usepackage{longtable}
\usepackage{float}
\usepackage{tikz}
\usepackage{pgfplots}
\pgfplotsset{compat=1.16}
\usepackage{subcaption}
\usepackage{enumitem}
\usepackage{xspace}
\usepackage{placeins}
\usepackage{hyperref}
\hypersetup{colorlinks=true,citecolor=blue!55!black,linkcolor=blue!55!black,urlcolor=blue!55!black}
\usepackage[capitalise]{cleveref}

\newtheorem{theorem}{Theorem}[section]
\newtheorem{proposition}[theorem]{Proposition}

\newtheorem{corollary}[theorem]{Corollary}
\theoremstyle{remark}
\newtheorem*{proofsketch}{Proof sketch}
\newtheorem{remark}[theorem]{Remark}

\newcommand{\Rtwo}{\ensuremath{R_{2}}\xspace}
\newcommand{\Rfour}{\ensuremath{R_{4}}\xspace}
\newcommand{\Rsix}{\ensuremath{R_{6}}\xspace}
\newcommand{\Reight}{\ensuremath{R_{8}}\xspace}
\newcommand{\OOD}{OOD\xspace}
\newcommand{\ERM}{ERM\xspace}
\newcommand{\IRM}{IRM\xspace}
\newcommand{\diag}{\operatorname{diag}}
\newcommand{\E}{\mathbb{E}}
\newcommand{\F}{\mathcal{F}}
\newcommand{\Var}{\operatorname{Var}}
\newcommand{\Cov}{\operatorname{Cov}}
\newcommand{\norm}[1]{\left\lVert#1\right\rVert}

\title{From Objectives to What Models Learn:\\A Landau Theory of Invariant Learning}
\author{%
Pinli Wang$^{1}$, Yue He$^{2}$, Peng Cui$^{2}$\\
$^{1}$Department of Physics, Tsinghua University, Beijing, China\\
$^{2}$Department of Computer Science and Technology, Tsinghua University, Beijing, China
}

\begin{document}
\maketitle
\fancyhead{}
\begingroup
\renewcommand{\thefootnote}{}
\footnotetext{\raggedright\hspace*{-1.8em}Contact: Pinli Wang (\texttt{wang-pl23@mails.tsinghua.edu.cn}) and Yue He (\texttt{hy865865@gmail.com})}
\addtocounter{footnote}{-1}
\endgroup
\addtocontents{toc}{\protect\setcounter{tocdepth}{-1}}

\begin{abstract}
Invariant-learning objectives pursue similar goals yet produce qualitatively different regularization paths, leaving unclear when shortcuts can be suppressed without damaging stable modes. Our starting point is simple: in the desired shortcut-suppressed regime, shortcut loading is small, so the objective's low-order expansion governs local stability and residual amplitude. This brings the problem into the domain of Landau phase-transition theory. We establish a mathematical isomorphism between the near-critical normal form of predictive-mode learning and Landau theory, identifying the learning objective as an effective free energy and critical-mode amplitude as an order parameter. The resulting low-order objective signatures predict distinct regularization phenotypes: quadratic \Rtwo{} terms shift phase boundaries and enable finite-strength elimination, whereas quartic \Rfour{} terms continuously attenuate acquired modes while leaving nonzero residual loading at finite strengths. Higher-order terms may further shape nonlinear tails. In a canonical bilinear model, we derive exact phase boundaries and equilibrium loadings, identifying conditions for a selective-retention window in which shortcut suppression preserves stable structure. Controlled bilinear and ReLU experiments, including an MNIST construction, support the predicted signature--phenotype relation, while coupled-feature experiments validate its extension to collective modes. The framework connects the mathematical structure of invariant-learning objectives to what models learn as regularization varies.
\end{abstract}

\section{Introduction}

The central promise of invariant learning is simple: exploit heterogeneity across training environments to learn predictive structure that survives distribution shift \citep{peters2016causal,arjovsky2019irm,sagawa2020groupdro,krueger2021rex}. In practice, however, objectives built around this common principle can behave very differently, while their formulas provide little indication of what will happen along the regularization path \citep{gulrajani2021domainbed,zhang2023missing}. This objective--behavior gap turns objective choice and strength selection into an empirical search problem \citep{chen2023pareto}. Fundamentally, it leaves unclear whether an objective can distinguish stable from shortcut modes \citep{geirhos2020shortcut} and suppress the latter without collateral damage.

These limitations call for a structural account of objective behavior: one that makes the regularization phenotype predictable directly from the form of the loss. Hard preselection is an incomplete solution, as spuriousness is mode-wise and is often distributed across interacting collective modes rather than isolated in known, single variables. Conventionally, invariant regularization is viewed as a continuous trade-off, reducing model selection to hunting for a single, fragile performance peak. In reality, suppression is not merely a continuous slide: a change in the local stability of the zero-loading state marks a sharp phase transition. As distinct modes hit this critical boundary at different strengths, the ideal outcome is not a point of compromise, but a robust selective-retention window.

The key to predicting these regularization paths lies in a simple mathematical reality: whether a shortcut is continuously attenuated or eliminated at a finite threshold, its loading in the desired selective-retention state is intrinsically small. The global nonlinear complexity of an invariant penalty is therefore largely irrelevant to this near-critical behavior. What governs the mode is the low-order expansion of the full objective, which determines the stability of the shortcut-free state and the leading behavior of small residual loading. This is precisely the domain of Landau phase-transition theory, which describes the emergence, attenuation, and loss of macroscopic order through a low-order free-energy expansion. Landau theory has successfully characterized critical behavior across diverse physical systems, from magnets to superconductors and ferroelectrics \citep{landau1937phase,landau1980statistical,ginzburg1950superconductivity,hohenberg2015ginzburg}.

We establish a mathematical isomorphism between the near-critical normal forms of predictive-mode learning and Landau theory. Through the lens of magnetization, the learning objective acts as an effective free energy, the critical-mode amplitude as the order parameter, regularization strength as a temperature-like control, and environment-dependent support as a mode-dependent external field. A mode that persists without such support resembles spontaneous ferromagnetic order; a shortcut driven mainly by the field resembles a paramagnetic response; and an intrinsically predictive but environment-sensitive mode resembles a soft magnet. Landau theory then predicts mode onset, equilibrium loading, and phase boundaries, as well as how learned modes respond to weak loss perturbations and how quickly they relax toward equilibrium near criticality.

Starting from concrete invariant-learning objectives, we reduce their local equilibria near a phase boundary to a low-order Landau free energy. Its coefficients form a low-order objective signature of the full loss. A quadratic correction \Rtwo{} changes the effective mass and moves the phase boundary, much as temperature controls the onset of magnetization; a quartic correction \Rfour{} changes the quartic stiffness and regulates post-onset amplitude; and higher-order terms shape behavior away from criticality. These signatures induce distinct regularization phenotypes, ranging from finite-threshold elimination and gradual attenuation to plateaus, non-monotonic tails, instability, or global collapse. They therefore organize invariant objectives by their regularization phenotypes rather than treating all regularizers as interchangeable penalties. In the canonical bilinear model, this correspondence additionally yields exact phase boundaries and equilibrium loadings.

\paragraph{Contributions.}
\textbf{(i) Ferromagnetic phase-transition perspective.}
We show that predictive-mode learning under invariant regularization admits a ferromagnetic Landau description, unifying mode onset, suppression, and selective retention with the language of order parameters, critical points, temperature-like controls, and spin-like collective modes.
\textbf{(ii) Objective signatures as a predictive methodology.}
We derive a Landau reduction in which effective quadratic, quartic, and higher-order terms form a low-order objective signature that predicts phase boundaries, post-onset loadings, and regularization phenotypes.
\textbf{(iii) Structural analysis of existing objectives.}
We apply this signature methodology to widely used invariant-learning objectives and obtain method-specific predictions, including finite-strength elimination, continuous shrinkage, nonlinear tails, instability, and collapse.
\textbf{(iv) Empirical validation.}
Controlled bilinear and ReLU experiments, including an MNIST-based construction, confirm the predicted phase boundaries, loading laws, and method-specific regularization phenotypes; coupled-feature experiments further validate the collective-mode extension.

\section{Related Work}

Landau theory describes ordering transitions through symmetry-constrained free-energy expansions. Its Ginzburg--Landau and relaxational extensions connect this framework to superconductivity, spatially varying order, and critical dynamics \citep{hohenberg1977dynamic,hohenberg2015ginzburg}. Related statistical-physics analyses of neural-network learning \citep{seung1992statistical,watkin1993statistical} have studied online learning in multilayer networks \citep{saad1995exact}, specialization transitions in committee machines \citep{biehl1998phase,aubin2018committee}, order-to-chaos transitions in random networks \citep{poole2016exponential}, and jamming near interpolation \citep{geiger2019jamming}.

Exact analyses of deep linear networks reveal distinctive learning dynamics, including plateaus and sequential mode learning \citep{saxe2014exact}, while complementary landscape analyses characterize their local-minimum structure and critical-point geometry \citep{laurent2018deep,achour2024loss}.

Invariant learning seeks predictors that remain stable across environments. This principle appears in invariant causal prediction \citep{peters2016causal}, \IRM and game-theoretic formulations of invariance \citep{arjovsky2019irm,ahuja2020irmgames}, GroupDRO \citep{sagawa2020groupdro}, REx/V-REx \citep{krueger2021rex}, Deep CORAL \citep{sun2016deepcoral}, and MMD alignment \citep{gretton2012kernel}. Existing theory has established limitations of finite-environment invariance and practical IRM objectives \citep{rosenfeld2021risks,kamath2021does}, while DomainBed highlights the sensitivity of domain-generalization results to hyperparameter and model-selection choices \citep{gulrajani2021domainbed}. WILDS further documents substantial performance degradation under naturally occurring distribution shifts \citep{koh2021wilds}.

\section{Formulation}

We consider supervised learning across training environments $e\in\mathcal E_{\mathrm{tr}}$ with joint distributions $P_e(X,Y)$. A stable feature remains predictive across environments, whereas a shortcut is predictive only under particular training conditions and may fail under distribution shift. We study equilibria of the learning objective, namely the stable states reached at the end of optimization. We focus on identifiable low-dimensional predictive modes, which isolate the structural effect of the learning objective from feature-discovery, optimization, and model-selection variability. This controlled setting makes mode onset, suppression, and retention directly measurable and enables quantitative, testable predictions of objective-induced behavior.

As in Landau theory, the form of the order parameter depends on the system. For a general finite network, we denote by $q$ the effective order parameter near a simple continuous transition; it represents the parameter combination that becomes soft at criticality. For closed-form analysis, we use the bilinear parameterization $\beta=w\theta$, in which $w$ provides a concrete realization of $q$; \cref{thm:app-network-soft-mode} gives the general construction.

Because the effective predictive coefficient $\beta=w\theta$ is unchanged under $(w,\theta)\mapsto(-w,-\theta)$, the objective obeys
$\F(-w,-\theta;\lambda)=\F(w,\theta;\lambda)$. This is the $\mathbb Z_2$ symmetry of a zero-field Landau free energy.
Assume that the $\theta$ direction remains noncritical at the reference critical point,
$\partial_{\theta\theta}\F(0,0;\lambda_c)>0$.
The stationary condition $\partial_\theta\F=0$ then uniquely determines, near the origin,
\begin{equation}
    \theta^\star(w;\lambda)
    =\kappa(\lambda)w+\mathcal O(w^3).
    \label{eq:linear-slaving}
\end{equation}
Thus $w$ is a valid local order-parameter coordinate, with $\theta$ locally slaved to it. Define
\begin{equation}
    \F_{\mathrm{eff}}(w;\lambda)
    =\F\!\left(w,\theta^\star(w;\lambda);\lambda\right)
    =L_0+\frac12r(\lambda)w^2
    +\frac14u(\lambda)w^4+\mathcal O(w^6).
\label{eq:general-reduced-landau}
\end{equation}
Consequently, $\F_{\mathrm{eff}}(-w;\lambda)=\F_{\mathrm{eff}}(w;\lambda)$, and its expansion contains only even powers.
As in Landau theory, the order parameter emerges continuously from zero near a continuous transition, so a low-order Taylor expansion captures the critical behavior; higher-order terms become important mainly away from criticality or at large regularization.
Here $r$ is the actual quadratic curvature along $w$ after the auxiliary parameter $\theta$ is allowed to readjust to its stationary value whenever $w$ changes:
\begin{equation}
    r(\lambda)
    =
    \left.
    \left(
    \F_{ww}-\frac{\F_{w\theta}^{\,2}}{\F_{\theta\theta}}
    \right)\right|_{(w,\theta)=(0,0)} .
    \label{eq:scalar-effective-mass}
\end{equation}
The first term is the direct curvature along $w$ at fixed $\theta$; the second accounts for the reduction in curvature produced by the readjustment of $\theta$. Meanwhile, $u(\lambda)$ collects fourth-order contributions after $\theta$ readjusts. Corrections of the forms $w^2$, $w\theta$, and $\theta^2$ in the original objective can all change the effective mass of the same critical mode. We call local operators that change $r$ and $u$ effective \Rtwo and \Rfour, respectively. Here $\lambda$ denotes the control strength of the corresponding method; when the two corrections are tuned independently, we write $\lambda_2$ and $\lambda_4$. The full reduction is given in Appendix~\ref{app:variational-reduction}. We call bilinear controls based on known environmental sensitivity oracle \Rtwo/\Rfour; their nonlinear counterparts are non-unique.

For closed-form, testable predictions, we specialize to the single-feature squared-loss model with explicit \Rtwo/\Rfour corrections. BCE has the same local form; a proof is given in Appendix~\ref{app:bce-local-equivalence}. The canonical objective is:
\begin{equation}
    \F_{\mathrm{can}}(w,\theta;\lambda_2,\lambda_4)
    =
    L_0-aw\theta
    +\frac{m+\lambda_4\gamma}{2}w^2\theta^2
    +\frac{\mu+\lambda_2\gamma}{2}w^2
    +\frac{\mu}{2}\theta^2 .
    \label{eq:local-learning-objective}
\end{equation}
Here $a$ is the task-predictive signal, $m$ is the local task-loss curvature with respect to the final loading $\beta$, $\gamma\geq0$ is the environmental sensitivity, and $\mu>0$ is the weight-decay coefficient. In this canonical case, $\lambda_2$ and $\lambda_4$ independently control the effective mass and quartic stiffness. Other algorithms need not take the form of \cref{eq:local-learning-objective}: joint quadratic and quartic corrections in $(w,\theta)$ respectively modify $r$ and $u$ after the same reduction. The critical-strength and loading formulas below are exact for this canonical model and asymptotic near the phase-transition critical point for other smooth objectives.

\section{A Landau Theory for a Single Feature}

\subsection{Variational Reduction}

Let $A=\mu+\lambda_2\gamma$ and $B=m+\lambda_4\gamma$, and assume $A>0$ and $B>0$. For fixed $w$, \cref{eq:local-learning-objective} is strictly convex in the auxiliary parameter $\theta$. The stationary condition $\partial_\theta\F_{\mathrm{can}}=0$ therefore uniquely determines the auxiliary equilibrium and the one-dimensional reduced objective:
\begin{equation}
    \theta^\star(w)=\frac{aw}{\mu+Bw^2},
    \qquad
    \F_{\mathrm{eff}}(w)
    \equiv
    \F_{\mathrm{can}}\!\left(w,\theta^\star(w)\right)
    =L_0+\frac{A}{2}w^2
    -\frac{a^2w^2}{2(\mu+Bw^2)}.
    \label{eq:reduced-objective}
\end{equation}
For this canonical model, strict convexity in $\theta$ makes the one-dimensional equilibrium reduction exact rather than merely local: stable equilibria of $\F_{\mathrm{eff}}$ uniquely recover those of the original two-dimensional objective.

\subsection{Landau Expansion and Equilibrium Phases}

\begin{figure}[t]
\centering
\begin{subfigure}[t]{0.325\linewidth}
\centering
\begin{tikzpicture}
\begin{axis}[
    width=\linewidth,height=3.40cm,
    axis lines=middle,
    xmin=-1.55,xmax=1.55,ymin=-0.34,ymax=0.95,
    xtick={-1,0,1},ytick=\empty,
    xlabel={$q$},ylabel={$\F_{\mathrm{eff}}(q)$},
    ylabel style={at={(axis cs:0,0.90)},anchor=south,rotate=0,font=\scriptsize},
    title={(a) $R_2$: finite-threshold elimination},
    title style={font=\scriptsize},label style={font=\scriptsize},tick label style={font=\scriptsize},
    legend style={font=\tiny,at={(axis description cs:-0.25,0.97)},anchor=north west,draw=none,fill=none},
    clip=true]
\addplot[very thick,blue!70!black,domain=-1.5:1.5,samples=180] {-0.5*x^2+0.25*x^4};
\addlegendentry{$r=-1$}
\addplot[very thick,orange!85!black,domain=-1.5:1.5,samples=180] {-0.125*x^2+0.25*x^4};
\addlegendentry{$r=-0.25$}
\addplot[very thick,red!70!black,domain=-1.5:1.5,samples=180] {0.175*x^2+0.25*x^4};
\addlegendentry{$r=0.35$}
\end{axis}
\end{tikzpicture}
\end{subfigure}
\hfill
\begin{subfigure}[t]{0.325\linewidth}
\centering
\begin{tikzpicture}
\begin{axis}[
    width=\linewidth,height=3.40cm,
    axis lines=middle,
    xmin=-1.20,xmax=1.20,ymin=-0.28,ymax=0.55,
    xtick={-1,0,1},ytick=\empty,
    xlabel={$q$},ylabel={$\F_{\mathrm{eff}}(q)$},
    ylabel style={at={(axis cs:0,0.50)},anchor=south,rotate=0,font=\scriptsize},
    title={(b) $R_4$: asymptotic shrinkage},
    title style={font=\scriptsize},label style={font=\scriptsize},tick label style={font=\scriptsize},
    legend style={font=\tiny,at={(axis description cs:-0.25,0.97)},anchor=north west,draw=none,fill=none},
    clip=true]
\addplot[very thick,blue!70!black,domain=-1.18:1.18,samples=220] {-0.5*x^2+0.25*x^4};
\addlegendentry{$u=1$}
\addplot[only marks,mark=*,mark size=1.25pt,blue!70!black,forget plot]
    coordinates {(-1,-0.25) (1,-0.25)};

\addplot[very thick,orange!85!black,domain=-1.18:1.18,samples=220] {-0.5*x^2+0.75*x^4};
\addlegendentry{$u=3$}
\addplot[only marks,mark=*,mark size=1.25pt,orange!85!black,forget plot]
    coordinates {(-0.57735,-0.08333) (0.57735,-0.08333)};

\addplot[very thick,red!70!black,domain=-1.18:1.18,samples=220] {-0.5*x^2+1.5*x^4};
\addlegendentry{$u=6$}
\addplot[only marks,mark=*,mark size=1.25pt,red!70!black,forget plot]
    coordinates {(-0.40825,-0.04167) (0.40825,-0.04167)};
\end{axis}
\end{tikzpicture}
\end{subfigure}
\hfill
\begin{subfigure}[t]{0.325\linewidth}
\centering
\begin{tikzpicture}
\begin{axis}[
    width=\linewidth,height=3.40cm,
    axis lines=middle,
    xmin=-1.55,xmax=1.55,ymin=-0.34,ymax=0.95,
    xtick={-1,0,1},ytick=\empty,
    xlabel={$q$},ylabel={$\F_{\mathrm{eff}}(q)$},
    ylabel style={at={(axis cs:0,0.90)},anchor=south,rotate=0,font=\scriptsize},
    title={(c) Stable-only reconstruction},
    title style={font=\scriptsize},label style={font=\scriptsize},tick label style={font=\scriptsize},
    clip=true]
\addplot[very thick,blue!70!black,domain=-1.5:1.5,samples=180] {-0.5*x^2+0.25*x^4};
\addplot[very thick,red!70!black,domain=-1.5:1.5,samples=180] {0.20*x^2+0.25*x^4};
\node[font=\scriptsize,blue!70!black] at (axis cs:1.03,-0.08) {stable};
\node[font=\scriptsize,red!70!black] at (axis cs:0.78,0.38) {shortcut};
\end{axis}
\end{tikzpicture}
\end{subfigure}
\caption{Schematic zero-field free-energy picture of finite-threshold \Rtwo elimination, asymptotic \Rfour shrinkage, and selective retention of the stable mode.}
\label{fig:physical-picture}
\end{figure}
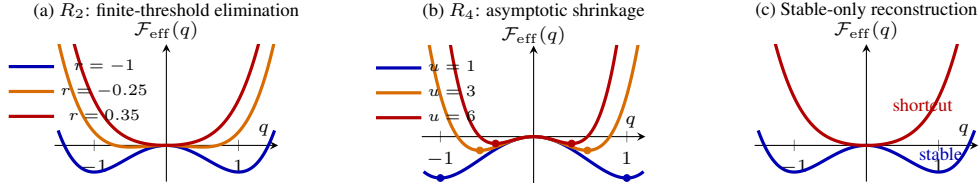

Expanding \cref{eq:reduced-objective} around $w=0$ gives
\begin{equation}
\begin{gathered}
    \F_{\mathrm{eff}}(w)
    =L_0+\frac{\gamma}{2}(\lambda_2-\lambda_{2,c})w^2
    +\frac{u}{4}w^4+\mathcal{O}(w^6),\\[-1mm]
    \lambda_{2,c}=\frac{a^2/\mu-\mu}{\gamma},
    \qquad
    u=\frac{2a^2B}{\mu^2}>0 .
\end{gathered}
\label{eq:landau-learning}
\end{equation}
The quadratic mass $r(\lambda_2)=\gamma(\lambda_2-\lambda_{2,c})$ determines the local stability of the origin, while the positive quartic stiffness $u$ stabilizes the nonzero ordered phase. We consider the ordinary case $u(\lambda_{2,c})>0$. \Cref{fig:physical-picture} illustrates the distinct effects of \Rtwo and \Rfour on the resulting wells.

The Landau free energy of a spatially uniform mean-field ferromagnet can be written as \citep{landau1937phase,landau1980statistical}
\begin{equation}
    F_{\mathrm{mag}}(M)
    =F_0+\frac{a_0}{2}(T-T_c)M^2+\frac{b}{4}M^4+O(M^6).
    \label{eq:landau-magnet}
\end{equation}
\Cref{eq:landau-learning,eq:landau-magnet} share the same local normal form, with the correspondence
\begin{equation}
    w\longleftrightarrow M,
    \qquad
    \lambda_2-\lambda_{2,c}\longleftrightarrow T-T_c,
    \qquad
    \gamma\longleftrightarrow a_0,
    \qquad
    u\longleftrightarrow b.
    \label{eq:landau-map}
\end{equation}
This exact mathematical matching formalizes the strict isomorphism proposed in the introduction. In this canonical control family, $\lambda_2$ is a temperature-like control parameter and $|w|$ is the order-parameter amplitude of the learning system: the origin is locally stable when $r>0$, a nonzero ordered solution appears when $r<0$, and $r=0$ gives the local continuous phase boundary.

In the continuous-time limit of gradient descent, the learning parameters obey
\begin{equation}
    \frac{dW_i}{dt}
    =-\eta\frac{\partial \F}{\partial W_i}
    \qquad\Longrightarrow\qquad
    \tau_c\frac{dq}{dt}
    =-r(\lambda)q-u(\lambda_c)q^3
    +\mathcal O\!\left(q^5+|\lambda-\lambda_c|q^3\right),
    \qquad \tau_c>0.
    \label{eq:main-landau-dynamics}
\end{equation}
Near a simple phase boundary, the reduced equation has the standard time-dependent Landau (Model-A) form for the order parameter $q$ \citep{hohenberg1977dynamic}. Its linear term changes sign at $r=0$, so the early-time growth-rate zero crossing used in \cref{fig:onset} identifies the static phase boundary. Appendix~\ref{app:dynamic-diagnostic} gives the center-manifold derivation and its scope. Beyond equilibrium loading, the objective's Landau signature predicts how learned modes respond to weak loss perturbations (Curie--Weiss law; \citealp{hohenberg2015ginzburg}) and how quickly they relax toward equilibrium during optimization. A conjugate field $-hq$ defines the susceptibility along an equilibrium branch:
\begin{equation}
    \chi_q
    \equiv\left.\frac{\partial q^\star}{\partial h}\right|_{h=0}
    =\left[\partial_q^2\F_{\mathrm{eff}}(q^\star)\right]^{-1},
    \qquad
    \tau_{\mathrm{relax}}\sim\tau_c\chi_q.
    \label{eq:main-susceptibility}
\end{equation}
For gradient descent with a small fixed step size $\eta$, $N_{90}$ denotes the steps required for a small perturbation along the slowest local mode to decay to $10\%$ of its initial amplitude. With $\kappa_{\min}>0$ the smallest Hessian eigenvalue at equilibrium, linearization predicts
\begin{equation}
    N_{90}^{\mathrm{pred}}
    =\frac{\log(0.1)}{\log(1-\eta\kappa_{\min})}
    \simeq\frac{\log 10}{\eta\kappa_{\min}}.
    \label{eq:main-recovery-steps}
\end{equation}
\subsection{Quantitative Phase Boundary and Loading}

\begin{theorem}[Task-signal--environment-sensitivity phase boundary]
Consider the canonical squared-loss model in \cref{eq:local-learning-objective}, and let $a=\E[XY]$. When $\gamma>0$, the feature-onset condition and critical regularization strength can be written jointly as
\begin{equation}
    |a|^2>\mu\bigl(\mu+\lambda_2\gamma\bigr)
    \quad\Longleftrightarrow\quad
    \lambda_2<\lambda_{2,c},
    \qquad
    \lambda_{2,c}=\frac{a^2/\mu-\mu}{\gamma}.
    \label{eq:onset-condition}
\end{equation}
If $\gamma=0$, quadratic environmental regularization does not move the local phase boundary of this feature, and the condition for a nonzero mode reduces to $|a|>\mu$.
\end{theorem}

\begin{proofsketch}
The exact reduced objective satisfies
\begin{equation}
    \frac{d\F_{\mathrm{eff}}}{dw}
    =
    w\left[
    A-\frac{a^2\mu}{(\mu+Bw^2)^2}
    \right].
    \label{eq:exact-reduced-derivative}
\end{equation}
The bracketed term is strictly increasing in $w^2$, from $A-a^2/\mu$ to $A>0$. Hence a stable nonzero stationary point exists if and only if $A-a^2/\mu<0$, which is \cref{eq:onset-condition}; equality gives the unique continuous phase boundary.
\end{proofsketch}

\begin{corollary}[Selective retention of stable features]
Let $\lambda_{2,c}^{(s)}$ and $\lambda_{2,c}^{(c)}$ denote the shortcut and stable-mode boundaries. If $\lambda_{2,c}^{(s)}<\lambda_{2,c}^{(c)}$, then for every $\lambda_2\in(\lambda_{2,c}^{(s)},\lambda_{2,c}^{(c)})$, the shortcut is stabilized at the origin while the stable direction remains in the nonzero acquired phase. Selective retention therefore depends on task signal relative to environmental sensitivity, not on marginal environment association alone.
\end{corollary}

\begin{proposition}[Steady-state feature loading]
Assume $B=m+\lambda_4\gamma>0$. The effective predictive loading at a stable stationary point of the canonical model is
\begin{equation}
    \left|\beta^\star\right|
    =\frac{\left[|a|-\sqrt{\mu(\mu+\lambda_2\gamma)}\right]_+}
    {m+\lambda_4\gamma},
    \label{eq:loading}
\end{equation}
where $[z]_+=\max\{z,0\}$. Thus, \Rtwo makes the loading vanish exactly at the finite boundary $\lambda_{2,c}$, whereas \Rfour continuously shrinks its nonzero post-onset loading.
\end{proposition}

\Cref{eq:loading} captures the distinct roles of the two corrections: increasing $\lambda_2$ reduces the numerator and eventually changes the sign of the quadratic mass, while increasing $\lambda_4$ only enlarges the denominator. The final loading therefore approaches zero linearly in $\lambda_{2,c}-\lambda_2$ near criticality; the square-root scaling of the order parameter is given in Appendix~\ref{app:equilibrium-results}.

\section{Multi-Component Order Parameters and Collective Modes}

A multicomponent order parameter is a natural extension of Landau theory: distinct components can form collective modes through mode coupling. In our setting, environmental responses of different features can reinforce or cancel one another in combination. Let $\bm{w},\bm{\theta}\in\mathbb{R}^p$ and $\bm{\beta}=\bm{w}\odot\bm{\theta}$. Here, $\Gamma_E$ denotes the environmental sensitivity matrix. The quadratic environmental correction can be written as
\begin{equation}
    R_2(\bm{w})
    =
    \frac{\lambda_2}{2}\sum_i\Gamma_{E,ii}w_i^2
    +
    \lambda_2\sum_{i<j}\Gamma_{E,ij}w_iw_j .
    \label{eq:spin-interactions}
\end{equation}
For $\Gamma_E\succeq0$, diagonal terms give componentwise masses, while off-diagonal terms provide spin-like pairwise coupling; task-supported low-eigenvalue directions form favored collective modes. The corresponding matrix \Rfour{} correction has the same matrix structure but acts on the final loading $\bm{\beta}$. Below, $C=\diag(\E[X_1Y],\ldots,\E[X_pY])$ denotes the task drive of each component.

\begin{theorem}[Collective-mode phase boundary]
For the multifeature bilinear extension of Eq.~(4), assume $\Gamma_E\succeq0$. The origin is locally unstable along a collective mode iff
\begin{equation}
    \lambda_{\max}\!\left[
    \frac{1}{\mu}C^{\top}
    (\mu I+\lambda_2\Gamma_E)^{-1}C
    \right]>1.
    \label{eq:matrix-onset}
\end{equation}
\end{theorem}

\begin{proposition}[Predictor-space selectivity of matrix \Rfour{}]
\label{prop:predictor-selectivity}
Let $\Gamma_E\succeq0$, $\lambda_2,\lambda_4>0$, and $\bm\beta=\bm w\odot\bm\theta$. Define
\begin{equation}
R_2(\bm w)=\frac{\lambda_2}{2}\bm w^\top\Gamma_E\bm w,
\qquad
R_4(\bm\beta)=\frac{\lambda_4}{2}\bm\beta^\top\Gamma_E\bm\beta.
\label{eq:matrix-r2-r4}
\end{equation}
Then $R_4(\bm\beta)=0$ iff $\bm\beta\in\ker\Gamma_E$, whereas $R_2(\bm w)=0$ iff $\bm w\in\ker\Gamma_E$. The latter need not imply $\bm\beta\in\ker\Gamma_E$, since coordinatewise multiplication by $\bm\theta$ need not preserve this nullspace. Thus matrix \Rfour directly penalizes environmental sensitivity in predictor space.
\end{proposition}

Environmental sensitivity specifies the penalized directions, while the penalty's local dependence on model parameters determines its signature order. The derivations and the Fig.~7 specialization are given in Appendix~\ref{app:collective-proofs}. The oracle operators below use the $b_F$-shift setup for controlled spectral-mechanism checks, but the formulation and criterion apply beyond this particular shift. Crucially, oracle $\Gamma_E$ is not uniquely identifiable from observational associations in finitely many training environments alone; uniquely recovering it requires additional structural assumptions or causal priors.

\section{Regularization-Path Predictions}

\begin{table}[H]
\centering
\small
\setlength{\tabcolsep}{4.0pt}
\renewcommand{\arraystretch}{1.00}
\setlength{\aboverulesep}{0.8pt}
\setlength{\belowrulesep}{0.8pt}
\caption{Predicted regularization phenotypes.}
\label{tab:regularization-phenotypes}
\begin{tabular}{>{\raggedright\arraybackslash}p{0.280\textwidth}>{\raggedright\arraybackslash}p{0.245\textwidth}>{\raggedright\arraybackslash}p{0.414\textwidth}}
\toprule
\textbf{Methods} & \textbf{Objective signature} & \textbf{Predicted regularization phenotype} \\
\midrule
\mbox{MMD-mean}, \mbox{CORAL-full}, IGA~\citep{koyama2020maximal}, \mbox{Gaussian MMD}
&
Stable positive \Rtwo{} component
&
Moves the local onset and enables finite-strength mode elimination. Higher-order terms in IGA and Gaussian MMD may modify the path shape.
\\
\midrule[0.3pt]
\mbox{CORAL-cov}, Fishr~\citep{rame2022fishr}, \mbox{V-REx}
&
Stable positive \Rfour{} component
&
Continuously shrinks an acquired mode. V-REx may exhibit a small nonlinear tail due to weak \Rsix/\Reight corrections.
\\
\midrule[0.3pt]
IRM
&
\Rfour{}-leading with negative \Rsix{} and positive \Reight{} terms
&
May exhibit non-monotonic paths and sharp drops at large strengths.
\\
\midrule[0.3pt]
GroupDRO
&
Reweighted \Rtwo{}/\Rfour{} terms
&
Suppresses shortcuts at small strengths but may retain a nonzero plateau.
\\
\midrule[0.3pt]
DANN~\citep{ganin2016dann}
&
State-dependent \Rtwo/\Rfour{} terms
&
May exhibit unstable paths and large seed variability.
\\
\midrule[0.3pt]
Fish~\citep{shi2022fish},\newline
MLDG~\citep{li2018mldg}
&
Indefinite \Rtwo{} and \Rfour{} gradient-alignment terms
&
Shows direction-dependent \Rtwo-like suppression at small strength and becomes unstable at large strength.
\\
\midrule[0.3pt]
ERM, Mixup~\citep{zhang2018mixup}
&
No fixed environmental correction
&
No sign-controlled shortcut suppression.
\\
\bottomrule
\end{tabular}
\end{table}

\clearpage
\begin{figure}[H]
\centering
\includegraphics[width=0.96\textwidth]{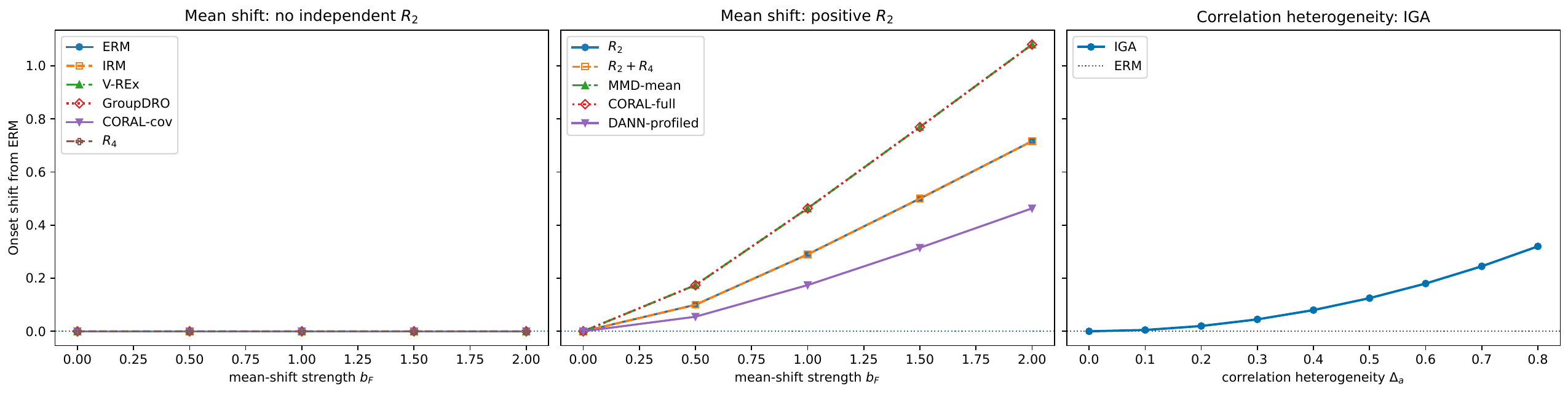}
\caption{Local taxonomy of onset shifts. The vertical axis shows the shift relative to \ERM{} in the critical signal strength $a_s$ at which the model begins to learn the shortcut. The first two panels use the mean-shift probe $b_F$; the third uses IGA's matched correlation-heterogeneity probe $\Delta_a$. In all cases, the analytic Landau onset agrees closely with the measurement, with maximum absolute deviation $4.7\times10^{-6}$. DANN-profiled is the locally profiled discriminator response (Appendix~\ref{app:dann-profile}).}
\label{fig:onset}
\end{figure}

Table~\ref{tab:regularization-phenotypes} summarizes the leading-order predictions in the low-loading regime. Their local expansions and empirical definitions are detailed in Appendices~\ref{app:fingerprint-derivations} and~\ref{app:empirical-objectives}, respectively. Actual selective retention also depends on whether the low-order matrices align with the oracle $\Gamma_E$: aligned matrix \Rtwo or \Rfour operators can preserve low-sensitivity stable modes, whereas misaligned operators gradually suppress them as regularization grows. Away from onset, higher-order or sign-indefinite corrections can dominate and produce nonlinear tails, instability, or collapse.

\section{Experiments}

The preceding analysis yields quantitative predictions for phase boundaries, steady-state loadings, and critical strengths (\cref{eq:onset-condition,eq:loading}), together with method-specific regularization phenotypes derived from the local objective signatures (\cref{tab:regularization-phenotypes}). We therefore design the experiments to test these conclusions in sequence: \cref{fig:onset,fig:boundaries,fig:critical-calibration} evaluate onset shifts, loading laws, and critical strengths; \cref{fig:paths,fig:nonlinear-paths} examine whether the predicted path distinctions persist along full bilinear and nonlinear regularization paths; and \cref{fig:collective-mode} tests the spectral prediction that stable structure may reside in coupled collective modes (\cref{eq:matrix-onset}).

\subsection{Quadratic Mass Moves the Onset; Quartic Stiffness Shrinks the Amplitude}

Figures~2--4 use scalar single-feature constructions. \Cref{fig:onset} uses matched environmental probes to measure the onset shift relative to \ERM. In this figure, onset is defined by the static condition $r=0$ in \cref{eq:onset-condition}; its numerical measurement is the zero crossing of the early-time local growth rate from a small perturbation, with their local equivalence detailed in Appendix~\ref{app:dynamic-diagnostic}. Its first two panels scan the mean shift $b_F$: methods without an independent quadratic correction show almost no onset shift, whereas a positive gate-block mass produces a finite shift. The third panel instead scans the correlation heterogeneity $\Delta_a$ for IGA, whose joint quadratic stiffness is non-degenerate only in that construction. These results agree with \cref{eq:onset-condition}: \Rtwo determines whether a mode can grow from the origin, while \Rfour mainly controls the amplitude after growth.

\begin{figure}[H]
\centering
\begin{minipage}[t]{0.46\textwidth}
\centering
\includegraphics[width=\linewidth]{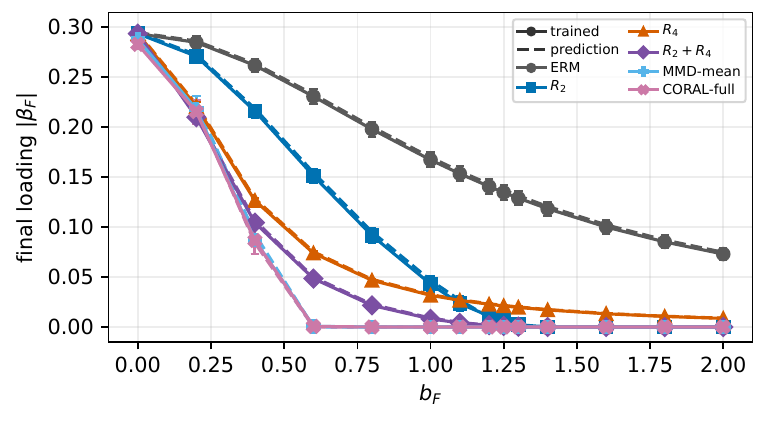}\\[-1mm]
{\scriptsize (a) Mean-shift probe ($b_F$)}
\end{minipage}\hspace{0.006\textwidth}
\begin{minipage}[t]{0.46\textwidth}
\centering
\includegraphics[width=\linewidth]{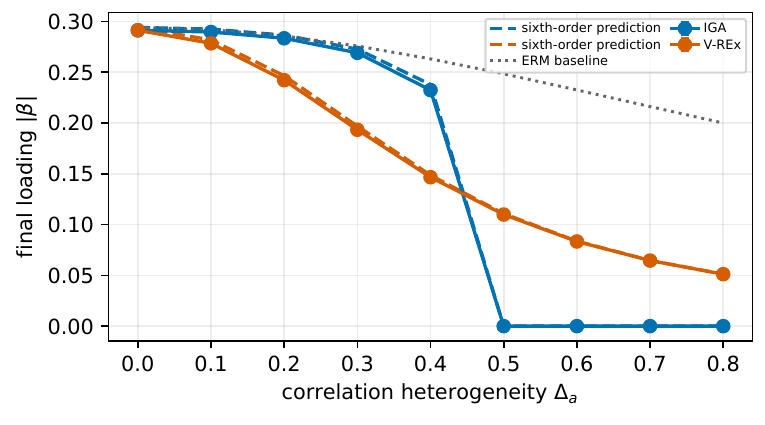}\\[-1mm]
{\scriptsize (b) Matched IGA--V-REx paths}
\end{minipage}
\caption{Quantitative validation of shortcut-loading predictions. Dashed curves show reduced Landau predictions; solid curves and error bars show the mean and s.e.m. Panel~(a) uses the mean-shift probe $b_F$; panel~(b) uses the matched correlation-heterogeneity probe $\Delta_a$.}
\label{fig:boundaries}
\end{figure}

\Cref{fig:boundaries} quantitatively tests the reduced Landau loading predictions under two matched environmental probes: the mean shift $b_F$ and correlation heterogeneity $\Delta_a$. Without fitting to the trained paths, objective-specific low-order coefficients derived from each learning objective are inserted into the same reduced Landau equilibrium, yielding accurate loading predictions for multiple methods in panel~(a); retaining the sextic coefficient extends the same construction to V-REx and IGA in panel~(b) (Appendix~\ref{app:equilibrium-results}).

\clearpage

\begin{figure}[H]
\centering
\includegraphics[width=0.88\textwidth]{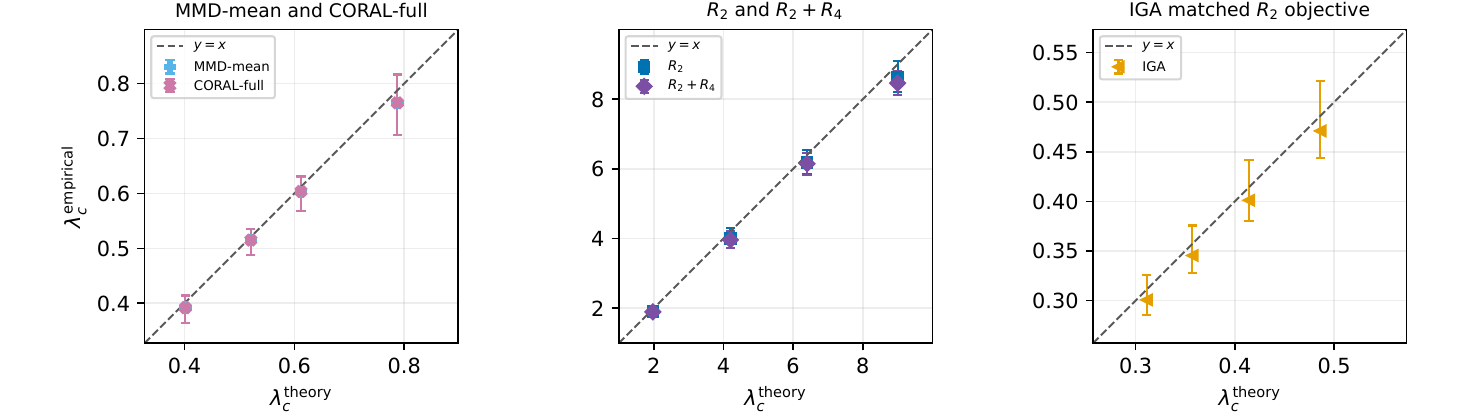}
\caption{Critical-strength calibration from final shortcut loading. Each marker denotes one setting; vertical bars span the central 95\% interval of the seed-resampled critical-strength estimates, and the dashed line is $y=x$. \Rtwo{}+\Rfour shares the \Rtwo onset, while MMD-mean, CORAL-full, and matched IGA track their corresponding effective-\Rtwo transition points.}
\label{fig:critical-calibration}
\end{figure}

\Cref{fig:critical-calibration} directly tests the predicted \Rtwo phase-transition point across explicit \Rtwo, \Rtwo{}+\Rfour, MMD-mean, CORAL-full, and matched IGA objectives. Near a simple continuous boundary, the equilibrium loading vanishes linearly with distance to the critical strength (\cref{eq:loading}), motivating a local linear extrapolation from the ordered side. The \Rtwo/\Rtwo{}+\Rfour comparison further confirms the distinct effects of \Rtwo and \Rfour: adding \Rfour changes the amplitude and near-critical slope of the loading branch without shifting the \Rtwo-controlled critical point. We estimate the empirical critical strength from a theory-blind coarse sweep followed by a dense rescan; the theoretical value is not used to select the fitting window or perform the fit. Only well-resolved local fits passing pre-specified quality controls are retained; detailed calibration settings and criteria are provided in Appendix~\ref{app:critical-calibration-protocol}.

\subsection{Regularization Paths}

\begin{figure}[H]
\centering
\begin{minipage}[t]{0.49\textwidth}
\centering
\includegraphics[width=\linewidth]{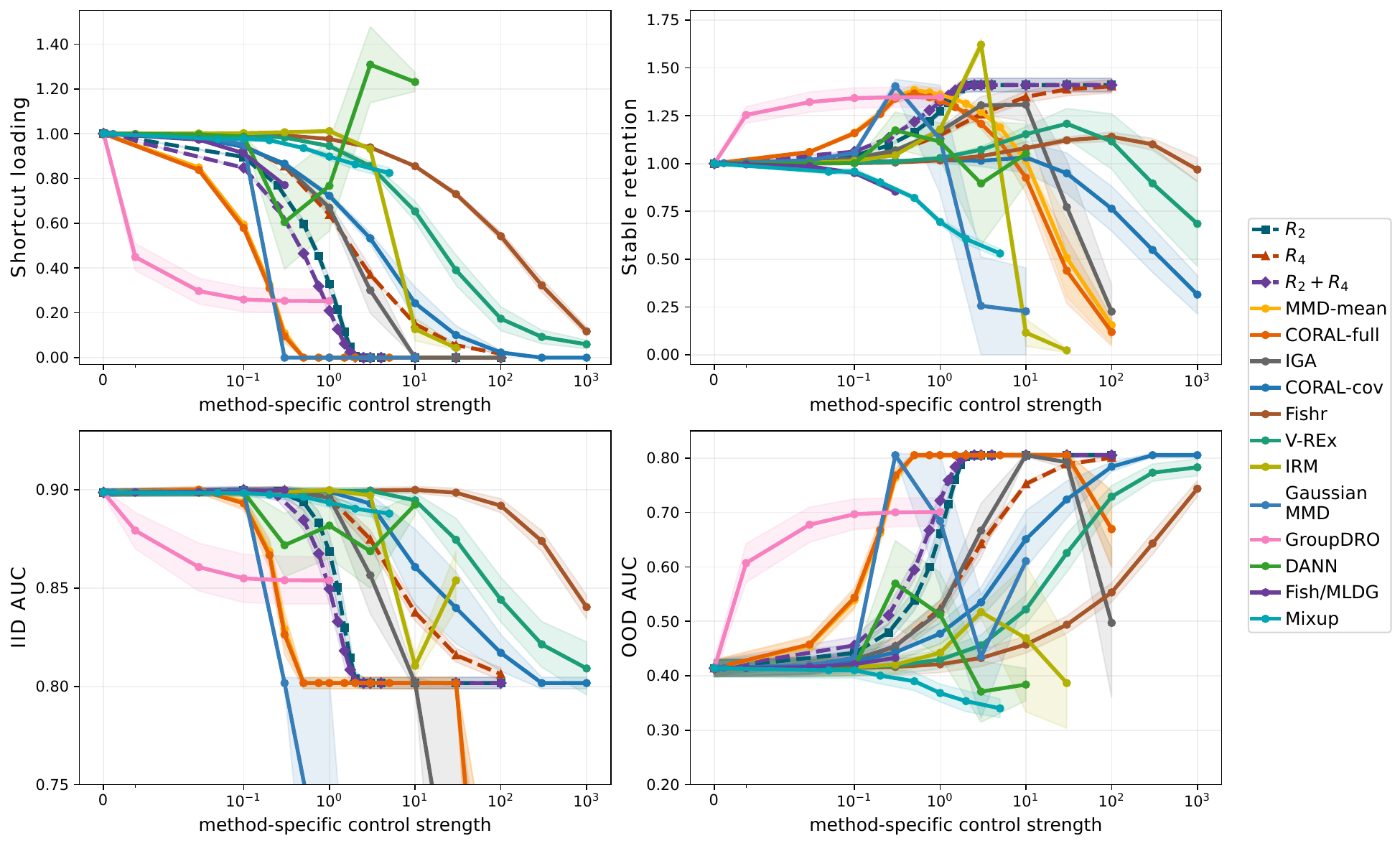}\\[-1mm]
{\scriptsize (a) 256 training samples per environment}
\end{minipage}\hspace{0.006\textwidth}
\begin{minipage}[t]{0.49\textwidth}
\centering
\includegraphics[width=\linewidth]{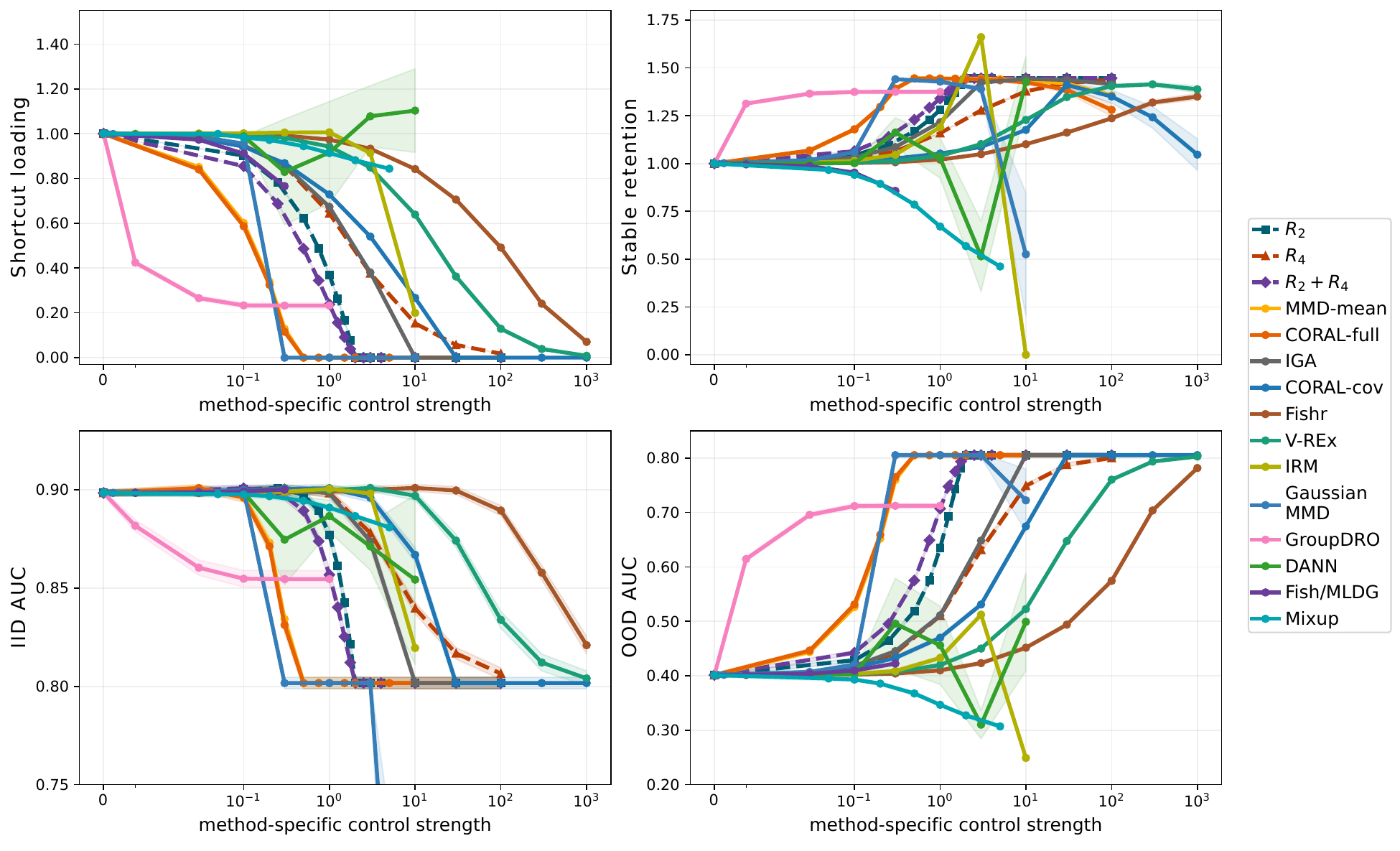}\\[-1mm]
{\scriptsize (b) 8192 training samples per environment}
\end{minipage}
\caption{Sample-size dependence of bilinear regularization paths. Curves and shaded bands show means and standard errors over five seeds, with shared IID/OOD test sets across sample sizes. Dashed curves with filled markers denote explicit oracle $R_2$, $R_4$, and $R_2+R_4$ controls. Shortcut loading and stable retention measure mean absolute input derivatives, normalized at $\lambda=0$.}
\label{fig:paths}
\end{figure}

Figures~5--6 use separate stable and shortcut features together with independent noise coordinates. In \cref{fig:paths}, stable features suffer earlier collateral damage with 256 samples per environment (left) than with 8192 (right), consistent with larger finite-sample fluctuations in the environmental statistics (Appendix~\ref{app:figure5-operator-alignment}). Oracle \Rtwo and \Rtwo{}+\Rfour controls preserve the stable mode after shortcut suppression throughout the tested range. \OOD AUC improves only when shortcut suppression preserves the stable mode. The observed regularization phenotypes agree with the qualitative predictions derived from the objective signatures in \cref{tab:regularization-phenotypes}. MMD-mean and CORAL-full show \Rtwo-like finite-strength suppression, whereas CORAL-cov, Fishr, and V-REx show \Rfour-leading gradual shrinkage. IGA and Gaussian MMD show modified path shapes, IRM non-monotonic paths, and V-REx a mild nonlinear tail, consistent with their higher-order structure. GroupDRO retains a nonzero shortcut plateau, while DANN shows unstable, seed-sensitive paths.

\clearpage

\begin{figure}[H]
\centering
\begin{minipage}[t]{0.49\textwidth}
\centering
\includegraphics[width=\linewidth]{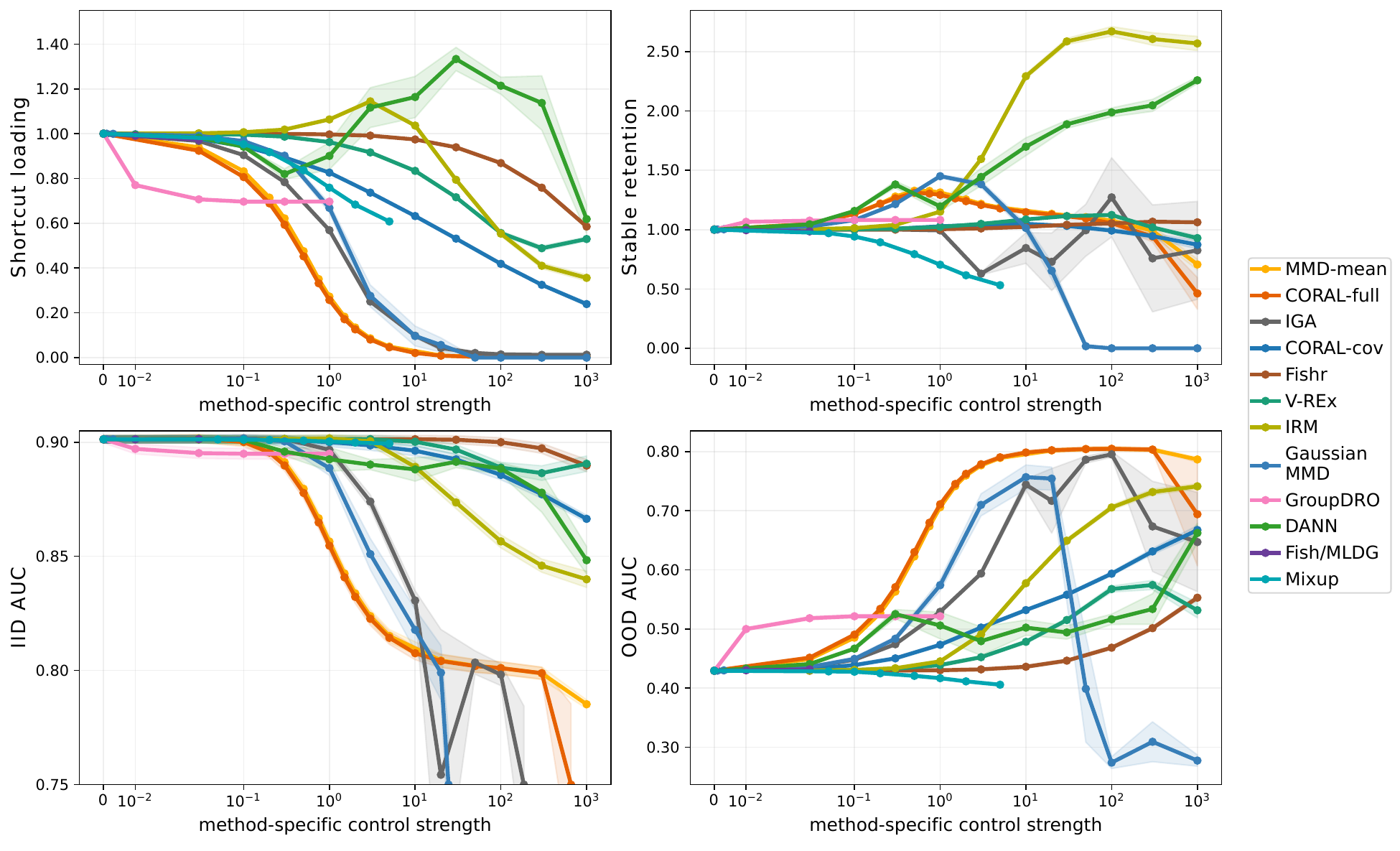}\\[-1mm]
{\scriptsize (a) One-hidden-layer ReLU}
\end{minipage}\hspace{0.006\textwidth}
\begin{minipage}[t]{0.49\textwidth}
\centering
\includegraphics[width=\linewidth]{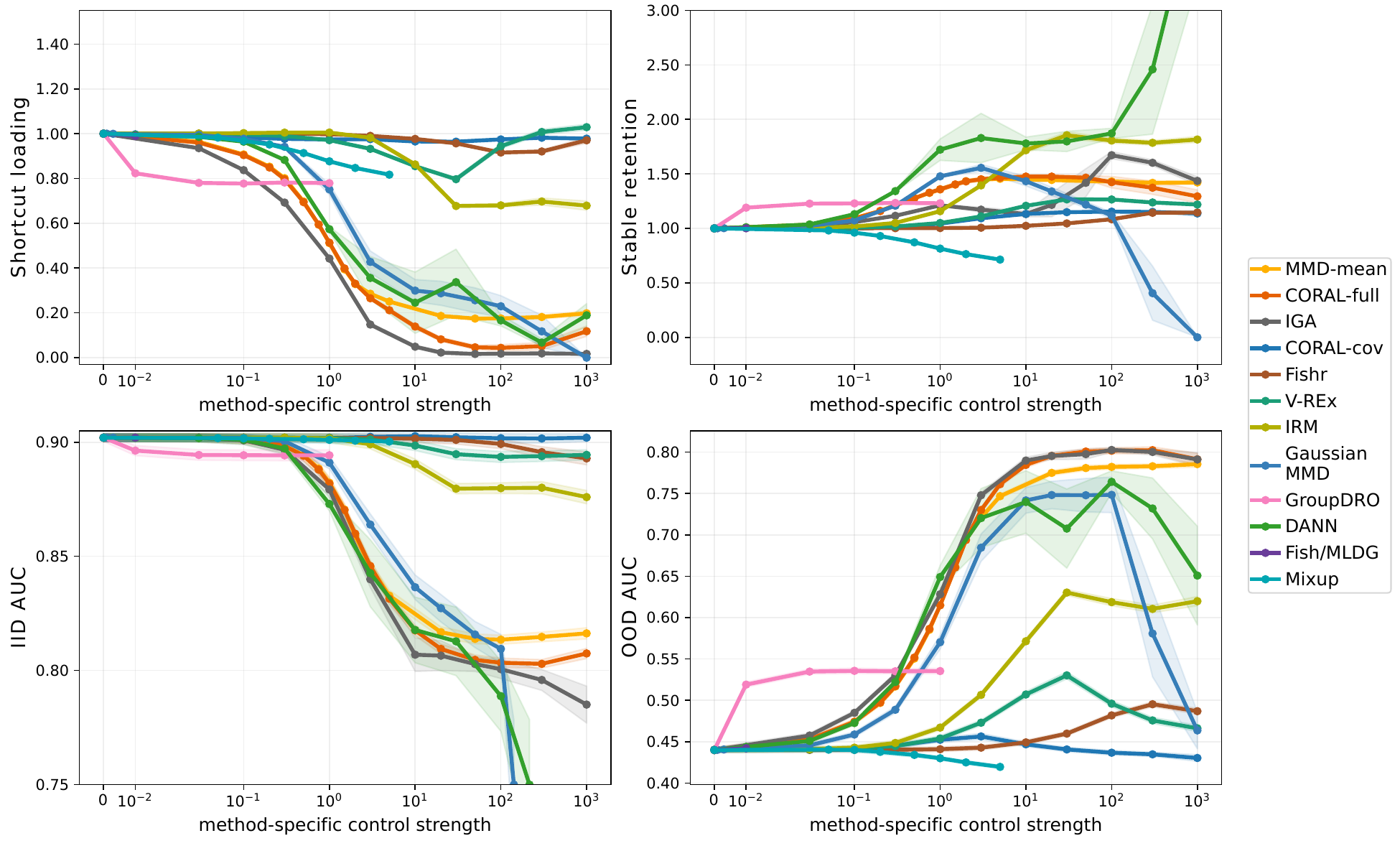}\\[-1mm]
{\scriptsize (b) Two-hidden-layer ReLU}
\end{minipage}
\caption{Nonlinear regularization paths using the same data as \cref{fig:paths}(b), with 8192 training samples per environment. The bilinear model's qualitative regularization phenotypes persist in both ReLU networks, with different transition scales. Objectives with higher-order terms exhibit network-dependent nonlinear tails.}
\label{fig:nonlinear-paths}
\end{figure}

\looseness=-1
We transfer the same datasets and objectives to ReLU networks. \Cref{fig:nonlinear-paths} shows that low-order signatures remain qualitatively predictive: \Rtwo{}-like objectives retain finite-strength shortcut control, whereas \Rfour{}-like objectives primarily yield continuous suppression with residual finite-strength shortcut loading. Across the tested depths and sample sizes, these qualitative signature--phenotype relations persist, while transition scales, large-strength behavior, and higher-order effects vary. Deeper networks and extended paths are reported in Appendix~\ref{app:extended-nonlinear-paths}. To study regularization paths with more complex, image-based stable features, we extend the experiments to MNIST (Appendix~\ref{app:mnist-paths}).

\subsection{Collective Feature Modes}

\begin{figure}[!t]
\centering
\includegraphics[width=0.98\textwidth]{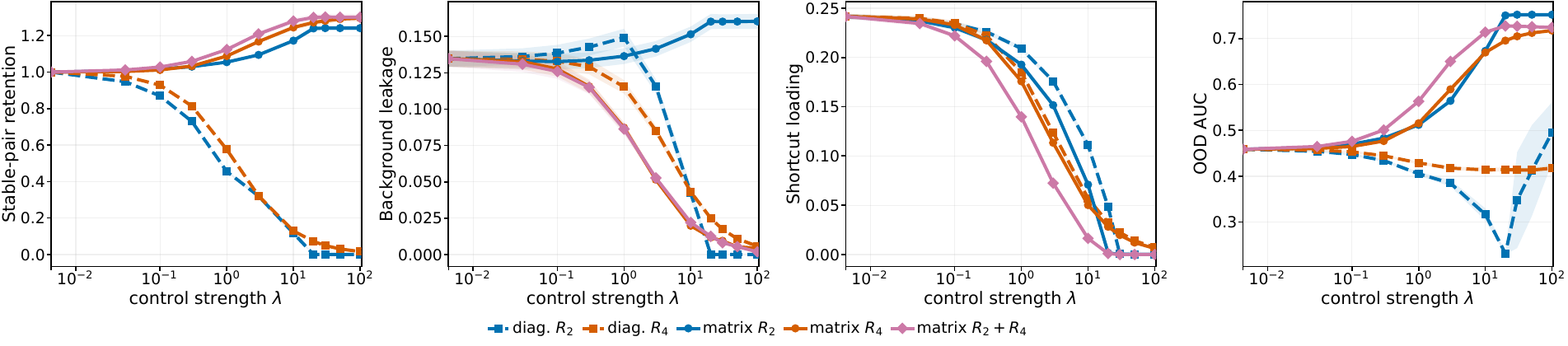}
\caption{Regularization paths for a compensatory collective mode. The panels report stable-pair retention, task-irrelevant background leakage, shortcut loading, and OOD AUC. Dashed curves use $\diag(\Gamma_E)$, while solid curves use the full oracle $\Gamma_E$. The matrix \Rfour and matrix \Rtwo{}+\Rfour curves nearly overlap in the background-leakage panel.}
\label{fig:collective-mode}
\end{figure}

\Cref{fig:collective-mode} uses a three-feature construction in which two individually environment-sensitive features form a compensatory collective mode, alongside an independent shortcut. \Cref{fig:collective-mode} compares coordinate-diagonal $R_2,R_4$ with coupled matrix $R_2,R_4,R_2+R_4$. Diagonal regularization destroys the stable combination while suppressing the shortcut. In contrast, coupled matrix \Rtwo, \Rfour, and their combination agree with the spectral theory: they suppress the shortcut while preserving the collective mode and improving OOD AUC. Matrix \Rfour{} is particularly effective at suppressing background leakage from imperfect pair cancellation. The supplementary full baseline comparison (\cref{fig:app-collective-all}) indicates that, in this construction, the non-oracle baselines do not consistently exhibit the same joint preservation--suppression behavior across all four diagnostics.

\begin{figure}[!t]
\centering
\includegraphics[width=\textwidth]{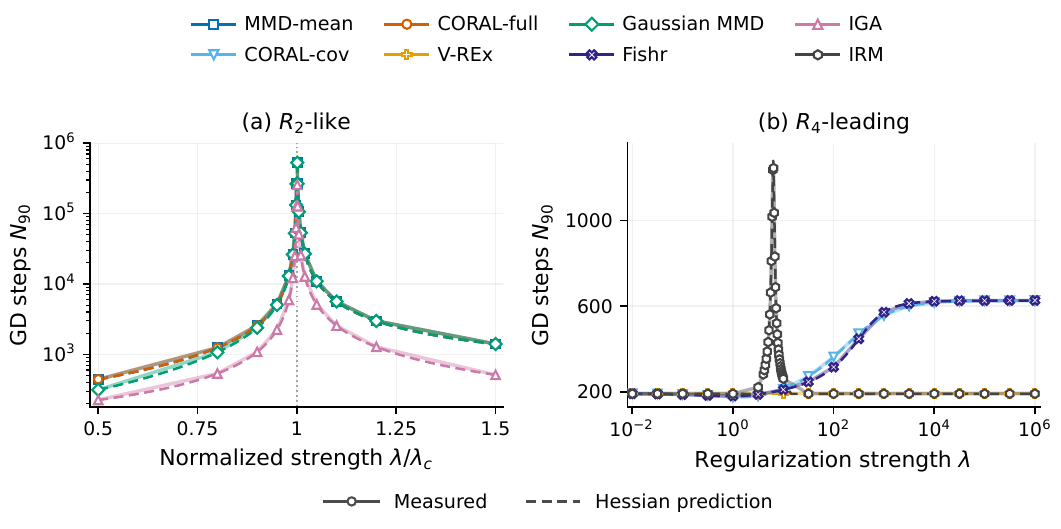}
\caption{Optimization dynamics in the bilinear model under \Rtwo-like (left) and \Rfour-leading (right) objectives. Solid curves with markers show GD steps $N_{90}$ for a small perturbation along the slowest local mode to decay to $10\%$ of its initial amplitude; dashed curves use the full objective Hessian at equilibrium. Left strengths are normalized by $\lambda_c$. All objectives share the same population environments and learning rate.}
\label{fig:objective-dynamics}
\end{figure}

\begin{samepage}
\subsection{Objective Signatures and Optimization Dynamics}
\label{sec:objective-dynamics}

Consistent with their signatures, \Rtwo-like objectives exhibit critical slowing near their suppression boundaries, whereas \Rfour-leading paths retain finite recovery times in the bilinear scan (\cref{fig:objective-dynamics}). IRM's higher-order terms produce a finite peak. Predictions from \cref{eq:main-recovery-steps} closely match measured recovery steps with fixed GD step size $\eta=0.03$ (Appendix~\ref{app:objective-dynamics-protocol}). Appendix~\ref{app:soft-mode-response} tests the $|r|^{-1}$ susceptibility and relaxation scaling near criticality; Appendix~\ref{app:objective-relaxation} provides additional recovery curves.

\end{samepage}

\section{Conclusion}

We develop a Landau-theoretic framework for predicting how invariant-learning objectives control the onset, suppression, and retention of predictive modes from their low-order structure. The resulting objective signatures organize existing methods by distinct regularization phenotypes and extend naturally from individual features to collective modes. Future work can build on the framework's equilibrium and dynamical descriptions to explain architecture-dependent algorithmic behavior and predict relevant regularization-strength ranges before empirical search. These insights can further guide the design of objectives with desired phenotypes, separating the estimation of environmental sensitivity $\Gamma_E$ from the construction of penalties with prescribed Landau signatures.

\label{main-text-end}

\subsection*{Reproducibility statement}
Complete proofs are provided in the appendices. Experimental configurations,
objective definitions, random seeds, and critical-strength estimation protocols
are documented in Appendix~\ref{app:reproducibility}.
Code is included in the supplementary material.

\subsection*{AI use statement}
Generative AI tools were used to aid and polish the writing of this paper, including improvements to clarity and readability. All AI-assisted content was reviewed by the authors, who take full responsibility for the final manuscript.

\bibliography{references}
\bibliographystyle{iclr2027_conference}

\clearpage
\appendix
\begin{center}
\Large\bfseries Appendix
\end{center}
\addtocontents{toc}{\protect\setcounter{tocdepth}{2}}
\begingroup
\edef\savedtocdepth{\number\value{tocdepth}}
\renewcommand{\contentsname}{Appendix Contents}
\hypersetup{linkcolor=black,linktoc=all}
\pdfbookmark[1]{Appendix Contents}{appendix-contents}
\tableofcontents
\setcounter{tocdepth}{\savedtocdepth}
\endgroup
\par\medskip

\section{Scope and notation}
\label{app:notation}

This appendix provides the complete derivations underlying the equilibrium Landau theory in the main text. Two levels of claims should be distinguished. For the canonical bilinear model with squared loss and explicit $R_2/R_4$ regularization, the reduced objective, critical strengths, and loading formula are exact. For general smooth losses and other algorithms, we compare only the leading local operators that emerge near a common reference point. The sign change of the quadratic mass at the origin, and hence the continuous local transition point, is determined exactly by the local curvature; nonzero equilibrium loading is predicted only asymptotically near criticality.

In environment $e$ with data distribution $P_e$, define
\begin{equation}
    c_e=\E_e[XY],\qquad q_e=\E_e[X^2],\qquad
    \beta=w\theta .
    \label{eq:app-moments}
\end{equation}
The environmental squared risk can be written exactly as
\begin{equation}
    \ell_e(w,\theta)
    =L_{0,e}-c_e\beta+\frac{q_e}{2}\beta^2 .
    \label{eq:app-env-risk}
\end{equation}
Hereafter, an overbar denotes an average over environments, e.g., $\bar c=\overline{c_e}$ and $\bar q=\overline{q_e}$, and we define $\Delta c_e=c_e-\bar c$ and $\Delta q_e=q_e-\bar q$. The reference objective with weight decay on both parameter blocks is
\begin{equation}
 \F_0(w,\theta)=\bar L_0-\bar c\,w\theta
 +\frac{\bar q}{2}w^2\theta^2
 +\frac{\mu_g}{2}w^2+\frac{\mu_\theta}{2}\theta^2 .
 \label{eq:app-reference-objective}
\end{equation}
We allow separate weight-decay coefficients $\mu_g$ and $\mu_\theta$; the main text sets $\mu_g=\mu_\theta=\mu$.

For any unit direction $v=(v_w,v_\theta)$, we define the coefficients of the objective along the ray $z=\rho v$ by
\begin{equation}
 \F(\rho v)-\F(0)=\frac12 a_2(v)\rho^2+\frac14 a_4(v)\rho^4+O(\rho^6).
 \label{eq:app-fingerprint-definition}
\end{equation}
\Cref{eq:app-reference-objective} gives the reference fingerprint
\begin{equation}
 \begin{aligned}
 a^{(0)}_2(v)&=\mu_gv_w^2+\mu_\theta v_\theta^2-2\bar c\,v_wv_\theta,\\
 a^{(0)}_4(v)&=2\bar q\,(v_wv_\theta)^2.
 \end{aligned}
 \label{eq:app-reference-fingerprint}
\end{equation}
Accordingly, a ``joint \Rtwo{} correction'' in this paper means an independent correction that changes $a_2$ or the Hessian at the origin, whereas a ``joint \Rfour{} correction'' begins at order $\rho^4$ in the joint variables $(w,\theta)$. To determine mode onset, these joint corrections must further be projected onto the critical soft mode to obtain the change in the effective mass of the reduced objective. The sign and magnitude of this change depend on the projection of the joint quadratic operator onto that mode.

\section{Variational reduction}
\label{app:variational-reduction}

This section states the precise conditions under which noncritical parameter directions can be eliminated to obtain a low-dimensional effective objective. We first give the two-coordinate reduction used by the canonical model and then extend it to the critical soft mode of a finite nonlinear network. The reduction characterizes equilibrium stationary points.

\begin{theorem}[Local elimination of the auxiliary degree of freedom]
\label{thm:app-ift-reduction}
Suppose that $\F(w,\theta;\lambda)$ is $C^6$ in a neighborhood of the critical reference point
$(w,\theta,\lambda)=(0,0,\lambda_c)$ and satisfies
\begin{equation}
    \partial_\theta\F(0,0;\lambda_c)=0,
    \qquad
    \partial_{\theta\theta}\F(0,0;\lambda_c)>0.
    \label{eq:app-ift-assumptions}
\end{equation}
Then there exists a neighborhood of this point and a unique $C^5$ function
$\theta^\star(w,\lambda)$ such that
\begin{equation}
    \partial_\theta
    \F\!\left(w,\theta^\star(w,\lambda);\lambda\right)=0.
    \label{eq:app-theta-stationary}
\end{equation}
Define
\begin{equation}
    \F_{\mathrm{eff}}(w;\lambda)
    =
    \F\!\left(w,\theta^\star(w,\lambda);\lambda\right).
    \label{eq:app-effective-definition}
\end{equation}
Then
\begin{equation}
    \frac{d\F_{\mathrm{eff}}}{dw}
    =
    \partial_w\F\!\left(w,\theta^\star(w,\lambda);\lambda\right).
    \label{eq:app-envelope}
\end{equation}
Consequently, stationary points of $\F_{\mathrm{eff}}$ are in one-to-one correspondence with local stationary points of the original objective satisfying
$\partial_\theta\F=0$.
\end{theorem}

\begin{proof}
By \cref{eq:app-ift-assumptions}, the function
$G(w,\theta,\lambda)=\partial_\theta\F(w,\theta;\lambda)$
satisfies, at the reference point,
$G=0$ and $\partial_\theta G=\partial_{\theta\theta}\F\neq0$.
The implicit function theorem therefore immediately yields a unique
$\theta^\star(w,\lambda)$. Applying the chain rule to
\cref{eq:app-effective-definition} gives
\begin{equation}
    \frac{d\F_{\mathrm{eff}}}{dw}
    =
    \partial_w\F
    +
    \partial_\theta\F\,
    \frac{\partial\theta^\star}{\partial w}
    =
    \partial_w\F,
\end{equation}
where the second term vanishes by \cref{eq:app-theta-stationary}.
\end{proof}

\begin{theorem}[Soft-mode reduction and generic phase boundary for a finite nonlinear network]
\label{thm:app-network-soft-mode}
Let $\F(\Theta;\lambda)$ be $C^7$ on a finite-dimensional parameter space, and let
$\bar\Theta(\lambda)$ be a $C^6$ branch of stationary points near $\lambda_c$:
\begin{equation}
    \nabla_\Theta\F\!\left(\bar\Theta(\lambda);\lambda\right)=0.
    \label{eq:app-network-stationary-branch}
\end{equation}
Suppose that, after restricting to a local slice transverse to any exact parameter symmetries, the critical Hessian
\begin{equation}
    H_c=\nabla_\Theta^2\F\!\left(\bar\Theta(\lambda_c);\lambda_c\right)
    \label{eq:app-network-critical-hessian}
\end{equation}
has a one-dimensional kernel $\ker H_c=\operatorname{span}\{v_c\}$, where $\|v_c\|=1$, and is positive definite on $v_c^\perp$. Then there is a neighborhood of $(q,\lambda)=(0,\lambda_c)$ and a unique $C^6$ map
$\eta^\star(q,\lambda)\in v_c^\perp$, with $\eta^\star(0,\lambda)=0$, such that
\begin{equation}
    P_\perp\nabla_\Theta\F\!\left(
    \bar\Theta(\lambda)+qv_c+\eta^\star(q,\lambda);\lambda
    \right)=0,
    \label{eq:app-network-transverse-stationarity}
\end{equation}
where $P_\perp$ denotes orthogonal projection onto $v_c^\perp$. Hence all nearby stationary points lie on the scalar soft-mode manifold
\begin{equation}
    \Theta^\star(q,\lambda)
    =\bar\Theta(\lambda)+qv_c+\eta^\star(q,\lambda).
    \label{eq:app-network-soft-parameterization}
\end{equation}
Defining
\begin{equation}
    \F_{\mathrm{eff}}(q;\lambda)
    =\F\!\left(\Theta^\star(q,\lambda);\lambda\right),
    \label{eq:app-network-effective-objective}
\end{equation}
we have
\begin{equation}
    \frac{d\F_{\mathrm{eff}}}{dq}
    =\left\langle v_c,
    \nabla_\Theta\F\!\left(\Theta^\star(q,\lambda);\lambda\right)
    \right\rangle.
    \label{eq:app-network-envelope}
\end{equation}
Consequently, stationary points of the full network objective near the critical branch are in one-to-one correspondence with stationary points of $\F_{\mathrm{eff}}$.

If, in addition, there is a local orthogonal involution $S$ satisfying
$S^2=I$, $Sv_c=-v_c$, $S\bar\Theta(\lambda)=\bar\Theta(\lambda)$, and
\begin{equation}
    \F\!\left(\bar\Theta(\lambda)+Sz;\lambda\right)
    =\F\!\left(\bar\Theta(\lambda)+z;\lambda\right),
    \label{eq:app-network-involution}
\end{equation}
then $\F_{\mathrm{eff}}(-q;\lambda)=\F_{\mathrm{eff}}(q;\lambda)$ and
\begin{equation}
    \F_{\mathrm{eff}}(q;\lambda)
    =\F_0(\lambda)+\frac12r(\lambda)q^2
    +\frac14u(\lambda)q^4+\mathcal O(q^6),
    \qquad r(\lambda_c)=0.
    \label{eq:app-network-landau-normal-form}
\end{equation}
If, furthermore,
\begin{equation}
    r'(\lambda_c)\neq0,
    \qquad
    u(\lambda_c)>0,
    \label{eq:app-network-transverse-crossing}
\end{equation}
then $\lambda_c$ is a generic codimension-one continuous phase boundary. The stationary branch $q=0$ is locally stable on the side where $r(\lambda)>0$; on the side where $r(\lambda)<0$, two symmetry-related stable branches emerge:
\begin{equation}
    q_\pm(\lambda)
    =\pm\sqrt{-\frac{r(\lambda)}{u(\lambda_c)}}
    +\mathcal O\!\left(|\lambda-\lambda_c|^{3/2}\right).
    \label{eq:app-network-bifurcating-branches}
\end{equation}
\end{theorem}

\begin{proof}
Write a nearby parameter vector as
$\Theta=\bar\Theta(\lambda)+qv_c+\eta$ with $\eta\in v_c^\perp$, and define
\begin{equation}
    G(q,\eta,\lambda)
    =P_\perp\nabla_\Theta\F\!\left(
    \bar\Theta(\lambda)+qv_c+\eta;\lambda
    \right).
\end{equation}
At $(0,0,\lambda_c)$, the derivative of $G$ with respect to $\eta$ is the restricted Hessian
$H_\perp=P_\perp H_c|_{v_c^\perp}$, which is positive definite and hence invertible by assumption. The implicit function theorem therefore yields the unique map $\eta^\star(q,\lambda)$ satisfying
\cref{eq:app-network-transverse-stationarity}. Because
\cref{eq:app-network-stationary-branch} implies $G(0,0,\lambda)=0$, uniqueness gives $\eta^\star(0,\lambda)=0$.

Along the reduced manifold, the transverse component of the gradient vanishes, so the gradient is parallel to $v_c$. Since $\partial_q\eta^\star\in v_c^\perp$, the chain rule gives
\begin{align}
    \frac{d\F_{\mathrm{eff}}}{dq}
    &=\left\langle
    \nabla_\Theta\F,
    v_c+\partial_q\eta^\star
    \right\rangle
    =\left\langle\nabla_\Theta\F,v_c\right\rangle,
\end{align}
which proves \cref{eq:app-network-envelope} and the stationary-point correspondence. At criticality, differentiating the transverse stationarity equation with respect to $q$ gives
$H_\perp\partial_q\eta^\star(0,\lambda_c)=-P_\perp H_cv_c=0$, hence
$\partial_q\eta^\star(0,\lambda_c)=0$. Therefore the reduced quadratic curvature vanishes:
$\partial_{qq}\F_{\mathrm{eff}}(0;\lambda_c)=v_c^\top H_cv_c=0$.

Under \cref{eq:app-network-involution}, orthogonality of $S$ implies that it preserves $v_c^\perp$. Both
$S\eta^\star(q,\lambda)$ and $\eta^\star(-q,\lambda)$ solve the same transverse stationarity equation, so uniqueness gives
$\eta^\star(-q,\lambda)=S\eta^\star(q,\lambda)$. Invariance of $\F$ then yields
$\F_{\mathrm{eff}}(-q;\lambda)=\F_{\mathrm{eff}}(q;\lambda)$. The even Taylor expansion and $r(\lambda_c)=0$ give
\cref{eq:app-network-landau-normal-form}.

Under \cref{eq:app-network-transverse-crossing}, the reduced stationarity equation factors as
\begin{equation}
    \partial_q\F_{\mathrm{eff}}(q;\lambda)
    =q\left[r(\lambda)+u(\lambda_c)q^2
    +\mathcal O\!\left(q^4+|\lambda-\lambda_c|q^2\right)\right].
    \label{eq:app-network-reduced-stationarity}
\end{equation}
The transverse condition $r'(\lambda_c)\neq0$ makes $r$ a valid local control coordinate, so the implicit function theorem applied to $q^2$ yields the two nonzero branches in \cref{eq:app-network-bifurcating-branches}. Positive definiteness on $v_c^\perp$ keeps every transverse direction stable. Along the reduced direction, $\partial_{qq}\F_{\mathrm{eff}}(0;\lambda)=r(\lambda)$, whereas on the nonzero branches
$\partial_{qq}\F_{\mathrm{eff}}(q_\pm;\lambda)=-2r(\lambda)+o(|r(\lambda)|)>0$ when $r(\lambda)<0$. This proves the stated local stability and the generic codimension-one continuous phase boundary.
\end{proof}

Here $q$ is a local Landau order-parameter coordinate. When the critical kernel is one-dimensional, the soft direction $v_c$ is unique up to sign, while the normalization of $q$ is a coordinate choice; the two branches $q_\pm$ are symmetry-related states along this same direction. In the canonical bilinear model, the nonzero $w$ projection lets $w$ serve directly as this coordinate.

To connect the full-network result with the scalar formula in the main text, choose a local coordinate $s$ having nonzero projection onto $v_c$ and collect all remaining parameters in $\psi$. If the Hessian is written as
\begin{equation}
    H=
    \begin{pmatrix}
        A & B^\top\\
        B & C
    \end{pmatrix},
    \qquad C\ \text{invertible},
    \label{eq:app-network-block-hessian}
\end{equation}
then eliminating the noncritical block gives the reduced mass
\begin{equation}
    r=A-B^\top C^{-1}B.
    \label{eq:app-network-schur-mass}
\end{equation}
At $r=0$, the critical soft direction is proportional to
$(1,-C^{-1}B)^\top$. Thus \cref{eq:scalar-effective-mass} is the two-parameter instance of the same Schur-complement reduction, while the vector $-C^{-1}B$ records the coordinated first-order response of all remaining layers.

\begin{remark}[Smooth, piecewise-smooth, and multicomponent cases]
\label{rem:app-network-scope}
The theorem applies directly to finite networks with smooth activations. For a ReLU network, the same argument applies inside an activation region whose interior contains the reference point. Transitions across activation-region boundaries require a nonsmooth analysis. The ReLU experiments assess phenotype-level transfer along finite regularization paths. If $\dim\ker H_c=k>1$, the same projection yields a vector order parameter $q\in\mathbb R^k$ and a multicomponent Landau reduction; only the critical subspace is intrinsic, while the basis and the components of $q$ may be rotated within it.
\end{remark}

\subsection{Joint quadratic corrections and the reduced effective mass}
\label{app:joint-quadratic-reduction}

To show how quadratic corrections on different parameter blocks jointly determine the same onset, consider the following local expansion with simultaneous sign-reversal symmetry:
\begin{equation}
 \F(w,\theta;\lambda)
 =L_0+\frac12A_\lambda w^2+K_\lambda w\theta
 +\frac12C_\lambda\theta^2
 +\F_{4,\lambda}(w,\theta)
 +O\!\left(\|(w,\theta)\|^6\right),
 \label{eq:app-general-joint-expansion}
\end{equation}
where $\F_{4,\lambda}$ is a homogeneous quartic polynomial. The quadratic Hessian is
\begin{equation}
 H_2(\lambda)
 =
 \begin{pmatrix}
 A_\lambda & K_\lambda\\
 K_\lambda & C_\lambda
 \end{pmatrix}.
 \label{eq:app-joint-hessian}
\end{equation}
If $C_\lambda>0$, the $\theta$ direction remains noncritical. Solving $\partial_\theta\F=0$ and applying the implicit function theorem gives
\begin{equation}
 \theta^\star(w;\lambda)
 =\kappa_\lambda w+O(w^3),
 \qquad
 \kappa_\lambda=-\frac{K_\lambda}{C_\lambda}.
 \label{eq:app-general-linear-slaving}
\end{equation}
Whenever the zero eigenvector of the critical Hessian has a nonzero $w$ component, $w$ is a valid local coordinate for the one-dimensional soft mode. If this component vanishes, the roles of the two coordinates can simply be interchanged.

Completing the square in the quadratic part yields
\begin{equation}
 \F_2(w,\theta;\lambda)
 =\frac12r_\lambda w^2
 +\frac{C_\lambda}{2}
 \left(\theta-\kappa_\lambda w\right)^2,
 \qquad
 r_\lambda=A_\lambda-\frac{K_\lambda^2}{C_\lambda}.
 \label{eq:app-schur-mass}
\end{equation}
Thus $r_\lambda$ is precisely the Schur complement of the $\theta$ block, and, for $C_\lambda>0$, the origin is locally stable if and only if $r_\lambda>0$. Substituting \cref{eq:app-general-linear-slaving} into the quartic term gives
\begin{equation}
 \F_{\mathrm{eff}}(w;\lambda)
 =L_0+\frac12r_\lambda w^2
 +\frac14u_\lambda w^4+O(w^6),
 \qquad
 u_\lambda=4\F_{4,\lambda}(1,\kappa_\lambda).
 \label{eq:app-general-reduced-coefficients}
\end{equation}
Note that $\theta^\star-\kappa_\lambda w=O(w^3)$ contributes only $O(w^6)$ through the completed-square quadratic term and therefore does not alter the quartic coefficient.

Now suppose that the algorithm adds a general quadratic correction at the reference point,
\begin{equation}
 \Delta\F_2
 =\frac{\lambda}{2}
 \left(q_{ww}w^2+2q_{w\theta}w\theta
 +q_{\theta\theta}\theta^2\right),
 \qquad
 Q=
 \begin{pmatrix}
 q_{ww}&q_{w\theta}\\
 q_{w\theta}&q_{\theta\theta}
 \end{pmatrix}.
 \label{eq:app-general-quadratic-correction}
\end{equation}
Letting $\kappa_0=-K_0/C_0$ and differentiating \cref{eq:app-schur-mass} directly gives
\begin{equation}
 \left.\frac{d r_\lambda}{d\lambda}\right|_{\lambda=0}
 =q_{ww}+2\kappa_0q_{w\theta}
 +\kappa_0^2q_{\theta\theta}
 =
 \begin{pmatrix}1&\kappa_0\end{pmatrix}
 Q
 \begin{pmatrix}1\\\kappa_0\end{pmatrix}.
 \label{eq:app-projected-mass-shift}
\end{equation}
The distinct operators $w^2$, $w\theta$, and $\theta^2$ all contribute to the effective mass along the critical soft mode. The sign after projection determines whether the quadratic correction suppresses or amplifies that mode. \Cref{eq:app-projected-mass-shift} gives the first-order shift at $\lambda=0$; the finite-$\lambda$ mass is given by \cref{eq:app-schur-mass}.

\subsection{Exact reduction of the canonical model}
\label{app:canonical-reduction}

For the truncated bilinear objective in the main text, define
\begin{equation}
    A=\mu+\lambda_2\gamma,
    \qquad
    B=m+\lambda_4\gamma,
    \label{eq:app-AB}
\end{equation}
and write
\begin{equation}
    \F(w,\theta)
    =
    L_0-aw\theta+\frac{B}{2}w^2\theta^2
    +\frac{A}{2}w^2+\frac{\mu}{2}\theta^2.
    \label{eq:app-single-energy}
\end{equation}
If $B>0$, then
\begin{equation}
    \partial_{\theta\theta}\F
    =
    \mu+Bw^2>0,
\end{equation}
Hence the elimination above is unique for every finite $w$, with
\begin{equation}
    \theta^\star(w)
    =
    \frac{aw}{\mu+Bw^2}.
    \label{eq:app-theta-branch}
\end{equation}

Moreover, the canonical objective satisfies the exact completed-square identity
\begin{equation}
    \F(w,\theta)
    =
    \F_{\mathrm{eff}}(w)
    +
    \frac{\mu+Bw^2}{2}
    \left(\theta-\theta^\star(w)\right)^2.
    \label{eq:app-completion-square}
\end{equation}
Because the second term is nonnegative, local minima of $\F_{\mathrm{eff}}$ are in one-to-one correspondence with local minima of the original two-dimensional objective.

\subsection{Local normal-form equivalence of BCE and squared loss}
\label{app:bce-local-equivalence}

The exact closed-form results in the main text use squared loss, but the same critical Landau structure does not depend on this choice. Let
$Y\in\{0,1\}$, let the single-feature logit be $z=\beta X=w\theta X$, and consider the binary cross-entropy
\begin{equation}
    \ell_{\mathrm{BCE}}(z,Y)
    =\log(1+e^z)-Yz .
    \label{eq:app-bce-pointwise}
\end{equation}
Near zero logit, the softplus expansion is
\begin{equation}
    \log(1+e^z)
    =\log 2+\frac{z}{2}+\frac{z^2}{8}
    -\frac{z^4}{192}+O(z^6).
    \label{eq:app-softplus-expansion}
\end{equation}
Assume that the required feature moments are finite, so that expectations may be taken term by term. Define
\begin{equation}
    a^{\mathrm{BCE}}_e
    =\E_e\!\left[\left(Y-\frac12\right)X\right],
    \qquad
    q_e=\E_e[X^2],
    \qquad
    h_e=\E_e[X^4],
\end{equation}
Then the BCE risk in environment $e$ satisfies
\begin{equation}
    L^{\mathrm{BCE}}_e(\beta)
    =\log 2-a^{\mathrm{BCE}}_e\beta
    +\frac{q_e}{8}\beta^2
    -\frac{h_e}{192}\beta^4+O(\beta^6).
    \label{eq:app-bce-beta-expansion}
\end{equation}
After substituting $\beta=w\theta$, the joint-parameter expansion through fourth order is
\begin{equation}
    L^{\mathrm{BCE}}_e(w,\theta)
    =L_{0,e}-a^{\mathrm{BCE}}_e w\theta
    +\frac{m^{\mathrm{BCE}}_e}{2}w^2\theta^2
    +O\!\left(\|(w,\theta)\|^8\right),
    \qquad
    m^{\mathrm{BCE}}_e=\frac{q_e}{4}.
    \label{eq:app-bce-joint-expansion}
\end{equation}
This has the same quadratic $w\theta$ coupling and quartic
$w^2\theta^2$ stabilizing term as the canonical squared-loss objective, after replacing the task signal and curvature by
$a^{\mathrm{BCE}}_e$ and $m^{\mathrm{BCE}}_e$. If the feature is centered,
$\E_e[X]=0$, then $a^{\mathrm{BCE}}_e=\E_e[YX]$, identical to the task signal under squared loss.
For the encoding $Y\in\{-1,+1\}$, using
$\log(1+e^{-Yz})$ similarly gives
$a^{\mathrm{BCE}}_e=\tfrac12\E_e[YX]$ and
$m^{\mathrm{BCE}}_e=q_e/4$; constant rescalings of the labels or logits merely reparameterize these coefficients.

Therefore, BCE and squared loss have the same Landau normal form near the zero-logit critical point
and induce the same effective \Rtwo/\Rfour order classification and phase-boundary mechanism. The equivalence holds near zero logit:
the higher-order terms in \cref{eq:app-bce-beta-expansion} generally do not vanish, so the exact reduced objective and
closed-form loading formula derived for squared loss are only asymptotic predictions near criticality for BCE. If an OOD algorithm constructs its penalty directly from risks or gradients,
its numerical coefficients must still be re-expanded under the BCE objective.

\section{Taylor expansion and Landau coefficients}
\label{app:landau-expansion}

Because
\begin{equation}
    \F(-w,-\theta)=\F(w,\theta),
    \qquad
    \theta^\star(-w)=-\theta^\star(w),
\end{equation}
the reduced objective satisfies
$\F_{\mathrm{eff}}(-w)=\F_{\mathrm{eff}}(w)$.
This symmetry excludes odd powers from the reduced Taylor expansion. The two branches $\pm w$ correspond to the same $\beta=w\theta$, so the physically relevant order-parameter magnitude is $|w|$.

Substituting \cref{eq:app-theta-branch} into \cref{eq:app-single-energy} gives the exact reduced objective
\begin{equation}
    \F_{\mathrm{eff}}(w)
    =
    L_0+\frac{A}{2}w^2
    -\frac{a^2w^2}{2(\mu+Bw^2)}.
    \label{eq:app-reduced-energy}
\end{equation}
Using
\begin{equation}
    \frac{1}{\mu+Bw^2}
    =
    \frac1\mu-\frac{B}{\mu^2}w^2
    +\frac{B^2}{\mu^3}w^4+O(w^6),
\end{equation}
we obtain
\begin{equation}
    \F_{\mathrm{eff}}(w)
    =
    L_0+\frac12rw^2+\frac14uw^4+O(w^6),
    \label{eq:app-landau-expansion}
\end{equation}
where
\begin{equation}
    r
    =
    A-\frac{a^2}{\mu}
    =
    \gamma(\lambda_2-\lambda_{2,c}),
    \qquad
    \lambda_{2,c}
    =
    \frac{a^2/\mu-\mu}{\gamma},
    \qquad
    u
    =
    \frac{2a^2B}{\mu^2}.
    \label{eq:app-landau-coefficients}
\end{equation}
Provided that
\begin{equation}
    B=m+\lambda_4\gamma>0,
    \label{eq:app-positive-quartic}
\end{equation}
we have $u>0$, so the quartic term stabilizes the reduced free energy near the critical point. Higher-order task terms, if present, modify only coefficients of order $O(w^6)$ and above and do not change the leading division of roles between the quadratic mass and quartic stiffness.

When the two layers have different baseline curvatures,
\begin{equation}
    \F
    =
    L_0-aw\theta+\frac{B}{2}w^2\theta^2
    +\frac{A_w}{2}w^2+\frac{\mu_\theta}{2}\theta^2,
\end{equation}
the same calculation gives
\begin{equation}
    r=A_w-\frac{a^2}{\mu_\theta},
    \qquad
    u=\frac{2a^2B}{\mu_\theta^2}.
\end{equation}
Thus the shared curvature $\mu$ in the main text is only a notational simplification.

\section{Equilibrium branches and loading}
\label{app:equilibrium-results}

\begin{theorem}[Exact phase boundary of the canonical model]
\label{thm:app-exact-onset}
Let $A>0$ and $B>0$. The derivative of the canonical reduced objective is
\begin{equation}
    \F_{\mathrm{eff}}'(w)
    =
    w\left[
    A-\frac{a^2\mu}{(\mu+Bw^2)^2}
    \right].
    \label{eq:app-exact-derivative}
\end{equation}
The origin is the unique stable equilibrium when $a^2\leq A\mu$; when $a^2>A\mu$, a pair of stable nonzero equilibria exists. Therefore, the exact phase boundary is
\begin{equation}
    a^2=A\mu
    \quad\Longleftrightarrow\quad
    \lambda_2=\frac{a^2/\mu-\mu}{\gamma}.
    \label{eq:app-exact-boundary}
\end{equation}
\end{theorem}

\begin{proof}
Let
\[
g(s)=A-\frac{a^2\mu}{(\mu+Bs)^2},
\qquad s=w^2\geq0.
\]
Because $B>0$, $g(s)$ is strictly increasing and converges to $A>0$. If
$g(0)=A-a^2/\mu\geq0$, then
$\F_{\mathrm{eff}}'(w)$ has the same sign as $w$, and the origin is the unique minimum.
If $g(0)<0$, then $g$ has a unique zero on the positive half-line, producing a symmetric pair of nonzero minima while the origin becomes a local maximum. The boundary is therefore $g(0)=0$.
\end{proof}

\begin{proposition}[Equilibrium branches and the mean-field critical exponent]
\label{prop:app-equilibrium-branches}
Suppose that $u(\lambda_c)>0$ and
\begin{equation}
    r(\lambda)
    =
    c(\lambda-\lambda_c)+o(\lambda-\lambda_c),
    \qquad c>0,
    \label{eq:app-linear-mass}
\end{equation}
Then the stable equilibrium branches at zero external field satisfy
\begin{equation}
    w^\star=0
    \quad (r>0),
    \qquad
    |w^\star|
    =
    \sqrt{-\frac{r}{u}}
    +O(|r|^{3/2})
    \quad (r<0).
    \label{eq:app-equilibrium-branches}
\end{equation}
Consequently,
\begin{equation}
    |w^\star|
    \sim
    \sqrt{\frac{c}{u(\lambda_c)}}\,
    (\lambda_c-\lambda)^{1/2},
    \label{eq:app-critical-scaling}
\end{equation}
i.e., the order parameter has the mean-field critical exponent $1/2$.
\end{proposition}

\begin{proof}
From
$\partial_w\F_{\mathrm{eff}}
=rw+uw^3+O(w^5)=0$,
the nonzero branch satisfies
$w^2=-r/u+O(r^2)$.
When $u(\lambda_c)>0$, this branch is a local minimum on the $r<0$ side.
Substituting \cref{eq:app-linear-mass} then gives the result.
\end{proof}

\begin{proposition}[Second-order nonanalyticity of the minimum free energy]
\label{prop:app-free-energy-singularity}
Under the conditions of Proposition~\ref{prop:app-equilibrium-branches},
\begin{equation}
    \F_{\min}(\lambda)-\F_{\min}(\lambda_c)
    =
    \begin{cases}
        0, & r\geq0,\\[1mm]
        \displaystyle-\frac{r^2}{4u(\lambda_c)}
        +O(|r|^3), & r<0.
    \end{cases}
    \label{eq:app-free-energy-singularity}
\end{equation}
Thus the equilibrium order parameter emerges continuously, while the first derivative of the minimum free energy with respect to the control parameter remains continuous and the second derivative jumps at the critical point. This is the standard mean-field second-order Landau transition.
\end{proposition}

\begin{proof}
For $r\geq0$, the local minimum is $w^\star=0$. For $r<0$, substituting
$w_\star^2=-r/u+O(r^2)$
into
$\F_{\mathrm{eff}}-\F_0=rw^2/2+uw^4/4+O(w^6)$
yields \cref{eq:app-free-energy-singularity}.
\end{proof}

\begin{theorem}[Exact loading and the distinct roles of quadratic and quartic terms]
\label{thm:app-loading}
If $B>0$, the stable equilibrium of \cref{eq:app-single-energy} satisfies
\begin{equation}
    |\beta^\star|
    =
    \frac{\left[|a|-\sqrt{A\mu}\right]_+}{B}.
    \label{eq:app-loading}
\end{equation}
Thus $\lambda_2$ drives the loading exactly to zero at the finite strength
$\lambda_{2,c}=(a^2/\mu-\mu)/\gamma$
whereas a pure $\lambda_4$ term only compresses a nonzero loading continuously by increasing $B$.
\end{theorem}

\begin{proof}
A nonzero stationary point satisfies
\begin{equation}
    Aw=\theta(a-B\beta),
    \qquad
    \mu\theta=w(a-B\beta).
    \label{eq:app-stationary-pair}
\end{equation}
Multiplying the two equations and canceling the nonzero factor $w\theta$ gives
\begin{equation}
    (a-B\beta)^2=A\mu.
\end{equation}
Selecting the branch with the same sign as the task drive and that connects continuously to zero as $a^2\downarrow A\mu$ gives \cref{eq:app-loading}.
\end{proof}

\paragraph{Sixth-order loading extension.}
For the higher-order predictions in \cref{fig:boundaries}(b), we apply the same soft-mode reduction but retain the sextic coefficient.  Writing the sixth-order truncation as
\begin{equation}
    \F_{\mathrm{eff}}^{(6)}(q)
    =\F_0+\frac12 r q^2+\frac14 u q^4+\frac16 v_6 q^6,
    \label{eq:app-sixth-order-free-energy}
\end{equation}
and setting $s=q^2$, every nonzero stationary branch satisfies
\begin{equation}
    r+us+v_6s^2=0,
    \qquad
    s_\star=
    \frac{-u+\sqrt{u^2-4v_6r}}{2v_6},
    \label{eq:app-sixth-order-branch}
\end{equation}
where the displayed root is the branch continuous with $s_\star=-r/u$ as $v_6\to0$ (and the stable positive root is selected whenever more than one real branch exists).  For a scalar bilinear soft mode, the slaved loading has the even expansion
\begin{equation}
    \beta(q)=b_2q^2+b_4q^4+O(q^6),
    \qquad
    |\beta^\star|
    =\left|b_2s_\star+b_4s_\star^2\right|+O(s_\star^3).
    \label{eq:app-sixth-order-loading}
\end{equation}
The limit $v_6\to0$ recovers the quartic branch, while the canonical explicit $R_2/R_4$ model further admits the exact loading formula in \cref{eq:app-loading}.  Thus the quartic and sixth-order predictions use the same reduced equilibrium law, differing only in the retained low-order coefficients.  For V-REx and IGA these coefficients are obtained from the objective expansions in \cref{eq:app-vrex-risk-variance,eq:app-iga-scalar-full}; they are not fitted to the trained loading paths.

\begin{proposition}[Selective-retention window]
\label{prop:app-window}
Suppose that the stable and shortcut directions have critical values
$\lambda^{(c)}_{2,c}$ and $\lambda^{(s)}_{2,c}$, respectively, with
$\lambda^{(s)}_{2,c}<\lambda^{(c)}_{2,c}$. Then any
\begin{equation}
    \lambda^{(s)}_{2,c}<\lambda_2<\lambda^{(c)}_{2,c}
    \label{eq:app-window}
\end{equation}
places the shortcut direction in the disordered phase while the stable direction remains ordered.
\end{proposition}

\section{Onset diagnostics, susceptibility, and Landau dynamics}
\label{app:dynamic-diagnostic}

We first explain why the early-time growth-rate zero crossing used in the experiments measures the same static phase boundary, and then give a local dynamical interpretation under deterministic continuous-time gradient flow.

\subsection{Linearized onset diagnostic}
Let $z=(w,\theta)^\top$. The continuous-time gradient flow of the canonical objective linearized near the origin is
\begin{equation}
    \dot z=-H_0z,
    \qquad
    H_0=
    \begin{pmatrix}
        A & -a\\
        -a & \mu
    \end{pmatrix}.
    \label{eq:app-linearized-flow}
\end{equation}
The fastest local growth rate is $-\lambda_{\min}(H_0)$, so its zero crossing occurs when
\begin{equation}
    \lambda_{\min}(H_0)=0
    \quad\Longleftrightarrow\quad
    \det H_0=A\mu-a^2=0
    \quad\Longleftrightarrow\quad
    r(\lambda_2)=0.
    \label{eq:app-dynamic-static-equivalence}
\end{equation}
Thus the measured dynamical zero crossing is a local numerical diagnostic of the static Landau boundary.

\subsection{Nonlinear soft-mode dynamics under gradient flow}
Consider deterministic continuous-time gradient flow
\begin{equation}
    \dot\Theta=-\nabla_\Theta\F(\Theta;\lambda).
    \label{eq:app-full-gradient-flow}
\end{equation}
At a simple soft-mode critical point satisfying the spectral assumptions of \cref{thm:app-network-soft-mode}, the linearized flow has one neutral direction and all transverse directions decay. Both reductions use the order-parameter coordinate $q$ along the critical direction $v_c$. The equilibrium map $\eta^\star$ enforces transverse stationarity; the dynamical map $\eta_{\rm cm}$ enforces invariance under gradient flow. The center-manifold theorem gives a locally invariant one-dimensional manifold, which we parameterize as
\begin{equation}
    \Theta_{\rm cm}(q,\lambda)
    =\bar\Theta(\lambda)+qv_c+\eta_{\rm cm}(q,\lambda),
    \qquad
    \eta_{\rm cm}\perp v_c .
    \label{eq:app-dynamic-center-manifold}
\end{equation}
Define the objective restricted to this manifold by
$\F_{\rm cm}(q;\lambda)=\F(\Theta_{\rm cm}(q,\lambda);\lambda)$.

\begin{proposition}[Time-dependent Landau normal form]
\label{prop:app-landau-dynamics}
Under the same local $\mathbb Z_2$ symmetry and transverse-crossing assumptions as in \cref{thm:app-network-soft-mode}, with $u(\lambda_c)>0$, the center-manifold dynamics satisfy
\begin{equation}
    \tau(q,\lambda)\dot q
    =-\frac{\partial\F_{\rm cm}}{\partial q},
    \qquad
    \tau(q,\lambda)
    =\left\|\partial_q\Theta_{\rm cm}(q,\lambda)\right\|^2>0.
    \label{eq:app-reduced-gradient-flow}
\end{equation}
Near criticality,
\begin{equation}
    \F_{\rm cm}(q;\lambda)
    =\F_0(\lambda)+\frac12r(\lambda)q^2
    +\frac14u(\lambda)q^4+\mathcal O(q^6),
    \label{eq:app-dynamic-landau-potential}
\end{equation}
so, after absorbing the positive local mobility into the time scale,
\begin{equation}
    \tau_c\dot q
    =-r(\lambda)q-u(\lambda_c)q^3
    +\mathcal O\!\left(q^5+|\lambda-\lambda_c|q^3\right),
    \qquad \tau_c>0.
    \label{eq:app-learning-landau-dynamics}
\end{equation}
\end{proposition}

\begin{proofsketch}
Because the center manifold is invariant under \cref{eq:app-full-gradient-flow},
$\partial_q\Theta_{\rm cm}\,\dot q=-\nabla_\Theta\F$ along it. Taking the inner product with $\partial_q\Theta_{\rm cm}$ gives
$\|\partial_q\Theta_{\rm cm}\|^2\dot q=-\partial_q\F_{\rm cm}$, which proves \cref{eq:app-reduced-gradient-flow}. The symmetry makes $\F_{\rm cm}$ even in $q$. With $r(\lambda_c)=0$ and the assumed $u(\lambda_c)>0$, differentiating the Landau expansion yields \cref{eq:app-learning-landau-dynamics}.
\end{proofsketch}

For comparison, the spatially uniform Landau free energy of a nonconserved scalar order parameter is
\begin{equation}
    F_{\rm mag}(M)
    =F_0+\frac12a_0(T-T_c)M^2+\frac14bM^4+\mathcal O(M^6),
\end{equation}
and its standard relaxational time-dependent Landau dynamics are \citep{hohenberg2015ginzburg}
\begin{equation}
    \tau_M\dot M
    =-\frac{\partial F_{\rm mag}}{\partial M}
    =-a_0(T-T_c)M-bM^3+\mathcal O(M^5).
    \label{eq:app-magnetic-landau-dynamics}
\end{equation}
Under the correspondence
$q\leftrightarrow M$, $r(\lambda)\leftrightarrow a_0(T-T_c)$, and $u\leftrightarrow b$, \cref{eq:app-learning-landau-dynamics,eq:app-magnetic-landau-dynamics} have the same local normal form. This spatially uniform relaxational dynamics is often referred to as Model-A or time-dependent Ginzburg--Landau dynamics. In particular, in the linear regime,
\begin{equation}
    \frac{d}{dt}\log|q|
    =-\frac{r(\lambda)}{\tau_c}+o(1),
    \label{eq:app-early-growth-mass}
\end{equation}
so the growth-rate zero crossing, the Hessian sign change, and the Landau mass condition $r=0$ identify the same local phase boundary.

The same normal form predicts critical slowing down. On the disordered side $r>0$, linearization around $q^\star=0$ gives
\begin{equation}
    \tau_{\rm relax}=\frac{\tau_c}{r}\,[1+o(1)].
\end{equation}
On the ordered side $r<0$, where $(q^\star)^2=-r/u+O(r^2)$, the local curvature is $\partial_{qq}\F_{\rm cm}(q^\star)=-2r+o(|r|)$ and hence
\begin{equation}
    \tau_{\rm relax}=\frac{\tau_c}{2|r|}\,[1+o(1)].
\end{equation}
Thus the relaxation time diverges as $|r|^{-1}$ when the phase boundary is approached from either side, giving the standard mean-field critical slowing down of relaxational Landau dynamics \citep{hohenberg2015ginzburg}.

\subsection{Susceptibility and critical slowing down}
\label{app:soft-mode-response}

We test whether the same critical curvature controls equilibrium response and relaxation in the canonical bilinear model and a smooth finite network. These experiments compare explicit \Rtwo{} and \Rfour{} controls under deterministic gradient flow.

\edef\responsesavedtopnumber{\number\value{topnumber}}
\setcounter{topnumber}{1}
\begin{figure}[t]
\centering
\includegraphics[width=\textwidth]{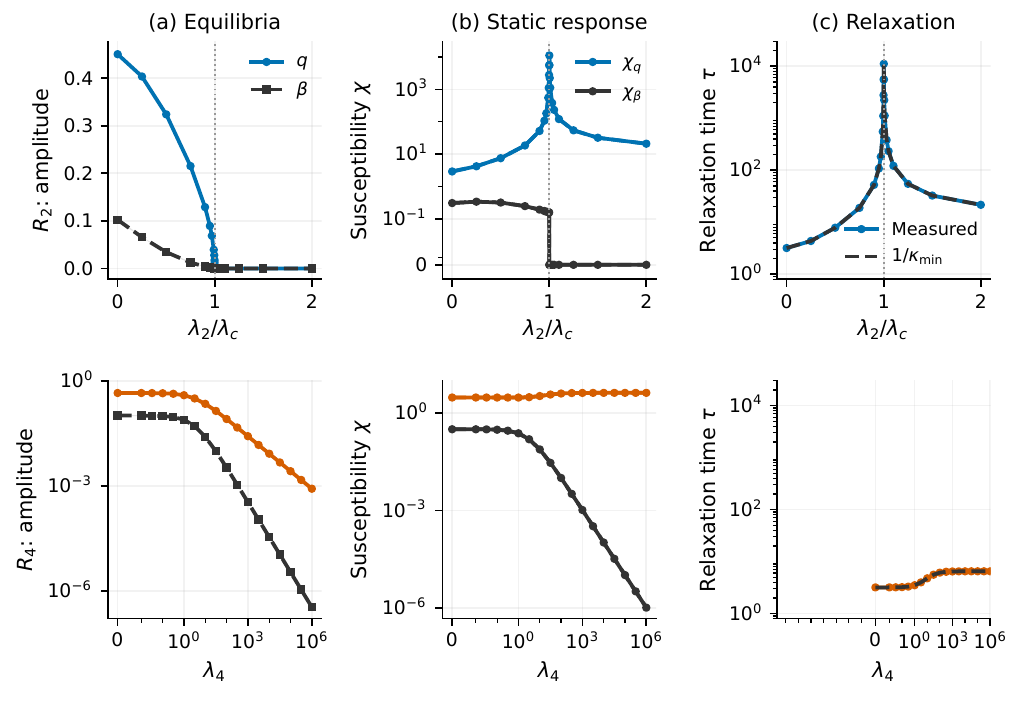}
\caption{Equilibrium amplitudes, static responses, and relaxation times in the smooth network under \Rtwo{} (top) and \Rfour{} (bottom). In the response and relaxation panels, markers are numerical measurements and dashed curves are Hessian predictions. Time is in gradient-flow units; \Rtwo{} strength is normalized by each seed's critical value.}
\label{fig:app-response-nonlinear}
\end{figure}

\begin{figure}[t]
\centering
\includegraphics[width=\textwidth]{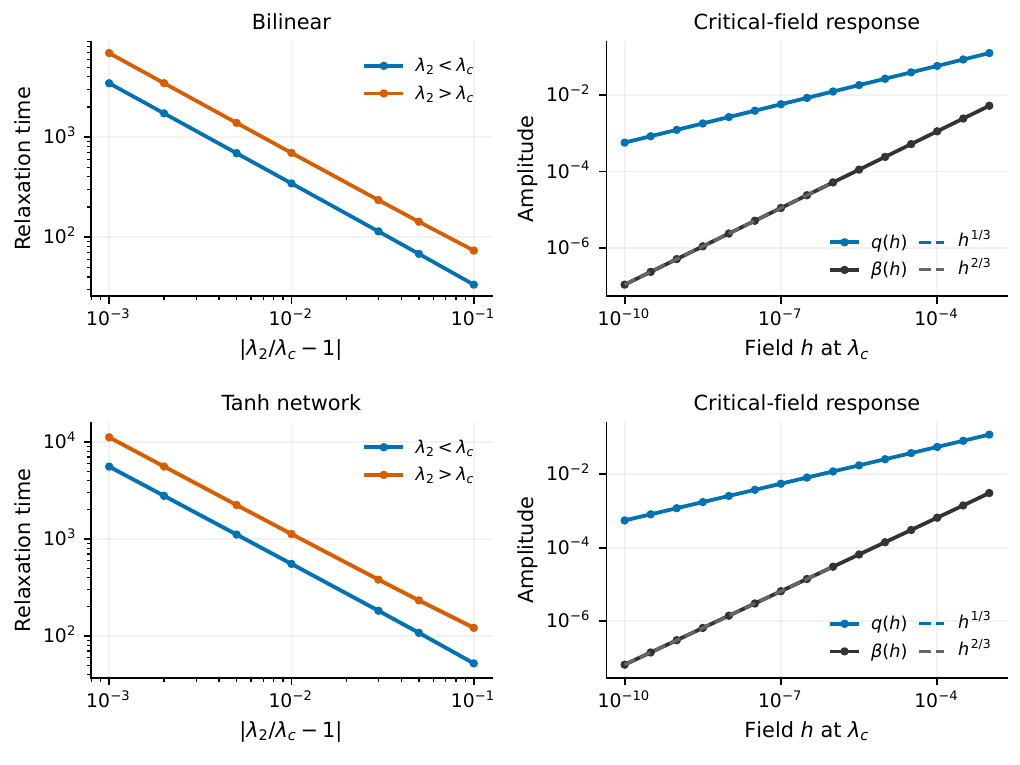}
\caption{Critical scaling in the bilinear and smooth-network controls. Left: relaxation time versus distance from the \Rtwo{} boundary. Right: amplitudes under $-hq$ at criticality; dashed guides show $h^{1/3}$ and $h^{2/3}$ scaling, anchored at the smallest sampled field.}
\label{fig:app-response-scaling}
\end{figure}

\paragraph{Response predictions.}
A field conjugate to the reduced coordinate gives
\begin{equation}
    \F_{\mathrm{eff}}(q;h)
    =\F_0+\frac12rq^2+\frac14uq^4-hq+O(q^6).
    \label{eq:app-response-potential}
\end{equation}
For $u>0$, differentiating the equilibrium condition on a fixed stable branch yields
\begin{equation}
    \chi_q
    =\left.\frac{\partial q^\star}{\partial h}\right|_{h=0}
    =\begin{cases}
        r^{-1}[1+o(1)], & r>0,\\[2pt]
        (2|r|)^{-1}[1+o(1)], & r<0.
    \end{cases}
    \label{eq:app-response-susceptibility}
\end{equation}
Together with the relaxation law above and a transverse crossing of $r$, this predicts
$\chi_q,\tau_{\mathrm{relax}}\propto|\lambda_2-\lambda_c|^{-1}$.
At $r=0$, the nonlinear response satisfies $q\sim(h/u)^{1/3}$.
Increasing quartic stiffness at fixed negative mass can reduce the equilibrium amplitude while its curvature remains finite, distinguishing continuous attenuation from critical softening.

For a differentiable probe $p(\Theta)$ of the full network, consider
$\F_h=\F-hp$. At a nondegenerate stable equilibrium with Hessian $H$, differentiating the full stationarity equations gives the exact local response
\begin{equation}
    \chi_p
    =\left.\frac{d\,p(\Theta_h^\star)}{dh}\right|_{h=0}
    =\nabla p^\top H^{-1}\nabla p.
    \label{eq:app-full-probe-response}
\end{equation}
\paragraph{Models and environments.}
The bilinear control uses \cref{eq:local-learning-objective} with
$(a,m,\gamma,\mu)=(0.5,2.25,1,0.2)$, giving $\lambda_c=1.05$.
The smooth network has eight inputs and one $\tanh$ unit,
$f_\Theta(X)=\theta\tanh(w^\top X)$, with parameters
$\Theta=(w,\theta)$.
For fixed orthonormal directions $b,c\in\mathbb R^8$, we draw
\begin{equation}
    \widetilde X_i=(0.5b+0.06c)Y_i+\xi_i,
    \qquad
    X_{ei}=\widetilde X_i-\overline{\widetilde X}+d_e b,
    \quad d_e\in\frac{1}{\sqrt5}\{-3,-1,1,3\},
    \label{eq:app-response-data}
\end{equation}
where $Y_i\in\{-1,1\}$ is balanced and $\xi_i\sim\mathcal N(0,I_8)$.
Each environment contains the same 1,024 base examples, differing only in its offset. The task objective is mean squared error with factor $1/2$ and ridge $\mu\|\Theta\|^2/2$, where $\mu=0.2$.
For the nonlinear experiment, we construct two mean-matching controls whose leading contributions near the zero reference are quadratic (\Rtwo{}) and quartic (\Rfour{}), respectively.
Writing $\bar\phi_e$ for the environment mean of $\tanh(w^\top X)$ and
$g_e=(\bar\phi_e-\bar\phi)/2$, we set
\begin{equation}
    \Omega_2=\frac{\lambda_2}{2}\|g(\Theta)\|^2,
    \qquad
    \Omega_4=\frac{\lambda_4}{2}\|\theta g(\Theta)\|^2.
    \label{eq:app-response-controls}
\end{equation}
At the zero reference, $g=0$; with $J_E=\partial g/\partial\Theta$, the environmental quadratic operator is $G_E=J_E^\top J_E$ for \Rtwo{}, whereas \Rfour{} has zero Hessian there. The critical reference has a one-dimensional Hessian kernel and positive transverse curvature in every run.

Both scans use $q=v_c^\top\Theta$, with $v_c$ the unit critical eigenvector at the \Rtwo{} boundary, held fixed along both scans. We separately measure predictive loading: $\beta=w\theta$ in the bilinear model and
\begin{equation}
    \beta(\Theta)
    =\frac{\widehat{\E}[(b^\top X)f_\Theta(X)]}
    {\widehat{\E}[(b^\top X)^2]}
    \label{eq:app-response-predictive-loading}
\end{equation}
in the smooth network, with empirical averages over the pooled training environments. Thus $q$ is a parameter-space critical coordinate, while $\beta$ measures the predictor's linear projection onto the designated input direction. Figures show unnormalized amplitudes.

\paragraph{Observable-specific responses.}
We denote the response to $\F-h_q q$ by $\chi_q$ (order-parameter susceptibility), and that to $\F-h_\beta\beta$ by $\chi_\beta$ (predictive-loading response). Each differentiates its own observable with respect to its own field at zero field, as in \cref{eq:app-full-probe-response}. In the canonical model, $-h_\beta\beta$ replaces the task drive $a$ by $a+h_\beta$, so $\chi_\beta=\partial\beta^\star/\partial a$ with the other coefficients fixed. Near criticality, $\beta\propto q^2$, so its conjugate perturbation changes the quadratic coefficient and its response can remain finite as $\chi_q$ diverges.

\paragraph{Measurements.}
We solve the full stationarity equations and verify positive Hessian curvature at each equilibrium. The Hessian zero crossing determines $\lambda_c$.
For each probe, central differences between equilibria of $\F\mp hp$ measure susceptibility; repeating the measurement at $h/2$ checks the linear-response regime. On the acquired side, both perturbations follow the same local branch.
Relaxation uses unit-mobility gradient flow $\dot\Theta=-\nabla_\Theta\F$. For ordinary gradient descent with a small fixed step size $\eta$, $t\simeq\eta k$ after $k$ steps, so $\tau_{\mathrm{steps}}\simeq\tau/\eta$.
At each equilibrium $\Theta^\star$, let $(\kappa_{\min},v_{\min})$ be the smallest eigenpair of $H=\nabla_\Theta^2\F(\Theta^\star)$, with $\|v_{\min}\|=1$ and $\kappa_{\min}>0$. We integrate from $\Theta_\pm(0)=\Theta^\star\pm\epsilon v_{\min}$ and fit the projected displacement:
\[
    d(t)=\frac12v_{\min}^\top[\Theta_+(t)-\Theta_-(t)]
    \simeq d(0)e^{-t/\tau},
    \qquad \tau_{\mathrm{pred}}=\kappa_{\min}^{-1}.
\]
The plotted $\tau$ is the fitted time for the local parameter-displacement amplitude to decrease by a factor $e$, in the above gradient-flow units. The current direction $v_{\min}$ determines the slowest local relaxation, while $q$ uses the fixed critical direction $v_c$; the relation in \cref{eq:main-susceptibility} holds asymptotically near criticality.
We use five independent data and initialization seeds for the smooth network, and five initialization repeats of the fixed bilinear population objective. Figures report means and standard errors.

\paragraph{Critical response and relaxation.}
The \Rtwo{} paths show the predicted inverse-distance scaling of $\chi_q$ and the relaxation time (\cref{fig:app-response-nonlinear,fig:app-response-scaling}). At criticality, the finite-field response follows $q\propto h^{1/3}$ and $\beta\propto h^{2/3}$, consistent with $\beta\propto q^2$. In these controls, \Rfour{} attenuation retains finite relaxation times even at small loading.

\FloatBarrier
\setcounter{topnumber}{\responsesavedtopnumber}

\begin{figure}[!t]
\centering
\includegraphics[width=\textwidth]{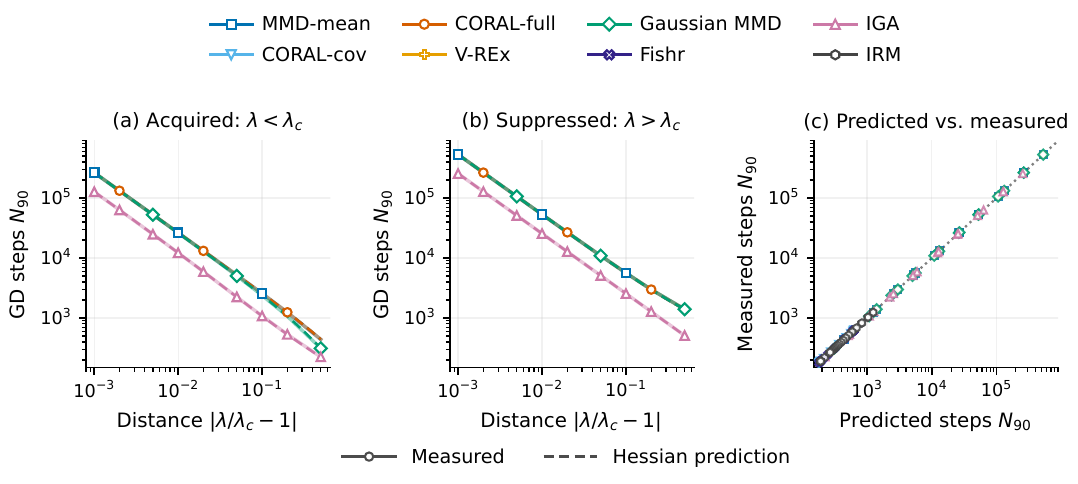}
\caption{Recovery-step diagnostics for the bilinear model. Panels~(a,b) show distance-to-boundary scaling for \Rtwo{}-like objectives on either side of $\lambda_c$: solid curves with markers are measured steps, and dashed curves are Hessian predictions. Panel~(c) compares predicted and measured steps for all eight objectives; the dotted line is $y=x$.}
\label{fig:app-objective-scaling}
\par\medskip
\includegraphics[width=\textwidth]{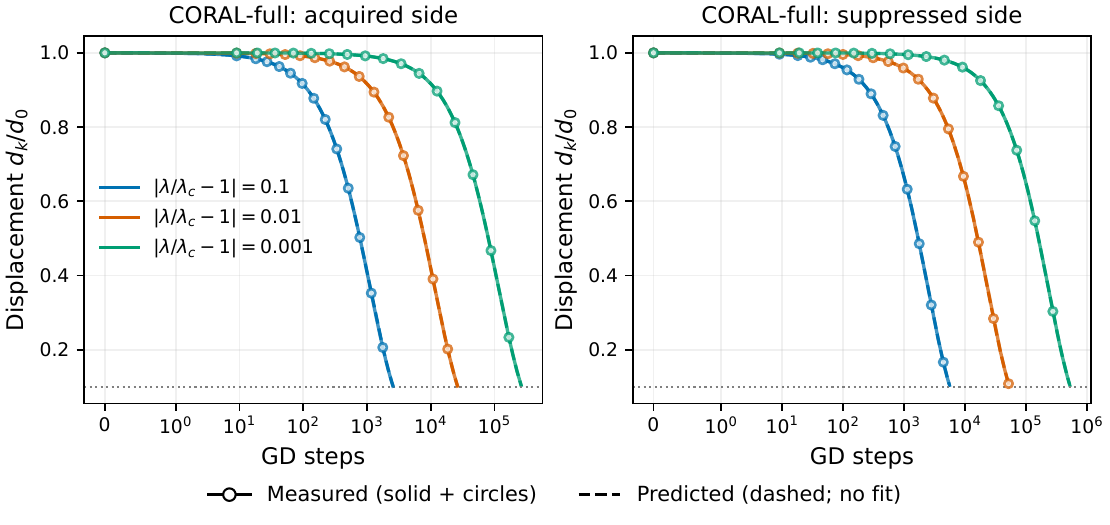}
\caption{CORAL-full relaxation on both sides of the boundary. Colors denote distance to criticality. Solid curves with circles show $d_k/d_0$; dashed curves show $(1-\eta\kappa_{\min})^k$. The dotted line marks the $0.1$ threshold defining $N_{90}$.}
\label{fig:app-objective-traces}
\end{figure}

\subsection{Optimization steps near equilibrium for invariant-learning objectives}
\label{app:objective-relaxation}

We supplement \cref{fig:objective-dynamics} with local recovery measurements under the same eight objectives in the bilinear model. Here $d_k$ is the projected half-difference of paired GD trajectories along the minimum-curvature Hessian direction, and $N_{90}$ is the first step with $|d_k/d_0|\leq0.1$. Predictions follow \cref{eq:main-recovery-steps}; the data and measurement protocol are given in Appendix~\ref{app:objective-dynamics-protocol}.

\Cref{fig:app-objective-scaling} checks the inverse-distance scaling and the agreement of predicted and measured steps, whose relative difference stays below $0.6\%$. \Cref{fig:app-objective-traces} shows the underlying CORAL-full trajectories. IRM's finite peak in \cref{fig:objective-dynamics} arises as its negative \Rsix{} term softens the acquired equilibrium while the minimum Hessian curvature remains positive.

\section{Collective modes and environment-induced mass}
\label{app:collective-proofs}

Let $C=\diag(c_1,\ldots,c_p)$ and $M=\mu I+\lambda_2\Gamma_E$, with $\Gamma_E\succeq0$. The multi-feature quadratic Hessian is
\begin{equation}
 H_0^{(p)}=
 \begin{pmatrix}
 M&-C\\-C^\top&\mu I
 \end{pmatrix}.
 \label{eq:app-matrix-hessian}
\end{equation}

\begin{theorem}[Collective-mode onset]
\label{thm:app-collective}
Assume $M\succ0$. A collective direction grows away from the origin if and only if
\begin{equation}
 \lambda_{\max}\!\left[
 \frac1\mu C^\top M^{-1}C
 \right]>1.
 \label{eq:app-collective-boundary}
\end{equation}
At criticality, the largest eigenvalue equals $1$. The corresponding gating direction can be written as $w_c\propto M^{-1}Cu_c$, where $u_c$ is the leading eigenvector in \cref{eq:app-collective-boundary}.
\end{theorem}

\begin{proof}
Because $M\succ0$, taking the Schur complement of the upper-left block in \cref{eq:app-matrix-hessian} gives
\begin{equation}
 H_0^{(p)}\succeq0
 \quad\Longleftrightarrow\quad
 S\equiv\mu I-C^\top M^{-1}C\succeq0.
 \label{eq:app-schur}
\end{equation}
Multiplying $S$ on both sides by $\mu^{-1/2}I$ shows that $S$ has a negative eigenvalue if and only if the matrix in \cref{eq:app-collective-boundary} has an eigenvalue larger than $1$. At the boundary, the quadratic stationarity equation $Mw=C\theta$ gives $w_c\propto M^{-1}Cu_c$.
\end{proof}

\begin{proof}[Proof of Proposition~\ref{prop:predictor-selectivity}]
For $\Gamma_E\succeq0$, $x^\top\Gamma_E x=\norm{\Gamma_E^{1/2}x}^2$ vanishes exactly when $x\in\ker\Gamma_E$. Applying this identity to $\bm w$ and $\bm\beta$, with positive $\lambda_2,\lambda_4$, proves both equivalences. To see why zero gate penalty need not imply zero predictor penalty, take $\Gamma_E=dd^\top$ with $d=(1,-1)^\top$, $\bm w=(1,1)^\top$, and $\bm\theta=(1,2)^\top$. Then $\bm\beta=(1,2)^\top$, so $R_2(\bm w)=0$ but $R_4(\bm\beta)=\lambda_4/2>0$.
\end{proof}

\paragraph{Compensatory example (Fig.~7).}
Let $\delta_c,\delta_s>0$ denote the environmental mean-shift amplitudes of the pair and shortcut, and set $d_{\mathrm{pair}}=(\delta_c,-\delta_c,0)^\top$ and $d_{\mathrm{short}}=(0,0,\delta_s)^\top$. The oracle sensitivity matrix is
\begin{equation}
\Gamma_E=d_{\mathrm{pair}}d_{\mathrm{pair}}^\top+d_{\mathrm{short}}d_{\mathrm{short}}^\top,
\label{eq:app-compensatory-gamma}
\end{equation}
so direct expansion gives
\begin{equation}
\bm w^\top\Gamma_E\bm w=\delta_c^2(w_1-w_2)^2+\delta_s^2w_3^2,
\qquad
\bm\beta^\top\Gamma_E\bm\beta=\delta_c^2(\beta_1-\beta_2)^2+\delta_s^2\beta_3^2.
\label{eq:app-compensatory-forms}
\end{equation}
Hence the compensated direction $(1,1,0)^\top/\sqrt2$ lies in $\ker(\Gamma_E)$, whereas the leakage direction $(1,-1,0)^\top/\sqrt2$ and shortcut direction $e_3$ have positive sensitivity. Matrix \Rfour directly penalizes predictor-space leakage $\beta_1-\beta_2$ and shortcut loading $\beta_3$, leaving the compensated direction unpenalized; matrix \Rtwo applies the same geometry to the gates $\bm w$.

\section{Derivation of objective signatures}
\label{app:fingerprint-derivations}

This section derives, term by term, the fingerprints of all methods appearing in the Figure~5 legend:
\ERM, GroupDRO, IRMv1, V-REx, IGA, Fish/MLDG, Fishr, Mixup, MMD-mean, Gaussian MMD, CORAL-cov, CORAL-full, DANN, and the explicit \Rtwo{}, \Rfour{}, and \Rtwo{}+\Rfour{} objectives. All identities refer to the bilinear population probe in \cref{eq:app-env-risk}. Their application to deep networks and alternating optimization describes local operators under the stated reference assumptions.

We use $\Var_{P_e}$ and $\Cov_{P_e}$ for within-environment moments, and $\Var_e$ and $\Cov_e$ for variation across environments. Let $m_e=\E_e[X]$ and $s_e=\Cov_{P_e}(X)$, and define
\begin{equation}
 \begin{aligned}
 M_2&=\overline{(m_e-\bar m)^2},\quad
 S_2=\overline{(s_e-\bar s)^2},\quad
 V_c=\overline{(c_e-\bar c)^2},\\
 C_{cq}&=\overline{(c_e-\bar c)(q_e-\bar q)},\quad
 V_q=\overline{(q_e-\bar q)^2}.
 \end{aligned}
 \label{eq:app-env-dispersions}
\end{equation}
If an implementation uses pairwise environment penalties, it differs from \cref{eq:app-env-dispersions} only by a constant depending on the number of environments; we absorb that constant into the corresponding $\lambda$. The quantities $\Delta a_2$ and $\Delta a_4$ below are reported under the convention of \cref{eq:app-fingerprint-definition}.

\subsection{ERM and explicit \texorpdfstring{$R_2/R_4$}{R2/R4} regularizers}

The environment-averaged \ERM objective is exactly \cref{eq:app-reference-objective}:
\begin{equation}
 \F_{\rm ERM}^{(2)}=-\bar c\,w\theta,\qquad
 \F_{\rm ERM}^{(4)}=\frac{\bar q}{2}w^2\theta^2.
 \label{eq:app-erm-components}
\end{equation}
Thus it defines the mixed task curvature and quartic data stabilization at zero regularization, without adding an independent \Rtwo{} or \Rfour{} operator.

The explicit quadratic and quartic environment regularizers are
\begin{equation}
 \Omega_{R_2}=\frac{\lambda_2\gamma}{2}w^2,\qquad
 \Omega_{R_4}=\frac{\lambda_4\gamma}{2}(w\theta)^2.
 \label{eq:app-explicit-operators}
\end{equation}
Substituting $w=\rho v_w$ and $\theta=\rho v_\theta$ gives
\begin{equation}
 \Delta a_2^{R_2}=\lambda_2\gamma v_w^2,\qquad
 \Delta a_4^{R_4}=2\lambda_4\gamma(v_wv_\theta)^2.
 \label{eq:app-explicit-fingerprints}
\end{equation}
For $R_2+R_4$, the two corrections add directly. This also shows that even when both terms appear in the learning objective, only the former changes the Hessian at the origin.

\subsection{MMD-mean, Gaussian MMD, and CORAL}

Here MMD-mean denotes MMD with a linear kernel, equivalently squared mean matching in representation space. For mean matching, if the representation is $h=wX$, the difference in environment means is $\E_e[h]-\overline{\E[h]}=w(m_e-\bar m)$. Hence the local MMD-mean term is
\begin{equation}
 \Omega_{\rm MMD\text{-}mean}=\lambda_\mu M_2w^2,\qquad
 \Delta a_2=2\lambda_\mu M_2v_w^2.
 \label{eq:app-mmd-mean}
\end{equation}
This is a genuine gate-block \Rtwo{} mass. By contrast, the centered covariance satisfies
\begin{equation}
 \Cov_{P_e}(h)=w^2s_e,\qquad
 \Omega_{\rm CORAL\text{-}cov}=\lambda_\Sigma S_2w^4,\qquad
 \Delta a_4=4\lambda_\Sigma S_2v_w^4.
 \label{eq:app-coral-cov}
\end{equation}
Thus covariance-only CORAL begins at quartic order. Full CORAL is the sum of the two: its \Rtwo{} component comes from mean matching and its \Rfour{} component from covariance matching.

For Gaussian MMD between two environment distributions $P,Q$, let $h=tX$ and consider the single-bandwidth kernel
\begin{equation}
 k_\sigma(tX,tX')=\exp\!\left[-\frac{t^2(X-X')^2}{2\sigma^2}\right].
 \label{eq:app-gaussian-kernel}
\end{equation}
Define
\begin{equation}
 C_{2j}(P,Q)=
 \E_{P,P}(X-X')^{2j}+\E_{Q,Q}(X-X')^{2j}
 -2\E_{P,Q}(X-X')^{2j}.
 \label{eq:app-gaussian-mmd-moment}
\end{equation}
Expanding the exponential term by term gives the exact even-power series
\begin{equation}
 {\rm MMD}^2_\sigma(tP,tQ)
 =\sum_{j\geq1}\frac{(-1)^jt^{2j}}{j!(2\sigma^2)^j}C_{2j}(P,Q)
 =A_2t^2+A_4t^4+O(t^6).
 \label{eq:app-gaussian-mmd-series}
\end{equation}
Averaging over environment pairs and kernel bandwidths preserves this form, with $A_2$ and $A_4$ averaged accordingly. Therefore,
\begin{equation}
 \Omega_{\rm MMD\text{-}Gaussian}
 =\lambda_G(A_2w^2+A_4w^4+O(w^6)),\quad
 \Delta a_2=2\lambda_GA_2v_w^2,\quad
 \Delta a_4=4\lambda_GA_4v_w^4.
 \label{eq:app-gaussian-mmd-fingerprint}
\end{equation}
When environment means differ, typically $A_2>0$, so the leading term is \Rtwo{}. If the mean term vanishes by symmetry, the leading term may instead be \Rfour{} or higher order. This is why Gaussian MMD cannot be assigned a data-independent fixed Landau order.

\subsection{IRMv1, V-REx, and GroupDRO}

\paragraph{IRMv1.}
Let the classifier scaling be $\alpha$. By \cref{eq:app-env-risk}, at $\alpha=1$,
\begin{equation}
 \partial_\alpha\ell_e(\alpha\beta)\big|_{\alpha=1}
 =-c_e\beta+q_e\beta^2.
 \label{eq:app-irm-gradient}
\end{equation}
Therefore, under the environment-average convention, the IRMv1 penalty satisfies exactly
\begin{equation}
 \begin{aligned}
 \Omega_{\rm IRM}
 &=\lambda_{\rm IRM}\overline{
 \bigl(-c_e\beta+q_e\beta^2\bigr)^2}\\
 &=\lambda_{\rm IRM}\left[
 \overline{c_e^2}\beta^2
 -2\overline{c_eq_e}\beta^3
 +\overline{q_e^2}\beta^4\right],\\
 \Delta a_4^{\rm IRM}
 &=4\lambda_{\rm IRM}\overline{c_e^2}(v_wv_\theta)^2.
 \end{aligned}
 \label{eq:app-irm-expansion}
\end{equation}
Because $\beta=w\theta=O(\norm{(w,\theta)}^2)$, the three terms are of fourth, sixth, and eighth order, respectively. IRMv1 has no independent \Rtwo{} term in $(w,\theta)$; using a sum rather than an average merely rescales $\lambda_{\rm IRM}$.

\paragraph{V-REx.}
Treat the environmental variation of $L_{0,e}$ as the constant term. Direct expansion of \cref{eq:app-env-risk} gives
\begin{equation}
 \begin{aligned}
 \Var_e(\ell_e)
={}&\Var_e(L_{0,e})-2\Cov_e(L_{0,e},c_e)\beta\\
&+\bigl[\Var_e(c_e)+\Cov_e(L_{0,e},q_e)\bigr]\beta^2\\
&-\Cov_e(c_e,q_e)\beta^3+\frac14\Var_e(q_e)\beta^4.
 \end{aligned}
 \label{eq:app-vrex-risk-variance}
\end{equation}
Hence the genuine \Rtwo{} fingerprint of $\lambda_{\rm V\text{-}REx}\Var_e(\ell_e)$ is
\begin{equation}
 \Delta a_2^{\rm V\text{-}REx}
 =-4\lambda_{\rm V\text{-}REx}\Cov_e(L_{0,e},c_e)v_wv_\theta.
 \label{eq:app-vrex-fingerprint}
\end{equation}
Under the equal-null-risk reference used in Figure~5, this term and $\Cov_e(L_{0,e},q_e)$ both vanish, so
\begin{equation}
 \Delta a_2^{\rm V\text{-}REx}=0,\qquad
 \Delta a_4^{\rm V\text{-}REx}
 =4\lambda_{\rm V\text{-}REx}V_c(v_wv_\theta)^2.
 \label{eq:app-vrex-equal-null}
\end{equation}
V-REx is therefore \Rfour{}-leading under this equal-null-risk reference.

\paragraph{GroupDRO.}
GroupDRO reweights environment-specific task risks toward high-loss environments. With equal weight decay $\mu$ on both parameter blocks, its hard objective is
\begin{equation}
 \F_{\rm GDRO}=\max_{p\in\Delta}\sum_ep_e\ell_e+\frac\mu2(w^2+\theta^2).
 \label{eq:app-hard-groupdro}
\end{equation}
During training, $p_{e,t+1}\propto p_{e,t}\exp[\eta\ell_e(w_t,\theta_t)]$, normalized over environments, with update step size $\eta$. Holding the current weights $p$ fixed during a predictor update gives
\begin{equation}
 \F_p=L_{0,p}-c_pw\theta+\frac{q_p}{2}w^2\theta^2+\frac\mu2(w^2+\theta^2),
 \label{eq:app-groupdro-active}
\end{equation}
where $c_p=\sum_ep_ec_e$ and $q_p=\sum_ep_eq_e$. Relative to uniform \ERM, $\Delta c_p=c_p-\bar c$ and $\Delta q_p=q_p-\bar q$ renormalize the \Rtwo{} task drive and \Rfour{} stiffness, with joint-variable coefficients
\begin{equation}
 \Delta a_2^{\rm GroupDRO}
=-2\Delta c_pv_wv_\theta,\qquad
\Delta a_4^{\rm GroupDRO}
=2\Delta q_p(v_wv_\theta)^2.
 \label{eq:app-groupdro-active-fingerprint}
\end{equation}
At a stable equilibrium with limiting weights $p^\star$ and $q_{p^\star}>0$, \cref{eq:app-loading} gives
\begin{equation}
 |\beta_{p^\star}^\star|
 =\frac{[|c_{p^\star}|-\mu]_+}{q_{p^\star}}.
 \label{eq:app-groupdro-loading}
\end{equation}
Reweighting can reduce shortcut loading while a residual task drive $|c_{p^\star}|>\mu$ sustains a nonzero plateau.

\subsection{IGA, Fish/MLDG, and Fishr}
\label{app:gradient-objectives}

Let $z=(w,\theta)^\top$, and write each environment risk as
\begin{equation}
 \ell_e=L_{0,e}+\frac12z^\top H_ez+r_e^{(4)}(z),\qquad
 H_e=\begin{pmatrix}0&-c_e\\-c_e&0\end{pmatrix},\qquad
 r_e^{(4)}(z)=\frac{q_e}{2}w^2\theta^2 .
 \label{eq:app-gradient-normal-form}
\end{equation}

\paragraph{IGA.}
IGA penalizes dispersion among environment gradients. Let $\Delta H_e=H_e-\bar H$ and $\Delta r_e^{(4)}=r_e^{(4)}-\bar r^{(4)}$. Then
\begin{equation}
 \begin{aligned}
 \Omega_{\rm IGA}
 &=\lambda_{\rm IGA}\overline{\left\lVert
 \nabla\ell_e-\overline{\nabla\ell_e}\right\rVert^2}\\
 &=\lambda_{\rm IGA}z^\top\overline{\Delta H_e^\top\Delta H_e}z
 +2\lambda_{\rm IGA}\overline{\left\langle
 \Delta H_ez,\nabla\Delta r_e^{(4)}(z)\right\rangle}
 +O(\norm z^6).
 \end{aligned}
 \label{eq:app-iga-matrix-expansion}
\end{equation}
Under the Hessian convention $\frac12z^\top Kz$,
\begin{equation}
 K_{\rm IGA}=2\lambda_{\rm IGA}
 \overline{\Delta H_e^\top\Delta H_e}\succeq0.
 \label{eq:app-iga-matrix-mass}
\end{equation}
Thus IGA contributes a positive-semidefinite matrix \Rtwo{} over the full parameter space and can rotate the unstable direction. In the scalar probe, \cref{eq:app-gradient-normal-form} gives
\begin{equation}
 \begin{aligned}
 \nabla_w\ell_e-\overline{\nabla_w\ell_e}
 &=-\Delta c_e\,\theta+\Delta q_e\,w\theta^2,\\
 \nabla_\theta\ell_e-\overline{\nabla_\theta\ell_e}
 &=-\Delta c_e\,w+\Delta q_e\,w^2\theta.
 \end{aligned}
 \label{eq:app-iga-gradients}
\end{equation}
\begin{equation}
\Omega_{\rm IGA}=\lambda_{\rm IGA}V_c(w^2+\theta^2)+O(\norm{(w,\theta)}^4),
\quad
\Delta a_2^{\rm IGA}=2\lambda_{\rm IGA}V_c(v_w^2+v_\theta^2).
\label{eq:app-iga-fingerprint}
\end{equation}
For the canonical squared-loss probe, this statement can be made explicit: substituting \cref{eq:app-iga-gradients} and collecting powers gives the exact identity
\begin{equation}
 \Omega_{\rm IGA}
 =\lambda_{\rm IGA}\left[
 V_c(w^2+\theta^2)
 -2C_{cq}w\theta(w^2+\theta^2)
 +V_qw^2\theta^2(w^2+\theta^2)
 \right].
 \label{eq:app-iga-scalar-full}
\end{equation}
The second term is the generally signed joint \Rfour{} correction, and the last is a nonnegative sixth-order correction. Thus IGA is \Rtwo{}-leading with quartic and sixth-order corrections. These higher-order terms preserve the strict local onset and can deform the finite-amplitude path as $\lambda_{\rm IGA}$ grows.

\paragraph{Fish/MLDG.}
The Fish/MLDG implementation in the figure uses cross-environment gradient inner products,
\begin{equation}
 \Omega_{\rm Fish/MLDG}
 =-\lambda_{\rm Fish}\overline{\left\langle\nabla\ell_e,\nabla\ell_f\right\rangle}_{e\ne f}.
 \label{eq:app-fish-objective}
\end{equation}
Substituting \cref{eq:app-gradient-normal-form} gives
\begin{equation}
 \begin{aligned}
 \Omega_{\rm Fish/MLDG}^{(2)}
 &=-\lambda_{\rm Fish}
 \overline{z^\top H_e^\top H_fz}_{e\ne f},\\
 \Omega_{\rm Fish/MLDG}^{(4)}
 &=-\lambda_{\rm Fish}\overline{
 \left\langle H_ez,\nabla r_f^{(4)}(z)\right\rangle+
 \left\langle H_fz,\nabla r_e^{(4)}(z)\right\rangle}_{e\ne f}.
 \end{aligned}
 \label{eq:app-fish-expansion}
\end{equation}
Because a scalar quadratic form depends only on the symmetric part of its matrix, define
\begin{equation}
 \begin{aligned}
 G_{\rm Fish}
 &:=-2\overline{\operatorname{sym}(H_e^\top H_f)}_{e\ne f},
 \qquad
 \operatorname{sym}(A)=\frac12(A+A^\top),\\
 \Phi_{\rm Fish}^{(4)}(z)
 &:=-\overline{
 \left\langle H_ez,\nabla r_f^{(4)}(z)\right\rangle+
 \left\langle H_fz,\nabla r_e^{(4)}(z)\right\rangle}_{e\ne f}.
 \end{aligned}
 \label{eq:app-fish-G-Phi}
\end{equation}
The low-order Fish/MLDG correction can therefore be written as
\begin{equation}
 \Omega_{\rm Fish/MLDG}(z)
 =\frac{\lambda_{\rm Fish}}{2}z^\top G_{\rm Fish}z
 +\lambda_{\rm Fish}\Phi_{\rm Fish}^{(4)}(z)
 +O(\norm z^6).
 \label{eq:app-fish-normal-form}
\end{equation}
The quadratic and quartic corrections have signs and directions determined by the geometry of cross-environment gradients; both stabilizing and destabilizing contributions are possible.

In the scalar probe, $H_e^\top H_f=c_ec_fI$, so
\begin{equation}
 G_{\rm Fish}
 =-2\overline{c_ec_f}_{e\ne f}I,\qquad
 \Delta a_2^{\rm Fish/MLDG}
 =-2\lambda_{\rm Fish}\overline{c_ec_f}_{e\ne f}
 (v_w^2+v_\theta^2).
 \label{eq:app-fish-fingerprint}
\end{equation}
The correction is then isotropic in the two-dimensional probe, but its sign is still determined by products of task gradients across environments. In a high-dimensional parameter space, projections along different feature directions can have different signs, producing positive-mass suppression in some directions and negative-mass amplification in others.

The canonical scalar probe also makes the first omitted term explicit. Define the ordered-pair averages
\begin{equation}
 A_{cc}=\overline{c_ec_f}_{e\ne f},\qquad
 A_{cq}=\overline{c_eq_f+q_ec_f}_{e\ne f},\qquad
 A_{qq}=\overline{q_eq_f}_{e\ne f}.
 \label{eq:app-fish-scalar-coefficients}
\end{equation}
The complete squared-loss correction is
\begin{equation}
 \Omega_{\rm Fish/MLDG}
 =-\lambda_{\rm Fish}A_{cc}(w^2+\theta^2)
 +\lambda_{\rm Fish}A_{cq}w\theta(w^2+\theta^2)
 -\lambda_{\rm Fish}A_{qq}w^2\theta^2(w^2+\theta^2).
 \label{eq:app-fish-scalar-full}
\end{equation}
Hence Fish/MLDG has an explicit sixth-order correction in addition to its sign-indefinite quadratic and quartic terms. Since $q_e=\E_e[X^2]\geq0$, $A_{qq}\geq0$ and this sixth-order contribution is nonpositive for positive $\lambda_{\rm Fish}$.

\begin{theorem}[Finite-strength local instability of Fish/MLDG]
\label{thm:app-fish-instability}
Translate any locally stable reference point to the origin, and let the Hessian without the Fish/MLDG correction be
$H_{\rm ref}\succ0$. If $G_{\rm Fish}$ has a negative-curvature direction, define
\begin{equation}
 \lambda_{\rm inst}
 =
 \min_{\substack{\norm v=1\\v^\top G_{\rm Fish}v<0}}
 \frac{v^\top H_{\rm ref}v}
 {-v^\top G_{\rm Fish}v}.
 \label{eq:app-fish-instability-threshold}
\end{equation}
Then $0<\lambda_{\rm inst}<\infty$, and
\begin{equation}
 H_{\rm ref}+\lambda_{\rm Fish}G_{\rm Fish}
 \begin{cases}
 \succ0, & 0\leq\lambda_{\rm Fish}<\lambda_{\rm inst},\\
 \text{has a zero eigenvalue}, & \lambda_{\rm Fish}=\lambda_{\rm inst},\\
 \text{has a negative eigenvalue}, & \lambda_{\rm Fish}>\lambda_{\rm inst}.
 \end{cases}
 \label{eq:app-fish-hessian-phases}
\end{equation}
In particular, when $H_{\rm ref}=\mu I$,
\begin{equation}
 \lambda_{\rm inst}
 =\frac{\mu}{\left|\lambda_{\min}(G_{\rm Fish})\right|}.
 \label{eq:app-fish-isotropic-threshold}
\end{equation}
\end{theorem}

\begin{proof}
Along any unit direction $v$, the total quadratic curvature is
\begin{equation}
 \kappa_2(v;\lambda_{\rm Fish})
 =v^\top H_{\rm ref}v
 +\lambda_{\rm Fish}v^\top G_{\rm Fish}v.
 \label{eq:app-fish-directional-curvature}
\end{equation}
If $v^\top G_{\rm Fish}v\geq0$, increasing
$\lambda_{\rm Fish}$ cannot destabilize that direction. If $v^\top G_{\rm Fish}v<0$,
$\kappa_2$ crosses zero exactly at
$v^\top H_{\rm ref}v/[-v^\top G_{\rm Fish}v]$.
Taking the minimum over all negative-curvature directions gives
\cref{eq:app-fish-instability-threshold}. Because
$H_{\rm ref}\succ0$ and the set of negative-curvature directions is nonempty, the threshold is strictly positive and finite.
The Rayleigh--Ritz criterion then gives
\cref{eq:app-fish-hessian-phases}. When $H_{\rm ref}=\mu I$, the first unstable direction is the minimum-eigenvalue eigenvector of $G_{\rm Fish}$, yielding
\cref{eq:app-fish-isotropic-threshold}.
\end{proof}

This theorem also explains how ``\Rtwo-like behavior at small regularization, followed by sudden instability'' can coexist. If the shortcut direction $e_s$ satisfies
$e_s^\top G_{\rm Fish}e_s>0$, then its effective mass
\begin{equation}
 m_s(\lambda_{\rm Fish})
 =e_s^\top H_{\rm ref}e_s
 +\lambda_{\rm Fish}e_s^\top G_{\rm Fish}e_s
 \label{eq:app-fish-shortcut-mass}
\end{equation}
increases with regularization and therefore behaves as a positive \Rtwo{} term under this projection. However, if another direction satisfies
$v^\top G_{\rm Fish}v<0$, the full system still loses local stability at
$\lambda_{\rm inst}$.

\begin{proposition}[Quartic bifurcation after instability]
\label{prop:app-fish-quartic}
Along a unit direction $v$, write the total local free energy as
\begin{equation}
 \F_\lambda(\rho v)-\F_\lambda(0)
 =\frac12\kappa_2(\lambda)\rho^2
 +\frac14\kappa_4(\lambda)\rho^4
 +O(\rho^6).
 \label{eq:app-fish-ray-normal-form}
\end{equation}
If $\kappa_2(\lambda)<0$, then:
\begin{enumerate}[leftmargin=1.6em]
 \item If $\kappa_4(\lambda)>0$, the quartic truncation produces a stable nonzero branch
 $\rho_\star^2=-\kappa_2/\kappa_4$, corresponding to mode amplification, switching, or a new metastable state;
 \item If $\kappa_4(\lambda)\leq0$, the quartic truncation has no stable nonzero well, and the radial gradient flow escapes outward.
\end{enumerate}
\end{proposition}

\begin{proof}
Neglecting $O(\rho^6)$,
$\partial_\rho\F_\lambda=\rho[\kappa_2+\kappa_4\rho^2]$.
When $\kappa_2<0<\kappa_4$, in addition to the unstable origin there is a stationary point at
$\rho_\star^2=-\kappa_2/\kappa_4$, whose second derivative is
$-2\kappa_2>0$. When $\kappa_4\leq0$, for every $\rho>0$,
$\kappa_2+\kappa_4\rho^2<0$, and hence
$-\partial_\rho\F_\lambda>0$; the gradient flow of the quartic truncation therefore moves continuously toward larger amplitude.
\end{proof}

\begin{remark}[Theoretical instability versus numerical collapse]
\Cref{thm:app-fish-instability} proves that the local Hessian loses positive definiteness at a finite regularization strength. Positive higher-order terms may restabilize the system farther away, in which case one observes a large mode rearrangement or a nonmonotone path. If the quartic and accessible higher-order terms fail to provide a stable well, optimization rapidly escapes along a negative-curvature direction and appears as parameter explosion or numerical collapse under finite-precision training. In the main experiments, the Fish/MLDG path is truncated at the first observed numerical instability.
\end{remark}

\paragraph{Fishr.}
Standard Fishr matches the coordinate-wise variances of per-example \emph{classifier} gradients. Let
\begin{equation}
 g^{\rm cls}_{e,n}=\nabla_{\theta_{\rm cls}}\ell_{e,n},
 \qquad
 v_e=\Var_{n\in e}\!\left(g^{\rm cls}_{e,n}\right),
 \label{eq:app-fishr-definition}
\end{equation}
where the variance is taken coordinate-wise. We study the instantaneous static objective
\begin{equation}
 \Omega_{\rm Fishr}
 =\frac{\lambda_{\rm Fishr}}{d_{\rm cls}}
 \sum_{j=1}^{d_{\rm cls}}\Var_e\!\left[(v_e)_j\right].
 \label{eq:app-fishr-static-objective}
\end{equation}
For a controlled comparison of objective geometries, no EMA, penalty annealing, or warm-up is used. In the scalar bilinear probe, take $\theta$ to be the classifier parameter. Then
\begin{equation}
 \nabla_\theta\ell_{e,n}=[(\beta X_n-Y_n)X_n]w,
 \qquad
 \nu_e(\beta)=\Var_{n\in e}[(\beta X_n-Y_n)X_n].
 \label{eq:app-fishr-scalar-gradient}
\end{equation}
For squared loss, define the environment-wise coefficients
\begin{equation}
 \nu_{0,e}=\Var_{P_e}(YX),\qquad
 \nu_{1,e}=-2\Cov_{P_e}(YX,X^2),\qquad
 \nu_{2,e}=\Var_{P_e}(X^2),
 \label{eq:app-fishr-nu-coefficients}
\end{equation}
so that $\nu_e(\beta)=\nu_{0,e}+\nu_{1,e}\beta+\nu_{2,e}\beta^2$ exactly. Let $V^{\rm Fishr}_{ij}=\Cov_e(\nu_{i,e},\nu_{j,e})$. Hence $v_e=\nu_e(\beta)w^2$ and the exact static scalar penalty is
\begin{equation}
 \begin{aligned}
 \Omega_{\rm Fishr}
 ={}&\lambda_{\rm Fishr}\bigl[
 V^{\rm Fishr}_{00}w^4
 +2V^{\rm Fishr}_{01}w^5\theta
 +(V^{\rm Fishr}_{11}+2V^{\rm Fishr}_{02})w^6\theta^2\\
 &\hspace{19mm}{}+2V^{\rm Fishr}_{12}w^7\theta^3
 +V^{\rm Fishr}_{22}w^8\theta^4
 \bigr].
 \end{aligned}
 \label{eq:app-fishr-fingerprint}
\end{equation}
Thus standard Fishr has no independent \Rtwo{} contribution and is locally \Rfour{}-leading. Its successive terms have joint orders $4,6,8,10,12$; only the leading $V^{\rm Fishr}_{00}\geq0$ and final $V^{\rm Fishr}_{22}\geq0$ coefficients have a fixed sign. Fishr and covariance-only CORAL are both quartic geometry-matching objectives, but Fishr matches classifier-gradient variances whereas CORAL matches representation covariances.

\subsection{Mixup and a DANN surrogate}
\label{app:dann-profile}

\paragraph{Mixup.}
For an environment pair $(e,f)$ and mixing coefficient $\alpha$, define
\begin{equation}
 \widetilde X=\alpha X_e+(1-\alpha)X_f,\qquad
 \widetilde Y=\alpha Y_e+(1-\alpha)Y_f .
 \label{eq:app-mixup-samples}
\end{equation}
Let $\widetilde c=\E[\widetilde X\widetilde Y]$ and $\widetilde q=\E[\widetilde X^2]$, where the expectation also averages over the sampled $(e,f,\alpha)$. The mixed risk remains exactly
\begin{equation}
 \F_{\rm Mixup}
=\widetilde L_0-\widetilde c\,w\theta
+\frac{\widetilde q}{2}w^2\theta^2+\text{weight decay}.
 \label{eq:app-mixup-risk}
\end{equation}
Mixup renormalizes the data moments of the task loss:
\begin{equation}
 \Delta a_2^{\rm Mixup}=-2(\widetilde c-\bar c)v_wv_\theta,\qquad
 \Delta a_4^{\rm Mixup}=2(\widetilde q-\bar q)(v_wv_\theta)^2.
 \label{eq:app-mixup-fingerprint}
\end{equation}

\raggedbottom
\paragraph{DANN.}
To derive a comparable local expression, first consider a ridge-regularized linear environment head $B$ that predicts the centered environment encoding $s$:
\begin{equation}
 \begin{aligned}
 L_D(g,B)&=\frac12\E\norm{s-BD_gX}^2+\frac{\mu_B}{2}\norm B_F^2,\\
 D_g&=\diag(g),\qquad C_s=\E[Xs^\top],\qquad\Sigma=\E[XX^\top].
 \end{aligned}
 \label{eq:app-dann-linear-head}
\end{equation}
Minimizing over $B$ yields the profiled loss
\begin{equation}
 L_D^\star(g)=L_D^\star(0)
 -\frac12\operatorname{Tr}\!\left[
 C_s^\top D_g(D_g\Sigma D_g+\mu_BI)^{-1}D_gC_s\right].
 \label{eq:app-dann-profile}
\end{equation}
The feature-side DANN term is $-\lambda_D L_D^\star(g)$. Expanding the inverse matrix at $g=0$ gives
\begin{equation}
 \begin{aligned}
 \Omega_{\rm DANN}^{(2)}
 &=\frac{\lambda_D}{2\mu_B}
 \operatorname{Tr}(C_s^\top D_g^2C_s),\\
 \Omega_{\rm DANN}^{(4)}
 &=-\frac{\lambda_D}{2\mu_B^2}
 \operatorname{Tr}(C_s^\top D_g^2\Sigma D_g^2C_s).
 \end{aligned}
 \label{eq:app-dann-matrix-expansion}
\end{equation}
The first term is a positive-semidefinite gate mass along environment-predictive directions, whereas the second is a saturating negative quartic correction. For a single feature and scalar environment encoding, $C_s=c_D:=\E[Xs]$ and $\Sigma=\bar q$ give the exact geometric series
\begin{equation}
 \Omega_{\rm DANN}^{\rm loc}
 =\frac{\lambda_Dc_D^2}{2\mu_B}w^2
 -\frac{\lambda_Dc_D^2\bar q}{2\mu_B^2}w^4
 +\frac{\lambda_Dc_D^2\bar q^2}{2\mu_B^3}w^6
 +O(w^8),
 \label{eq:app-dann-expansion}
\end{equation}
and hence
\begin{equation}
 \Delta a_2^{\rm DANN}=\lambda_D\frac{c_D^2}{\mu_B}v_w^2,\qquad
 \Delta a_4^{\rm DANN}=-2\lambda_D\frac{c_D^2\bar q}{\mu_B^2}v_w^4.
 \label{eq:app-dann-fingerprint}
\end{equation}
The expansion above applies to the ridge-profiled linear MSE surrogate. The experiments use a linear environment classifier trained by alternating cross-entropy updates.

\subsection{Summary of local operators}

\begin{table}[H]
\centering
\caption{Effective local expansions at the controlled bilinear reference state. Here $z=(w,\theta)$, $\beta=w\theta$, and $v$ denotes a measurement direction. Terms denoted $O(\norm z^k)$ are the remaining higher-order terms; every ``quartic'' statement refers to order in the joint variables $(w,\theta)$.}
\label{tab:app-operators}
\scriptsize
\setlength{\tabcolsep}{4pt}
\begin{tabular}{@{}>{\raggedright\arraybackslash}p{0.16\linewidth}>{\raggedright\arraybackslash}p{0.43\linewidth}>{\raggedright\arraybackslash}p{0.365\linewidth}@{}}
\toprule
\textbf{Method in figure} & \textbf{Effective correction relative to ERM} & \textbf{Local order, sign, and conditions}\\
\midrule
ERM & None: $-\bar c\,\beta+\tfrac{\bar q}{2}\beta^2$ & Reference mixed task curvature and quartic data term; see \cref{eq:app-erm-components}.\\
GroupDRO & $-\Delta c_p\beta+\tfrac12\Delta q_p\beta^2$ & Reweighted \Rtwo{}/\Rfour{} task terms at fixed weights; residual drive can sustain a nonzero plateau.\\
IRMv1 & $\lambda_{\rm IRM}[\overline{c_e^2}\beta^2-2\overline{c_eq_e}\beta^3+\overline{q_e^2}\beta^4]$ & \Rfour{}-leading, with explicit sixth- and eighth-order terms; no \Rtwo{} contribution.\\
V-REx & $-2\Cov(L_{0,e},c_e)\beta+[\Var(c_e)+\Cov(L_{0,e},q_e)]\beta^2-\Cov(c_e,q_e)\beta^3+\tfrac14\Var(q_e)\beta^4$ & \Rtwo{} under unequal null risks; \Rfour{}-leading under equal null risks, with sixth- and eighth-order terms.\\
IGA & $\lambda z^\top\overline{\Delta H_e^\top\Delta H_e}z+2\lambda\overline{\langle\Delta H_ez,\nabla\Delta r_e^{(4)}(z)\rangle}+O(\norm z^6)$ & PSD matrix-\Rtwo{} plus generally signed \Rfour{} and nonnegative $R_6$ gradient couplings.\\
Fish/MLDG & $\frac{\lambda}{2}z^\top G_{\rm Fish}z+\lambda\Phi_{\rm Fish}^{(4)}(z)+O(\norm z^6)$ & Quadratic and quartic corrections are sign-indefinite; a negative direction of $G_{\rm Fish}$ causes finite-strength local instability; see \cref{thm:app-fish-instability}.\\ 
Fishr & $\lambda_{\rm Fishr}\Var_e[\nu_e(0)]w^4+O(\norm z^6)$ & Classifier-gradient variance matching; \Rfour{}-leading with state-dependent finite-amplitude higher-order corrections.\\
Mixup & $-(\widetilde c-\bar c)\beta+\tfrac12(\widetilde q-\bar q)\beta^2$ & Renormalizes the task curvature and quartic data term; no fixed additive operator.\\
MMD-mean & $\lambda_\mu M_2w^2$ & Gate-block \Rtwo{}; see \cref{eq:app-mmd-mean}.\\
Gaussian MMD & $\lambda_G(A_2w^2+A_4w^4+\cdots)$ & Determined by moment differences; may begin at \Rtwo{}, \Rfour{}, or higher order.\\
CORAL-cov & $\lambda_\Sigma S_2w^4$ & Quartic gate-covariance geometry.\\
CORAL-full & $\lambda_\mu M_2w^2+\lambda_\Sigma S_2w^4$ & Contains both the MMD-mean \Rtwo{} and the CORAL-cov \Rfour{}.\\
DANN & $+\operatorname{Tr}(C_s^\top D_g^2C_s)-O(g^4)$ & The profiled linear MSE surrogate yields a PSD gate \Rtwo{} and a saturating negative \Rfour{}; experiments use an alternating linear cross-entropy classifier.\\
\Rtwo & $\frac12\lambda_2\gamma w^2$ & Explicit gate \Rtwo{}.\\
\Rfour & $\frac12\lambda_4\gamma(w\theta)^2$ & Explicit joint \Rfour{}.\\
$R_2+R_4$ & $\frac12\lambda_2\gamma w^2+\frac12\lambda_4\gamma(w\theta)^2$ & \Rtwo{} determines the exact onset; \Rfour{} determines the post-onset amplitude.\\
\bottomrule
\end{tabular}
\end{table}

\flushbottom
\section{Supplementary results}
\label{app:supplementary-results}

\subsection{Finite-sample operator alignment}
\label{app:figure5-operator-alignment}

The MMD-mean and CORAL-full objectives in Figure~5 are evaluated from empirical environment moments.  Let $\widehat m_e=n^{-1}\sum_{i\in e}X_i$ and define
\[
 \widehat\Gamma_\mu=
 \frac{2}{\binom{E}{2}}\sum_{e<f}
 \diag\!\left((\widehat m_e-\widehat m_f)^{\odot 2}\right).
\]
For the gate vector $g$, the MMD-mean penalty is $\Omega_{\rm MMD}=\frac{\lambda}{2}g^\top\widehat\Gamma_\mu g$.  CORAL-full has this identical leading \Rtwo operator and adds a covariance term that begins at \Rfour order.  Although the population stable-coordinate mean is invariant by construction, its empirical counterpart fluctuates across finite training environments.  With $\gamma_c=e_c^\top\widehat\Gamma_\mu e_c$ and $\gamma_s=e_s^\top\widehat\Gamma_\mu e_s$, \cref{tab:app-figure5-alignment} audits this projection before training on the five Figure~5 datasets at each sample size.

The stable projection is small but nonzero. At $\lambda=100$, its mean contribution is $\lambda\bar\gamma_c\simeq2.86$ for $n=256$, compared with the gate weight-decay coefficient $\mu=0.2$. For $n=8192$, the corresponding contribution is only $0.0527$, consistent with the weaker collateral suppression in \cref{fig:paths}(b).

\begin{table}[H]
\centering
\scriptsize
\setlength{\tabcolsep}{3pt}
\renewcommand{\arraystretch}{1.08}
\caption{Leading empirical \Rtwo signatures in the finite-sample Figure~5 objective. Values are mean $\pm$ s.e.m. over five pre-training datasets at each sample size $n$ per environment. $\gamma_c$ and $\gamma_s$ are the stable- and shortcut-mode gate masses.}
\label{tab:app-figure5-alignment}
\begin{tabular}{@{}lcccc@{}}
\toprule
\textbf{Objective} & $n$ & $\bar\gamma_c$ & $\bar\gamma_s$ & $\overline{\gamma_c/\gamma_s}$ \\
\midrule
oracle \Rtwo & both & $0$ & $1$ & $0$ \\
MMD-mean & 256 & $0.0286\pm0.0078$ & $5.092\pm0.212$ & $(5.73\pm1.57)\!\times\!10^{-3}$ \\
CORAL-full & 256 & $0.0286\pm0.0078$ & $5.092\pm0.212$ & $(5.73\pm1.57)\!\times\!10^{-3}$ \\
MMD-mean & 8192 & $(5.27\pm0.65)\!\times\!10^{-4}$ & $5.342\pm0.021$ & $(9.88\pm1.24)\!\times\!10^{-5}$ \\
CORAL-full & 8192 & $(5.27\pm0.65)\!\times\!10^{-4}$ & $5.342\pm0.021$ & $(9.88\pm1.24)\!\times\!10^{-5}$ \\
\bottomrule
\end{tabular}
\end{table}

\subsection{Extended nonlinear regularization paths}
\label{app:extended-nonlinear-paths}

\begin{figure}[!htbp]
\centering
\includegraphics[width=\textwidth]{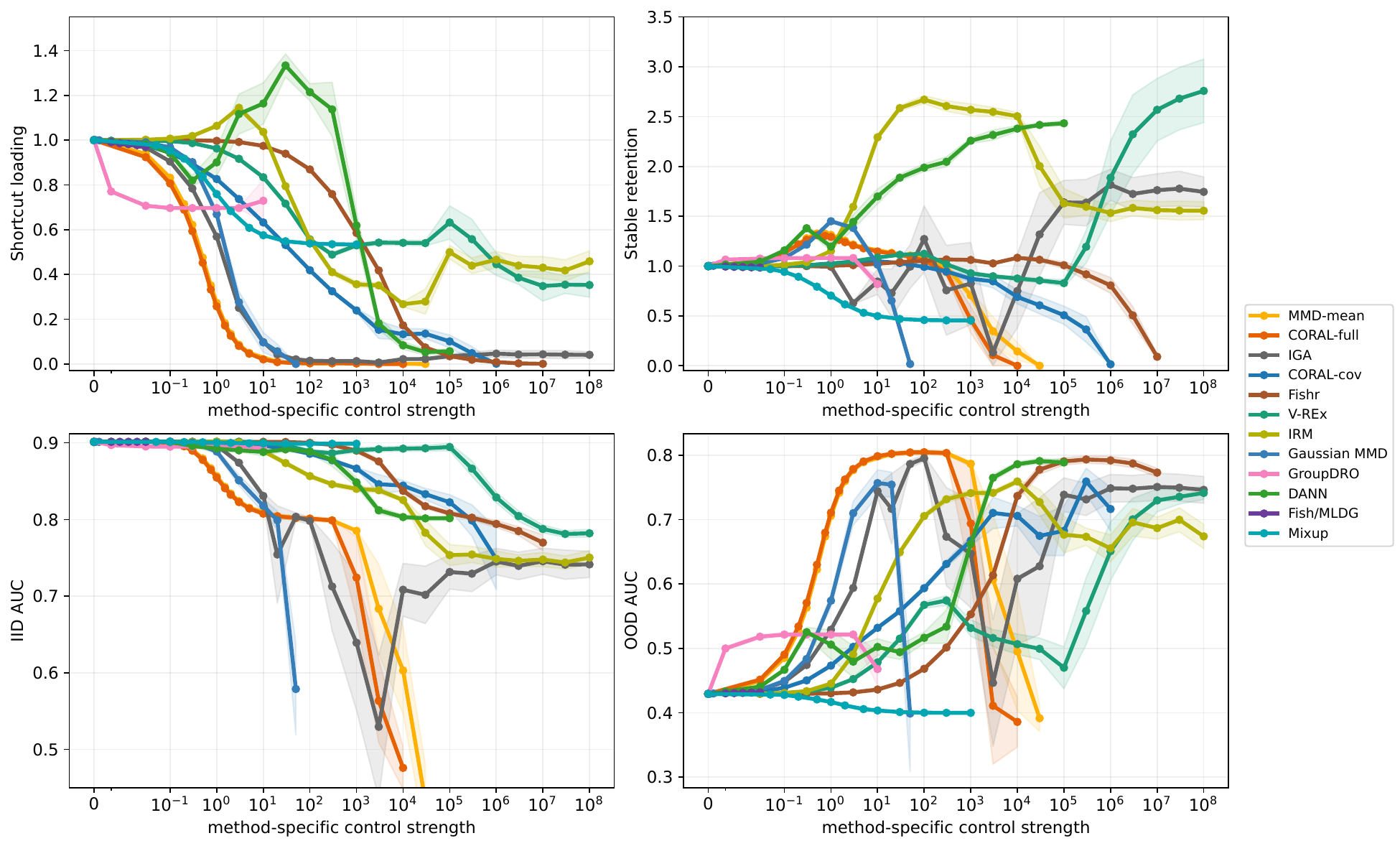}
\caption{Extended regularization paths for the one-hidden-layer ReLU network with 8192 training samples per environment.}
\label{fig:app-relu1-extended}
\end{figure}

\begin{figure}[!htbp]
\centering
\includegraphics[width=\textwidth]{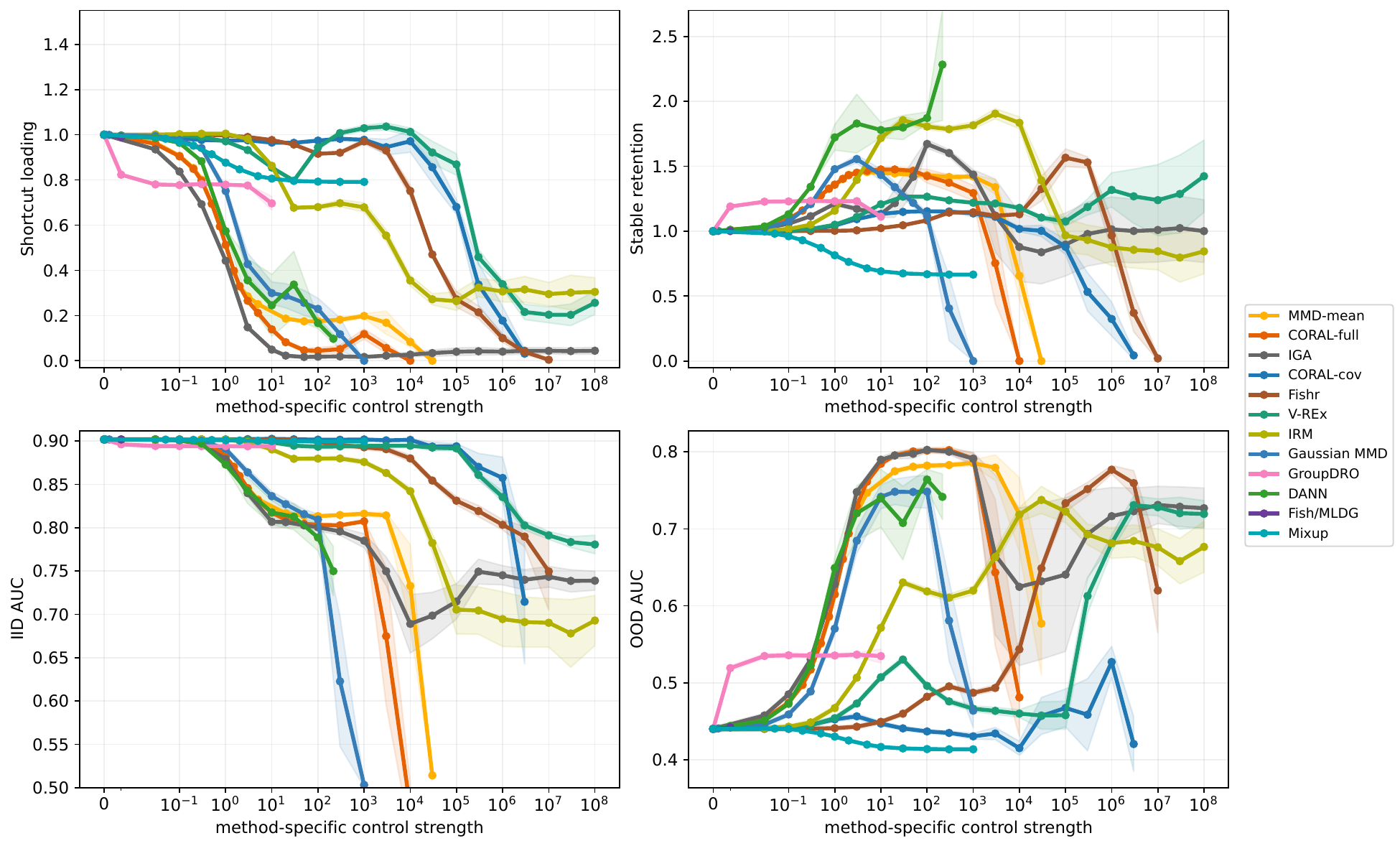}
\caption{Extended regularization paths for the two-hidden-layer ReLU network with 8192 training samples per environment.}
\label{fig:app-relu2-extended}
\end{figure}

\begin{figure}[!htbp]
\centering
\includegraphics[width=\textwidth]{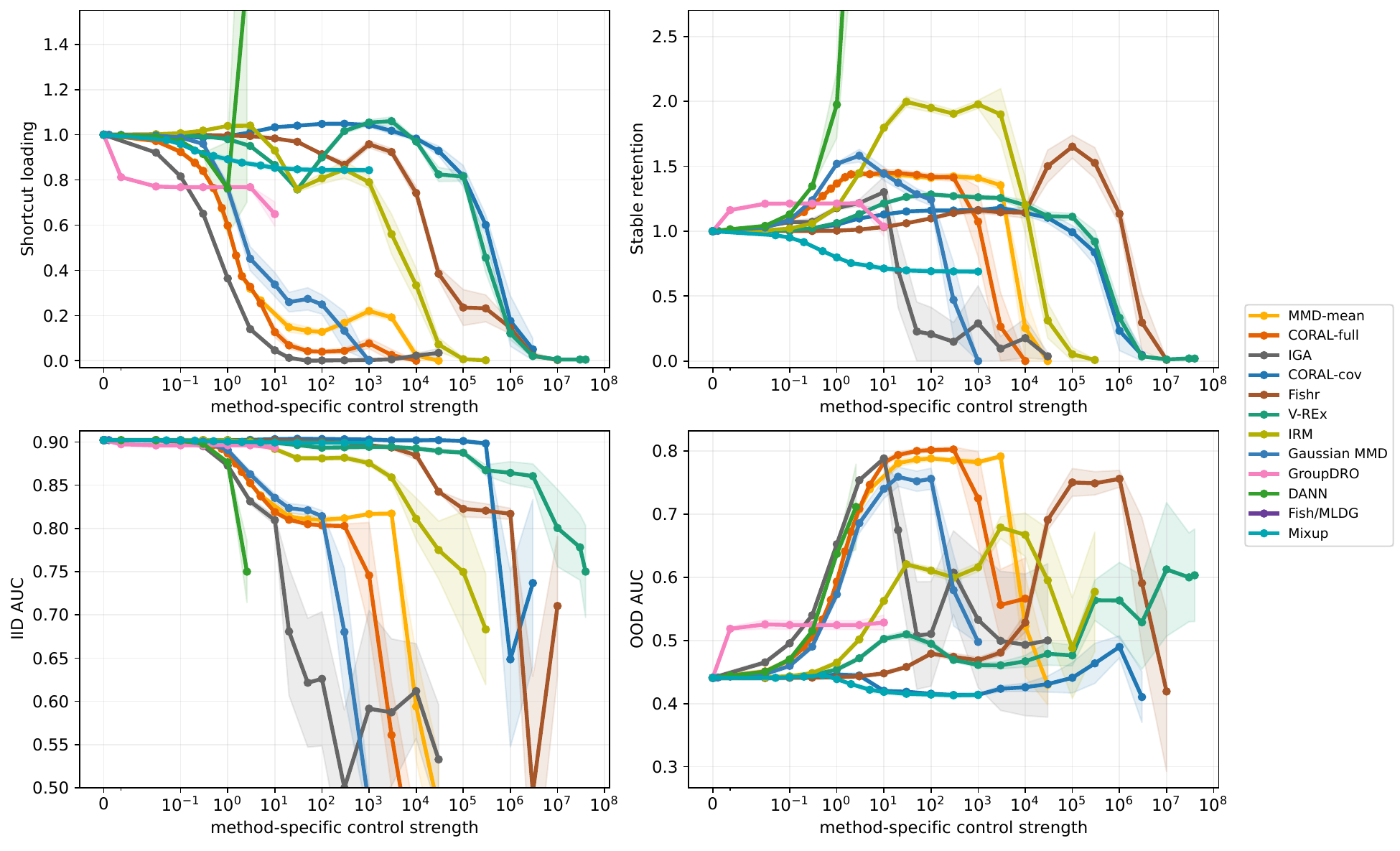}
\caption{Extended regularization paths for the three-hidden-layer ReLU network with 8192 training samples per environment, using the same data as \cref{fig:app-relu1-extended,fig:app-relu2-extended}.}
\label{fig:app-relu3-extended}
\end{figure}

\Cref{fig:app-relu1-extended,fig:app-relu2-extended} extend the same runs shown in \cref{fig:nonlinear-paths} to larger control strengths, and \cref{fig:app-relu3-extended} adds a three-hidden-layer network. All three architectures use exactly the same training and IID/OOD test sets as \cref{fig:paths}(b), with 8192 training samples per environment. The three-hidden-layer network has widths $(32,32,32)$ and uses the two-hidden-layer optimization settings; loading definitions are unchanged. Curves and shaded bands report means and standard errors over five seeds. Each method uses the same displayed endpoint across all four metrics.

\FloatBarrier

\subsection{Regularization paths on MNIST}
\label{app:mnist-paths}

Prior invariant-learning studies have used MNIST constructions with label-correlated corner patches \citep{chen2022scill} or environment-specific white squares \citep{norman2026unsupervised}. We adapt these localized cues to a binary bright marker with controlled environment-dependent prevalence and label association.

We examine whether the qualitative signature--phenotype relation in Figure~6 remains informative when stable features are learned from MNIST images. We evaluate thirteen established methods under two controlled environment allocations.

\paragraph{Task and shortcut construction.}
Following the binary digit task of \citet{arjovsky2019irm}, the clean label is $Y_c=2\mathbf{1}\{d<5\}-1$ for digit $d\in\{0,\ldots,9\}$; the task label $Y\in\{-1,+1\}$ flips $Y_c$ with rate $0.25$, enforced by rounded counts within each digit. The stable signal is therefore the grayscale digit shape. The shortcut is a bright corner marker: when $B=1$, initially zero-valued pixels within an upper-left $8\times8$ patch are set to unit brightness, whereas $B=0$ leaves the background unchanged. This intervention never overwrites nonzero digit pixels. With $S=2B-1$ and training environment index $\zeta_e\in\{-3,-1,1,3\}/\sqrt5$, the label-conditional marker probabilities are chosen to realize the target moments
\begin{equation}
 \E[S\mid e]=\delta \zeta_e,\qquad
 c_e:=\E[SY\mid e]=a_s+\Delta_a \zeta_e,\qquad
 (a_s,\delta,\Delta_a)=(0.70,0.10,0.03).
 \label{eq:app-mnist-moments}
\end{equation}
Equivalently, $P(B=1\mid Y=y,e)=(1+\delta \zeta_e+y c_e)/2$ for balanced $Y$. Conditional marker counts are rounded within the two task-label classes. This separates the marker's marginal mean shift from its label-correlation heterogeneity: $\Var_e(\E[S\mid e])=0.01$, whereas $\Var_e(c_e)=0.0009$. The nominal marker-only AUC ranges from $0.830$ to $0.870$ across training environments, averaging $0.85$. OOD evaluation uses held-out MNIST test images in two environments with $\zeta_e=\pm2$, $\E[S\mid e]=0.10\zeta_e$, and $\E[SY\mid e]=-0.70$, reversing the shortcut association (marker-only AUC $0.15$). IID AUC is measured on a stratified $20\%$ holdout from each training environment; OOD AUC is averaged over the two test environments. Both use the noisy task label.

\edef\mnistsavedtopnumber{\number\value{topnumber}}
\setcounter{topnumber}{1}
\begin{figure}[t]
\centering
\includegraphics[width=\textwidth]{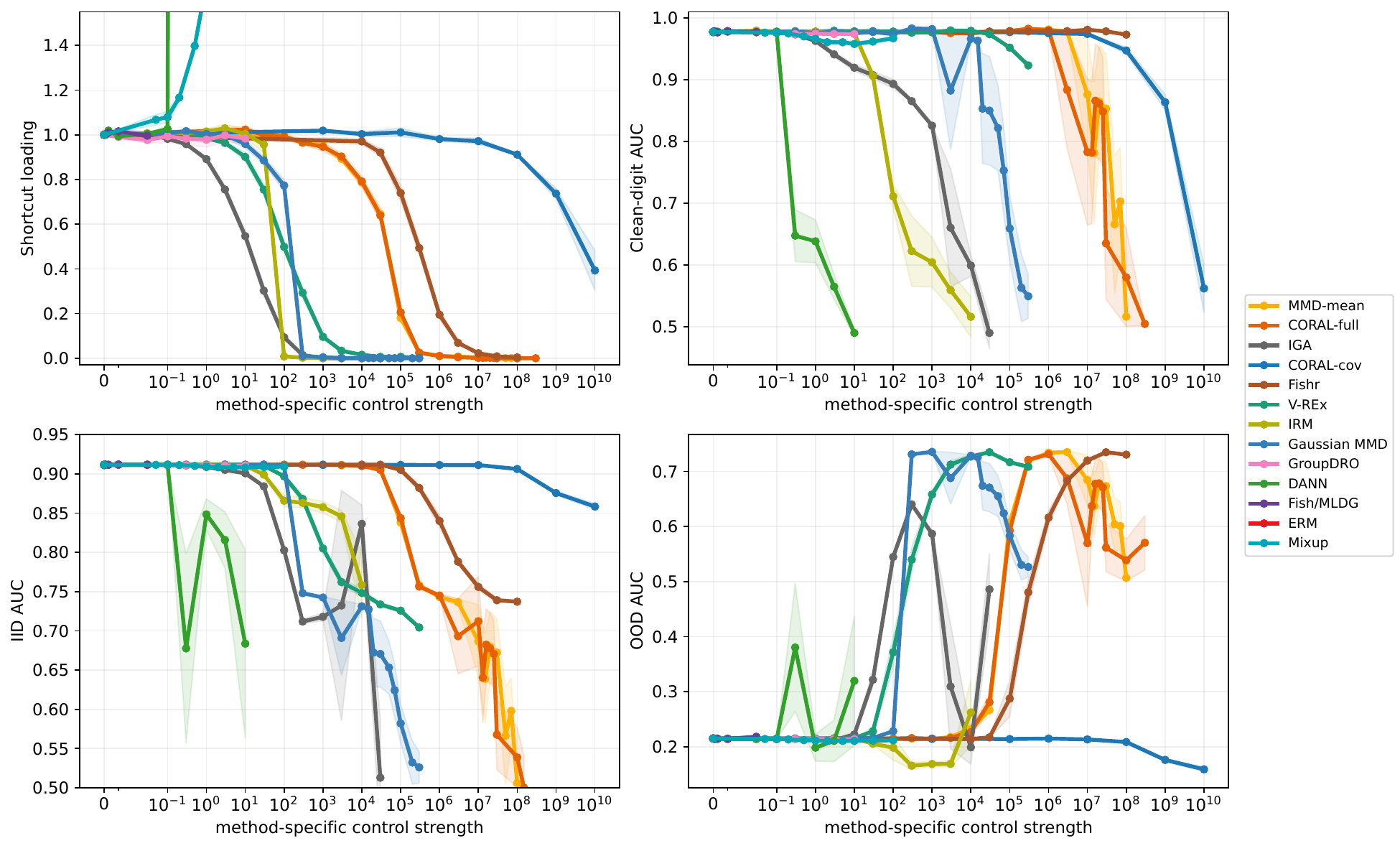}
\caption{MNIST regularization paths with identical digit images and label-flip masks across four training environments. Curves and shaded bands show means and standard errors over five training seeds. Several objectives exhibit a selective-retention window in which marker response decreases while Clean-digit AUC remains near its ERM value.}
\label{fig:app-mnist-replicated}
\end{figure}

\begin{figure}[t]
\centering
\includegraphics[width=\textwidth]{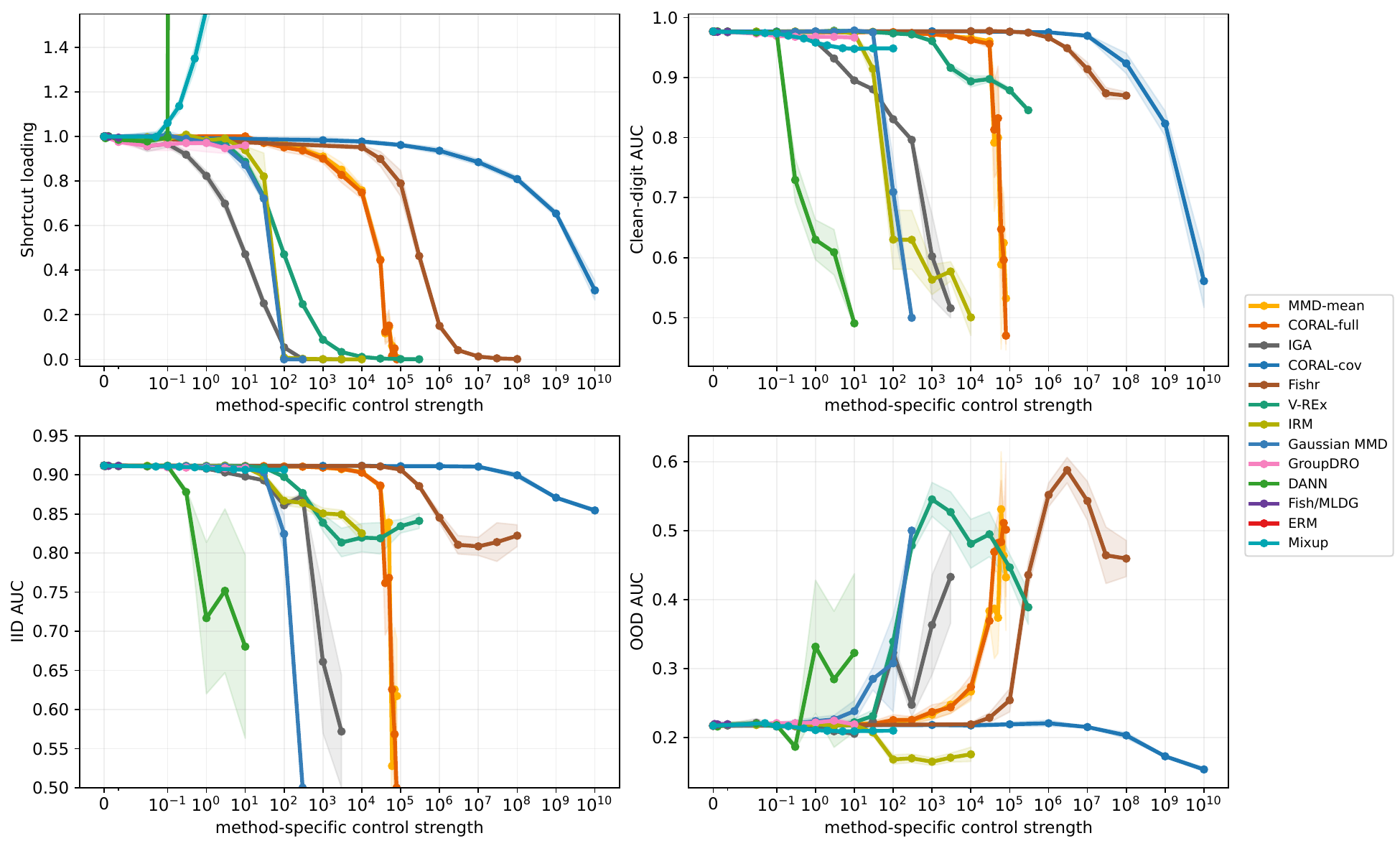}
\caption{MNIST regularization paths with matched but non-overlapping digit images across four environments. Digit and label counts, label-flip positions, and minibatch indices are aligned. Curves and shaded bands show means and standard errors over five training seeds. Shortcut attenuation and digit-feature loss overlap more closely than in the replicated setting.}
\label{fig:app-mnist-disjoint}
\end{figure}

\paragraph{Two environment allocations.}
In the \emph{replicated-digit} setting, every environment contains the same digit images and the same task-label flip mask. Each has $43{,}370$ training and $10{,}840$ IID-holdout examples, with respectively $4{,}337$ and $1{,}084$ examples per digit. There are $1{,}084$ flipped training labels and $271$ flipped IID labels per digit in every environment. A shared holdout assignment keeps the training and IID image sets disjoint across environments. Only the marker assignment differs.

In the \emph{matched-disjoint} setting, the environments receive four distinct images from each same-digit matched quartet. Matching uses six principal components of $7\times7$ average-pooled image descriptors and a Morton ordering; within-quartet assignment approximately balances the coarse image means across environments. Each quartet shares its label-flip status. The resulting $54{,}200$ source images are unique across environments. Each environment has $10{,}840$ training and $2{,}710$ IID examples: respectively $1{,}084$ and $271$ per digit, with $271$ and $68$ flipped labels per digit. Digit counts, task-label counts, and the positions of flipped examples within the matched sequence agree exactly; grayscale images and their higher-order statistics may still differ. Before the IID split, the maximum pairwise mean-image RMSE is $1.54\times10^{-3}$ on the $[0,1]$ scale. Both settings use synchronized minibatch indices and no environment-specific augmentation. The comparison jointly varies residual stable-feature mismatch and per-environment sample size.

\paragraph{Replicated-digit results.}
In the replicated-digit run, \cref{fig:app-mnist-replicated} is consistent with a broad selective-retention window for several objectives when stable-image and label variation across environments is removed. At $\lambda=10^6$, MMD-mean yields the across-seed mean triplet $(Q_s,\text{Clean-digit AUC},\text{OOD AUC})=(0.0098\mathbin{\pm}0.0011,0.981\mathbin{\pm}0.002,0.734\mathbin{\pm}0.002)$, compared with \ERM's $(1,0.977\mathbin{\pm}0.001,0.215\mathbin{\pm}0.001)$, where uncertainties denote standard errors over five seeds. CORAL-full at $10^6$ gives $(0.010\mathbin{\pm}0.002,0.980\mathbin{\pm}0.003,0.731\mathbin{\pm}0.004)$. V-REx at $3\times10^4$ and Fishr at $3\times10^7$ give $(0.006\mathbin{\pm}0.002,0.974\mathbin{\pm}0.001,0.735\mathbin{\pm}0.002)$ and $(0.008\mathbin{\pm}0.002,0.978\mathbin{\pm}0.001,0.735\mathbin{\pm}0.003)$, respectively. Across these four method--strength pairs, every seed has $Q_s<0.016$ and Clean-digit AUC above $0.967$. These sampled path points approach the noisy-label digit-only performance level at method-specific transition scales. At the larger sampled strengths, the mean-matching and Gaussian MMD paths lose stable digit information or collapse; IGA and \IRM show earlier damage in this run, while covariance-only CORAL retains substantial shortcut loading over much of its scan.

\paragraph{Matched-disjoint results.}
The matched-disjoint run in \cref{fig:app-mnist-disjoint} shows greater seed variation and more apparent collateral damage despite exact matching of digit and flip counts. At $\lambda=10^3$, V-REx gives the across-seed mean triplet $(Q_s,\text{Clean-digit AUC},\text{OOD AUC})=(0.088\mathbin{\pm}0.008,0.961\mathbin{\pm}0.003,0.545\mathbin{\pm}0.025)$, compared with \ERM's $(1,0.977\mathbin{\pm}0.001,0.217\mathbin{\pm}0.002)$. Across the five V-REx runs, $Q_s$ ranges from $0.055$ to $0.101$ and Clean-digit AUC from $0.953$ to $0.966$, whereas OOD AUC ranges from $0.505$ to $0.637$. Fishr at $10^6$ gives $(0.150\mathbin{\pm}0.014,0.967\mathbin{\pm}0.001,0.552\mathbin{\pm}0.017)$; at $10^7$, the mean marker response decreases further, giving $(0.013\mathbin{\pm}0.002,0.914\mathbin{\pm}0.011,0.543\mathbin{\pm}0.028)$, with clear digit-feature loss. MMD-mean at $4\times10^4$ and CORAL-full at $5\times10^4$ have still larger seed variation, including runs with low Clean-digit AUC. On the sampled grid, shortcut attenuation and stable-feature loss overlap more closely than in the replicated setting. \ERM, Mixup, and GroupDRO retain substantial marker response over their sampled ranges in this run. The sampled covariance-only, adversarial, and gradient-alignment paths also do not display a uniformly high-retention suppression window. In particular, a high OOD AUC accompanied by large marker response and near-chance Clean-digit AUC is not counted as stable-feature recovery.

\paragraph{Model, objectives, and scan protocol.}
The predictor uses the width and spatial-subsampling convention of the two-hidden-layer ColoredMNIST MLP in \citet{arjovsky2019irm}: the rendered luminance image is subsampled from $28\times28$ to $14\times14$, passed through two width-$390$ ReLU layers, and mapped to a single logit. Every path point starts afresh and runs $2{,}000$ float32 Adam updates with learning rate $\simeq2.82\times10^{-3}$, an explicit parameter-squared penalty with coefficient $\simeq8.60\times10^{-4}$, gradient-norm clipping at $10$, and $128$ matched samples per environment per update. The task loss is binary cross-entropy (BCE). There is no penalty annealing, learning-rate schedule, or early stopping. The dataset construction is fixed at seed $0$; model initialization and minibatch sampling use training seeds $\{0,1,2,3,4\}$. Curves report the across-training-seed mean and standard error.

The objectives follow Appendix~\ref{app:empirical-objectives}, with BCE risks on minibatches and the following normalization details. Mean- and covariance-matching terms divide squared discrepancies by $390$ and $390^2$, respectively, following the elementwise-mean convention of DomainBed \citep{gulrajani2021domainbed}; CORAL-full sums the mean and covariance terms. Gaussian MMD averages five RBF bandwidths $\{0.1,0.5,1,2,10\}$ using the biased V-statistic. Fishr matches instantaneous diagonal variances of per-example classifier gradients, including the bias, without EMA. Fish/MLDG uses the same full-parameter gradient-inner-product surrogate as in Figure~6. DANN alternates one linear-domain-head Adam update (learning rate $0.01$, weight decay $0.001$) with each predictor update. The horizontal axis is the group-weight update rate for GroupDRO, the Beta concentration for Mixup, and the penalty/adversarial coefficient for the remaining non-ERM methods. Control strengths therefore have method-specific scales.

Within each seed, all methods share one zero-strength baseline. MMD-mean, CORAL-full, and Gaussian MMD curves end at the first point with mean Clean-digit AUC $\leq0.55$; this rule also applies to IGA and IRMv1 in the matched-disjoint setting. Fish/MLDG curves end at the first observed loading instability.

\paragraph{Measurements.}
For a fixed test digit image $x$, let $f_\lambda(x,B)$ be the logit with marker state $B$. Define the marker-marginalized score and counterfactual response by
\begin{equation}
 \bar f_\lambda(x)=\tfrac12\bigl[f_\lambda(x,0)+f_\lambda(x,1)\bigr],\qquad
 Q_s(\lambda)=
 \frac{\E_x|f_\lambda(x,1)-f_\lambda(x,0)|}
      {\E_x|f_0(x,1)-f_0(x,0)|}.
 \label{eq:app-mnist-measurements}
\end{equation}
The panels show shortcut loading $Q_s$, \emph{Clean-digit AUC} of $\bar f_\lambda$ against the unflipped digit label, and factual IID/OOD AUC against the noisy task label. Clean-digit AUC is the task-based stable-retention diagnostic. Selective retention combines low marker response with preserved Clean-digit AUC.

\paragraph{Interpretation.}
We compare the MNIST paths qualitatively with the local signatures in Appendix~\ref{app:fingerprint-derivations}. These signatures depend on the reference state; in particular, the \Rfour{}-leading V-REx expression in \cref{eq:app-vrex-equal-null} assumes equal null risks across environments. The matched-disjoint V-REx tail illustrates that small but nonzero shortcut loading can still alter OOD rankings when stable-score differences also shrink, even while Clean-digit AUC remains relatively high.

\FloatBarrier
\setcounter{topnumber}{\mnistsavedtopnumber}

\subsection{Collective-mode baseline comparison}
\label{app:collective-additional}

\begin{figure}[htbp]
\centering
\includegraphics[width=\textwidth]{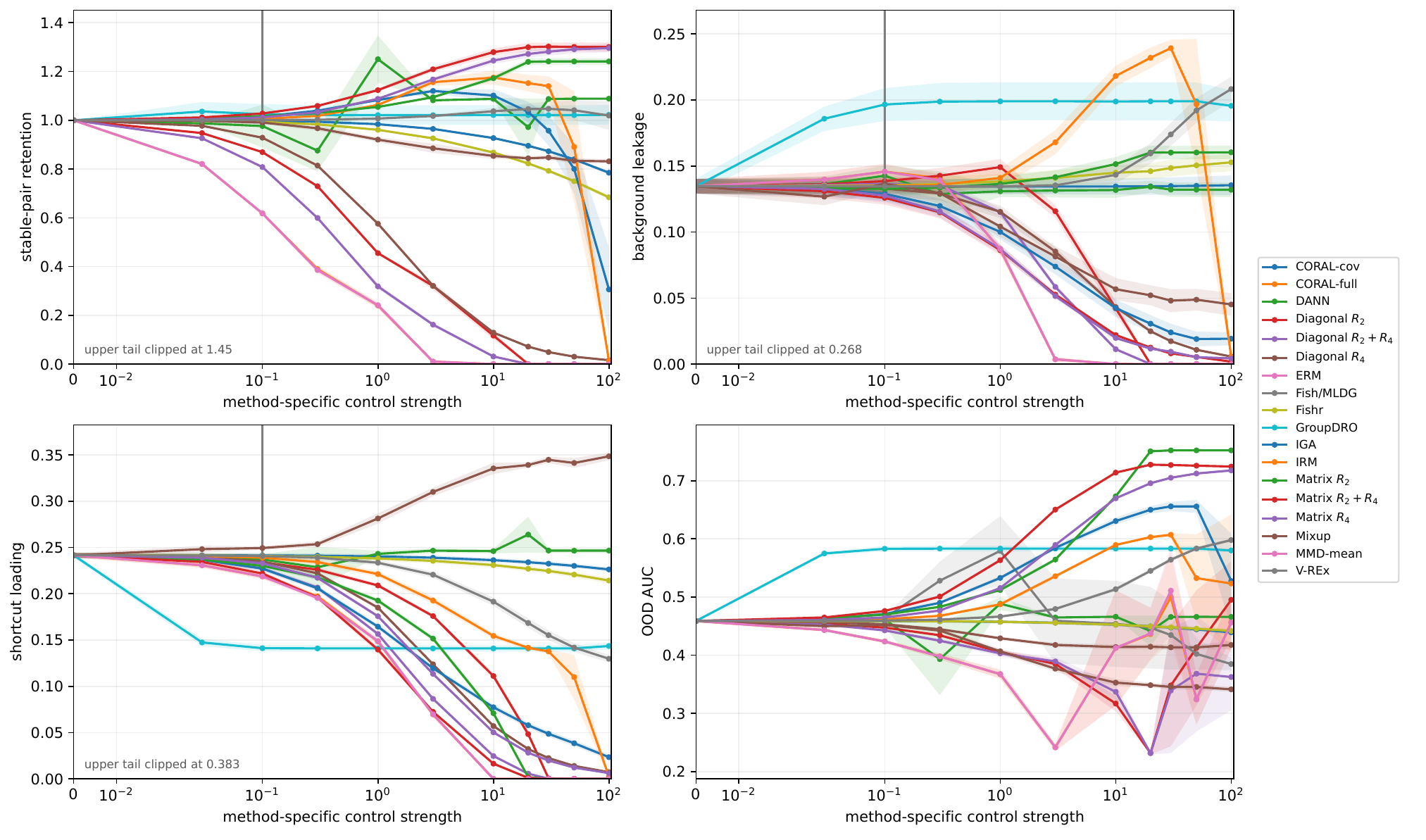}
\caption{Supplementary full algorithm comparison for the collective-mode experiment (\cref{fig:collective-mode} in the main text). The four panels report retention of the compensating pair, background leakage, shortcut loading, and \OOD AUC, respectively. In this construction, the broader set of sampled paths does not uniformly combine preservation of the compensating features, removal of leakage, and improved \OOD performance.}
\label{fig:app-collective-all}
\end{figure}

\Cref{fig:app-collective-all} reports the full baseline comparison for the collective-mode experiment.

\FloatBarrier

\section{Experimental configurations and reproducibility}
\label{app:reproducibility}

Unless stated otherwise, all stochastic training curves report the mean and standard error over five random seeds. Figure~4 uses the pre-specified, theory-blind two-stage empirical calibration protocol detailed below. Settings for the response and MNIST experiments are given in Appendices~\ref{app:soft-mode-response} and~\ref{app:mnist-paths}, respectively.

\subsection{Training protocol}
\label{app:training-protocol}

Table~\ref{tab:app-training-protocol} summarizes the training protocol for Figures~2--7. Except for the deterministic population audit in Figure~2, each update in these experiments computes the environment risk on all samples from each of the four training environments and then averages across environments; no mini-batch sampling is used. Synthetic stochastic curves use seeds $\{0,1,2,3,4\}$ and report mean $\pm$ standard error. No penalty annealing, learning-rate schedule, optimizer reset, or early stopping is used.

\begin{table}[H]
\centering
\footnotesize
\setlength{\tabcolsep}{4pt}
\renewcommand{\arraystretch}{1.08}
\caption{Training protocol for Figures~2--7. Sample counts are per environment; the listed experiments except Figure~2 use Adam with its default momentum parameters.}
\label{tab:app-training-protocol}
\begin{tabular}{@{}>{\raggedright\arraybackslash}p{0.12\textwidth}>{\raggedright\arraybackslash}p{0.26\textwidth}>{\raggedright\arraybackslash}p{0.22\textwidth}>{\raggedright\arraybackslash}p{0.335\textwidth}@{}}
\toprule
\textbf{Figure} & \textbf{Model} & \textbf{Samples and batching} & \textbf{Optimization settings} \\
\midrule
Figure~2 & Single-shortcut population objective & No finite-sample batches & Gradient descent for 50 steps with step size $0.03$; initialized from a perturbation of radius $10^{-3}$. \\
Figure~3 & Bilinear predictor & 256 training samples in (a), 1024 in (b); full-batch over four environments & Adam for 2000 steps with learning rate $0.03$ and $\ell_2$ coefficient $\mu=0.2$. \\
Figure~4 & Single-feature bilinear predictor & 512 training samples; full-batch over four environments; no test set is used & Adam for 2500 steps with learning rate $0.03$, gradient clipping at $20$, and $\mu\in\{0.15,0.2\}$. \\
Figures~5--6 & Bilinear predictor; or plain ReLU MLP with one hidden layer (32) / two hidden layers (32,32) & 256/8192 training samples in Figure~5(a)/(b); 8192 in Figure~6; 2048 test samples; full-batch over four environments & 3000 steps with gradient clipping at $10$; learning rate $0.01$ and $\mu=0.2$ for the bilinear and one-hidden-layer models, and learning rate $0.005$ and $\mu=0.05$ for the two-hidden-layer model. \\
Figure~7 & Three-feature bilinear collective-mode predictor & 1024 training samples and 4096 samples for each of IID/OOD testing; full-batch over four environments & Adam for 3000 steps; learning rate $0.03$ for both predictor and DANN domain head, gradient clipping at $10$, $\mu=0.02$ (domain-head $\mu_B=10^{-3}$). \\
\bottomrule
\end{tabular}
\end{table}

\subsection{Empirical objectives and implementation choices}
\label{app:empirical-objectives}

This section specifies the objectives optimized in the finite-sample synthetic experiments. Let $R_e=\frac{1}{2n_e}\sum_i(y_{ei}-f_\phi(x_{ei}))^2$ denote the full-batch MSE in environment $e$, let $\bar R=E^{-1}\sum_eR_e$, let $h_{ei}$ be the representation used for alignment or domain discrimination, and let $g_e=\nabla_\phi R_e$. Except for the specialized updates of Mixup, GroupDRO, and DANN, all methods minimize $\bar R+\mathcal R_{\ell_2}+\lambda\Omega$. Every finite-sample synthetic experiment uses float64 full-batch training, five random seeds, and the default momentum parameters of PyTorch Adam. For the bilinear model, $h=g\odot x$; for the plain ReLU MLP, $h$ is the final hidden layer.

\begingroup
\footnotesize
\setlength{\LTpre}{5pt}
\setlength{\LTpost}{5pt}
\setlength{\tabcolsep}{4pt}
\renewcommand{\arraystretch}{1.00}
\begin{longtable}{@{}>{\raggedright\arraybackslash}p{0.16\textwidth}>{\raggedright\arraybackslash}p{0.43\textwidth}>{\raggedright\arraybackslash}p{0.365\textwidth}@{}}
\caption{Empirical objectives and implementation choices for the finite-sample regularization-path experiments. Here $\mathsf C_e$ is the sample covariance of representations in environment $e$. Every pairwise average is normalized as $\binom{E}{2}^{-1}\sum_{e<f}$.}\label{tab:app-empirical-objectives}\\
\toprule
\textbf{Method} & \textbf{Objective or additional term used in training} & \textbf{Implementation choice} \\
\midrule
\endfirsthead
\toprule
\textbf{Method} & \textbf{Objective or additional term used in training} & \textbf{Implementation choice} \\
\midrule
\endhead
\bottomrule
\endfoot
ERM; explicit \Rtwo{} / \Rfour{} & ERM uses $\bar R+\mathcal R_{\ell_2}$. For the bilinear parameterization $\beta=g\odot\theta$, explicit \Rtwo{} adds $\frac{\lambda}{2}g^\top\Gamma g$, explicit \Rfour{} adds $\frac{\lambda}{2}\beta^\top\Gamma\beta$, and the joint objective uses their sum. & Scalar experiments set $\Gamma=\gamma$; the matrix version in Figure~7 uses the known oracle $\Gamma_E$, while the diagonal baseline uses $\diag(\Gamma_E)$. \\
IRMv1 & $\lambda E^{-1}\sum_e[\partial_\alpha R_e(\alpha f_\phi)|_{\alpha=1}]^2$. & Uses the autograd derivative at the shared scalar scaling $\alpha=1$ and averages across environments. \\
V-REx & $\lambda E^{-1}\sum_e(R_e-\bar R)^2$. & Uses the population-variance convention over the finite set of environments. \\
GroupDRO & Optimizes $\sum_ep_eR_e+\mathcal R_{\ell_2}$, with $p_e\leftarrow p_e\exp(\eta R_e)/\sum_fp_f\exp(\eta R_f)$. & Initializes $p_e$ uniformly; the scanned control is the group-weight update step size $\eta$. \\
IGA & $\lambda E^{-1}\sum_e\|g_e-\bar g\|_2^2$. & Computes environment-risk gradients with respect to all trainable predictor parameters. \\
Fish/MLDG & $-\lambda\binom{E}{2}^{-1}\sum_{e<f}g_e^\top g_f$. & Uses a cross-environment gradient-alignment surrogate with full-parameter gradient inner products. \\
Fishr & $\lambda d_{\rm cls}^{-1}\sum_{j=1}^{d_{\rm cls}}\Var_e[(v_e)_j]$. & Matches the instantaneous coordinate-wise variances of per-example classifier gradients; the penalty is averaged over classifier coordinates. No EMA or penalty annealing is used. \\
MMD-mean; Gaussian MMD & The former adds $\lambda\overline{\|\bar h_e-\bar h_f\|_2^2}$; the latter adds $\lambda\overline{\operatorname{MMD}^2(h_e,h_f)}$. & Gaussian MMD averages RBF kernels with bandwidths $\{0.1,0.5,1,2,10\}$, uses a biased V-statistic including diagonal entries, and uses at most the first 128 samples from each environment. \\
CORAL-full; CORAL-cov & CORAL-full adds $\lambda\overline{\|\bar h_e-\bar h_f\|_2^2+\|\mathsf C_e-\mathsf C_f\|_F^2}$; CORAL-cov retains only the covariance term. & CORAL-full uses the DomainBed mean-and-covariance implementation with one shared $\lambda$; CORAL-cov is the covariance-only component corresponding to the original Deep CORAL loss. Covariances use the sample denominator $n_e-1$. \\
Mixup & Replaces ERM with the mean MSE on $\alpha x_e+(1-\alpha)x_{e+1}$ and $\alpha y_e+(1-\alpha)y_{e+1}$. & Uses $\alpha\sim\operatorname{Beta}(\lambda,\lambda)$ and cyclically pairs adjacent environments; $\lambda=0$ is defined as the ERM baseline. \\
DANN & The predictor minimizes $\bar R+\mathcal R_{\ell_2}-\lambda\,\operatorname{CE}(D_\psi(h),e)$. & $D_\psi$ is a linear environment-classification head; before each predictor update, the domain head receives one full-batch Adam update on cross-entropy. \\
\end{longtable}
\endgroup

Table~\ref{tab:app-empirical-objectives} describes the finite-sample training objectives, and Appendix~\ref{app:fingerprint-derivations} gives their local $R_2/R_4$ expansions in the controlled bilinear population setting. The DANN derivation uses a ridge-profiled linear MSE surrogate; the DANN curves in Figures~5--6 use the alternating cross-entropy domain classifier listed above.

\subsection{Figure-specific configurations}

\subsubsection{Local onset taxonomy (Figure~2)}
This figure computes the Hessian at the origin directly for a single-shortcut population objective and runs gradient descent for 50 steps from a perturbation of radius $10^{-3}$ using step size $0.03$; the early-growth rate is fitted from the first 20 steps, and its zero crossing in $a_s$ defines the reported dynamic onset. GroupDRO starts from uniform environment weights and updates them with step size $\eta=1$ before each predictor step. Equal null risks make the initial weight changes quadratic in $(w,\theta)$, preserving the \ERM onset. The four training environments are $\zeta_e=(-3,-1,1,3)/\sqrt5$. The first two panels scan $b_F=\delta\in\{0,0.5,1,1.5,2\}$ over $a_s\in[0,2]$ with step size $0.02$, with shared parameters $\mu_w=\mu_\theta=0.2$, $\sigma=1$, and $\Delta_a=0.5$. The IGA panel instead fixes $b_F=0$ and scans $\Delta_a\in\{0,0.1,\ldots,0.8\}$ using the same onset diagnostic.

\subsubsection{Scalar loading validation (Figure~3)}
Panel~(a) uses $X_F=a_FY+b_F\zeta_e+\sigma_F\epsilon$, where $Y\in\{-1,+1\}$ is balanced and $\zeta_e=(-3,-1,1,3)/\sqrt5$. It sets $a_F=0.6$, $\sigma_F=1$, and scans $b_F\in\{0,\allowbreak 0.2,\ldots,1.0,\allowbreak 1.1,\allowbreak 1.2,\allowbreak 1.25,\allowbreak 1.3,\allowbreak 1.4,\allowbreak 1.6,\allowbreak 1.8,\allowbreak 2.0\}$; each environment has 256 training samples, and the bilinear model uses Adam (learning rate $0.03$) for 2000 steps with weight decay $\mu=0.2$. Panel~(b) uses the matched probe $X=(0.6+\Delta_a\zeta_e)Y+\epsilon$, fixes $b_F=0$, and scans $\Delta_a\in\{0,0.1,\ldots,0.8\}$ with $\lambda_{\rm IGA}=1$ and $\lambda_{\rm V\text{-}REx}=5$. It uses 1024 training samples per environment and otherwise the same optimizer and number of steps.

\subsubsection{Theory-blind critical-strength calibration (Figure~4)}
\label{app:critical-calibration-protocol}
The calibration uses the same four-environment scalar construction $X_F=a_FY+b_F\zeta_e+\epsilon$ with $\zeta_e=(-3,-1,1,3)/\sqrt5$ for explicit \Rtwo, \Rtwo{}+\Rfour, MMD-mean, and CORAL-full. We enumerate $a_F\in\{0.5,0.6\}$, $\mu\in\{0.15,0.2\}$, and $b_F\in\{0.5,0.7,0.9\}$, yielding 12 settings per method. Matched IGA instead uses $X_F=(0.5+\Delta_a\zeta_e)Y+\epsilon$ with $(\mu,\Delta_a)=(0.15,0.6),\allowbreak (0.15,0.65),\allowbreak (0.15,0.7),\allowbreak (0.15,0.75)$, so the calibration contains 52 settings in total. For every setting, we first train a fixed coarse grid $\{0,\allowbreak 0.025,\allowbreak 0.05,\allowbreak 0.075,\allowbreak 0.1,\allowbreak 0.15,\allowbreak 0.2,\allowbreak 0.3,\allowbreak 0.4,\allowbreak 0.5,\allowbreak 0.6,\allowbreak 0.8,\allowbreak 1,\allowbreak 1.25,\allowbreak 1.5,\allowbreak 2,\allowbreak 2.5,\allowbreak 3,\allowbreak 4,\allowbreak 5,\allowbreak 6,\allowbreak 8,\allowbreak 10,\allowbreak 12\}$, independently of $\lambda_c^{\rm theory}$. Let $(\bar b_{\mathrm{floor}},s_{\mathrm{floor}})$ be the mean and s.e.m.\ of the final loading at the three largest coarse strengths. At strength $\lambda$, let $(\bar b_\lambda,s_\lambda)$ denote the corresponding mean and s.e.m.\ of final $|\beta|$. We call this point \emph{resolved ordered} iff
\[
  \bar b_\lambda-1.96s_\lambda>
  \bar b_{\mathrm{floor}}+1.96s_{\mathrm{floor}} .
\]
Thus $\bar x\pm1.96\,\mathrm{s.e.m.}$ is used as the usual approximate two-sided $95\%$ normal confidence interval for a mean. The first persistent non-resolved tail and the preceding resolved point form the empirical bracket $[\lambda_{\rm lo},\lambda_{\rm hi}]$. Settings without such a bracket are discarded.

We next use a uniform 9-point refinement as an initial empirical screen within each bracket (21 points for IGA). A setting is retained only when this scan contains at least five \emph{resolved ordered} strengths; its five largest such strengths, i.e., those closest to the transition from the ordered side, define a common local fitting window. To quantify finite-seed variation, we enumerate all $5^5=3125$ ordered five-seed bootstrap draws with replacement. For each draw, we average the loading at the five fixed strengths and fit its zero by ordinary least squares. Figure~4 reports the mean and the $2.5$th--$97.5$th percentile bootstrap interval of these 3125 critical-strength estimates. The bracket and fitting points are selected from empirical loading data. The aggregate weighted local fit is used for quality control: we reject a nonnegative slope, $R^2<0.95$, or an extrapolated root more than half a bracket width outside the empirical bracket. The one-sided confidence criterion restricts the fit to resolved ordered points.

Of the 52 settings, 24 were excluded for having fewer than five \emph{resolved ordered} strengths: four each for \Rtwo{} and \Rtwo{}+\Rfour{}, and eight each for MMD-mean and CORAL-full. The remaining 28 passed all quality checks. Figure~4 displays four settings per method (20 in total), selected using fixed theoretical critical-strength targets.

\subsubsection{Regularization paths and nonlinear transfer (Figures~5--6)}
The bilinear and ReLU experiments share a four-environment stable--shortcut construction: $X_c=a_cY+\sigma_c\epsilon_c$, $X_s=(a_s+\Delta_a\zeta_e)Y+\delta \zeta_e+\sigma_s\epsilon_s$, together with four independent noise features. We use $(a_c,a_s,\delta,\Delta_a,\sigma_c,\sigma_s)=(0.6,0.8,1,0.3,1,0.5)$. Figure~5 compares 256 and 8192 training samples per environment with shared IID/OOD test sets within each seed. Figure~6 and Appendix~\ref{app:extended-nonlinear-paths} use the same 8192-sample training sets and IID/OOD test sets as Figure~5(b), for all five seeds and all methods. Each test environment contains 2048 samples. OOD testing uses $\zeta_e=\pm2$ and reverses the shortcut label coefficient, $a_s\mapsto-a_s$. The bilinear model is trained with Adam for 3000 steps at learning rate $0.01$. Nonlinear validation uses plain ReLU MLPs on the raw inputs: either one hidden layer of width 32 or two hidden layers of widths $(32,32)$, trained for 3000 steps at learning rates $0.01$ and $0.005$, respectively. Functional loading is the mean absolute Jacobian of the prediction with respect to each input coordinate on OOD samples, normalized by the corresponding zero-regularization value for the same method.

\subsubsection{Collective-mode experiment (Figure~7)}
The compensating-pair construction uses two features with opposite environmental backgrounds to test whether the stable signal is a collective direction. Figure~7 and its supplementary diagnostics use four composite training environments. Both $\zeta_{\mathrm{pair}}=(-3,-1,1,3)/\sqrt5$ and $\zeta_{\mathrm{short}}=(1,-3,3,-1)/\sqrt5$ have zero mean and unit second moment, and they are mutually orthogonal. With independent noise $\epsilon_j\sim\mathcal N(0,\sigma^2)$, the training and IID-test features are
\[
\begin{aligned}
X_1&=(a_c+\Delta_a\zeta_{\mathrm{pair}})Y+\delta_c\zeta_{\mathrm{pair}}+\epsilon_1,
&X_2&=-\delta_c\zeta_{\mathrm{pair}}+\epsilon_2,\\[-1mm]
X_3&=(a_s+\Delta_{a,s}\zeta_{\mathrm{short}})Y+\delta_s\zeta_{\mathrm{short}}+\epsilon_3.
\end{aligned}
\]
Thus, when $\Delta_a=0$, $X_1+X_2$ cancels the pair background, whereas $X_1-X_2$ is the background-leakage direction. The OOD environments keep both background axes fixed and only reverse $a_s\mapsto-a_s$. The corresponding oracle mass is $\Gamma_E=d_{\mathrm{pair}}d_{\mathrm{pair}}^\top+d_{\mathrm{short}}d_{\mathrm{short}}^\top$, where $d_{\mathrm{pair}}=(\delta_c,-\delta_c,0)^\top$ and $d_{\mathrm{short}}=(0,0,\delta_s)^\top$. Hence $\bm\beta^\top\Gamma_E\bm\beta=\delta_c^2(\beta_1-\beta_2)^2+\delta_s^2\beta_3^2$. The diagonal baseline uses $\diag(\Gamma_E)$, whereas the matrix \Rtwo{} and \Rfour{} objectives use the full oracle $\Gamma_E$ to isolate the spectral mechanism.

The construction uses $a_c=0.6$, $\delta_c=1$, $a_s=0.75$, $\delta_s=1$, $\sigma=1$, and $\Delta_a=\Delta_{a,s}=0$. Each training environment contains 1024 samples, and each IID and OOD test environment contains 4096 samples. For each $\lambda\in\{0,0.03,0.1,0.3,1,3,10,20,30,50,100\}$, training runs for 3000 steps over five seeds. The stable-mode loading is $(\beta_1+\beta_2)/\sqrt2$, normalized by its absolute value for the same method and seed at $\lambda=0$; the shortcut loading is $|\beta_3|$.

\subsubsection{Optimization dynamics (Figure~\ref{fig:objective-dynamics})}
\label{app:objective-dynamics-protocol}

Figure~\ref{fig:objective-dynamics} and Appendix~\ref{app:objective-relaxation} use the bilinear predictor $f_\Theta(X)=w\theta X$, with $\Theta=(w,\theta)$ and representation $wX$. Four equally weighted environments have balanced $Y\in\{-1,1\}$ and
\begin{equation}
    X_e=(0.5+0.2\zeta_e)Y+0.5\zeta_e+\epsilon,
    \qquad \zeta_e\in\frac{1}{\sqrt5}\{-3,-1,1,3\},
    \quad \epsilon\sim\mathcal N(0,1).
    \label{eq:app-objective-relaxation-data}
\end{equation}
We use population squared loss with factor $1/2$, weight decay $0.2\|\Theta\|^2/2$, and the environment-variance normalization of Appendix~\ref{app:fingerprint-derivations}. Gaussian MMD uses a unit-bandwidth RBF kernel; Fishr uses per-example classifier-gradient variances.

At each equilibrium $\Theta^\star$ with positive Hessian curvature, let $(\kappa_{\min},v_{\min})$ be the smallest eigenpair of the full objective Hessian, with $\|v_{\min}\|=1$. We run deterministic gradient descent with fixed step size $\eta=0.03$ from $\Theta_0^\pm=\Theta^\star\pm\varepsilon v_{\min}$ and measure the projected half-difference
\begin{equation}
    d_k=\frac12v_{\min}^{\top}(\Theta_k^+-\Theta_k^-),
    \qquad
    N_{90}=\min\{k\geq0:|d_k/d_0|\leq0.1\}.
    \label{eq:app-objective-recovery-measurement}
\end{equation}
Perturbation amplitudes are $\varepsilon=10^{-3}\|\Theta^\star\|$ on the acquired branch and $\varepsilon=10^{-3}\min(0.1,\sqrt{\kappa_{\min}})$ at the origin; all sampled points satisfy $0<\eta\kappa_{\max}<1$. The predicted steps use \cref{eq:main-recovery-steps}; predicted trajectories use $d_k/d_0\simeq(1-\eta\kappa_{\min})^k$.

\end{document}